%% file: paper.tex
\pdfoutput=1

\documentclass[journal]{IEEEtran}

\usepackage{graphicx} 
\usepackage{amsmath}
\usepackage{amssymb}
\usepackage[usenames,dvipsnames]{xcolor}
\usepackage[lined,boxed,commentsnumbered]{algorithm2e}

\newcommand{\red}[1]{{#1}}

\newcommand{\R}{\mathbb{R}}
\renewcommand{\P}{\mathcal{P}}
\newcommand{\Ri}{\mathcal{R}}
\renewcommand{\S}{\mathcal{S}}
\renewcommand{\k}{\kappa}

\renewcommand{\H}{\mathcal{H}} 
\newcommand{\Nm}{\vec{\mathcal{N}}} 
\newcommand{\K}{K} 
\newcommand{\si}{SI} 

\begin{document}
\title{A Complete System for Candidate Polyps Detection in Virtual Colonoscopy}




\author{Marcelo~Fiori, Pablo~Mus\'{e}, and~Guillermo~Sapiro
\thanks{M. Fiori and P. Mus\'{e} are with the Instituto de Ingenier\'{i}a El\'{e}ctrica, Universidad de la Rep\'{u}blica, Herrera y Reissig 565, 11300, Uruguay.}
\thanks{G. Sapiro is with the Department of Electrical and Computer Engineering, Duke University, Durham, North Carolina, 27708, USA.}}


\maketitle

\begin{abstract}
Computer tomographic colonography, combined with computer-aided detection, is a promising emerging technique for colonic polyp analysis. We present a complete pipeline for polyp detection, starting with a simple colon segmentation technique that enhances polyps, followed by an adaptive-scale candidate polyp delineation and classification based on new texture and geometric features that consider both the information in the candidate polyp location and its immediate surrounding area. The proposed system is tested with ground truth data, including flat and small polyps which are hard to detect even with optical colonoscopy. For polyps larger than $6mm$ in size we achieve $100\%$ sensitivity with just $0.9$ false positives per case, and for polyps larger than $3mm$ in size we achieve $93\%$ sensitivity with $2.8$ false positives per case. 
\end{abstract}



\section{Introduction}

Colorectal cancer is nowadays the second leading cause of cancer-related deaths in the United States (only surpassed by lung cancer), and the third cause worldwide \cite{whocancer}.
The early detection of polyps is fundamental, allowing to reduce mortality rates up to $90\%$ \cite{prevention}. 
Polyps can be classified into sessile, pedunculated, and flat, according to their morphology. Pedunculated polyps are attached to the colon wall by a stalk, sessile polyps grow directly from the wall, and flat polyps have less than $3mm$ of elevation above the colonic mucosa \cite{zalis}.
Nowadays, optical colonoscopy (OC) is the most used detection method due in part to its 
 high performance. However, this technique is invasive and expensive, making it hard to use in large screening campaigns.

Computed Tomographic Colonography (CTC), or Virtual Colonoscopy (VC), is a promising alternative technique that emerged in the 90's \cite{Vining1994}. It uses volumetric Computed Tomographic data of the cleansed and air-distended colon. Distention is carried out by placing a thin tube into the rectum, and
performing the colon insufflation with room air or carbon dioxide. It is less invasive than optical colonoscopy, and much more suitable for screening campaigns once its performance is demonstrated. However, VC is less popular than OC not only because it is a relatively new technique, but also because, contrarily to OC, it is not yet reimbursed by insurance companies. 
On the other hand, in OC, on the average only around $70\%-80\%$ of the colon can be explored \cite{voyage}. Incomplete studies due to obstructing colorectal lesions, colon twists, or anatomical variations are not rare ($5\%$ to $15\%$ of OC examinations) \cite{blachar2007}, and there is an additional important risk of colon perforation. In a large study by Kim {\it et al.} \cite{kim2007}, where about $3000$ patients went through OC and another $3000$ through VC, seven perforations occurred in OC while none was reported in VC.

Nevertheless, it takes more than $15$ minutes for a trained radiologist to complete a CTC study, and the performance of the overall optical colonoscopy is still considered better. In this regard, Computer-Aided Detection (CAD) algorithms can play a key role, assisting the expert to both reduce the procedure time and improve its accuracy \cite{sundaram,hong,VanRavesteijn2010,Nappi2002,Nappi2007,Yoshida2002}.

Flat polyps are of special interest because these are an important source of false negatives in CTC, and although there are different opinions, many authors claim that flat polyps are around $10$ times more likely to contain high-grade epithelial dysplasia\footnote{An abnormality of development in cells that may become \textit{cancer in situ} or \textit{invasive cancer}.} \cite{fidler,fidler2008,wobler}.

\red{There are numerous discussions regarding the potential risks of the polyps according to their size. Even though some authors consider that ``small'' polyps may not represent risk, some gastroenterologists disagree \cite{Aschoff2008}. 
Summers \cite{Summers2010} claims that one of the major challenges in the field is in increasing sensitivity for smaller polyps, and Church \cite{church} states that
small adenomas can still be clinically significant and should not be ignored.} 
At the same time, Bond \cite{Bond2002} declares that the major disadvantage of VC is its current low performance for flat polyps.

The goal of the work presented in this paper is to exploit VC precisely to automatically flag (mark for attention of the expert) colon regions with high probability of being polyps, with special attention to results in challenging small and flat polyps. 
Towards this aim, we propose a complete pipeline that starts with a novel and simple segmentation step \red{(segmentation of the colon surface/lumen)}. We then introduce geometrical and textural features that take into account not only the candidate polyp region, but the surrounding area at multiple scales as well. 
This way, our proposed CAD algorithm is able to accurately detect candidate polyps by measuring local variations of these features. The whole algorithm is completely automatic and produces state-of-the-art results. This paper extends our previous conference publication
\cite{Fiori2010}.

The rest of this paper is organized as follows. In Section \ref{review} we briefly review prior related work and in Section \ref{pipeline} we present an overview of the whole proposed pipeline. We address the colon segmentation problem in Section \ref{segment} and the feature extraction and classification in Section \ref{features}. 
In Section \ref{sect:classif} we describe the classification and evaluation method, and in Section \ref{sect:results} we present numerical results. The discussion is presented in Section \ref{discussion} and we conclude in Section \ref{conclusions}.

\subsection{Virtual Colonoscopy CAD Review}
\label{review}

Automatic polyp detection is a very challenging problem, not only because the polyps can have different shapes and sizes, but also because they can be located in very different surroundings.
Most of the previous work on CAD of colonic polyps is based on geometric features, some using additional CT image density information, but none of them takes into account the (geometric and texture) information of the tissues {\it surrounding} the polyp. 
This local and adaptive differential analysis is part of the contributions of this work.

Early work on CAD methods by Vining {\it et al.} \cite{vining} is based on the detection of abnormal wall thickness. Since then, several different approaches were proposed. Most of them have a segmentation step first, and then the classification step itself. We briefly discuss the main ideas in each stage.

\noindent \textbf{Segmentation}

The most common segmentation techniques are based on region growing and thresholding methods. 
Among others, Yoshida {\it et al.} \cite{yoshida} use a Gaussian smoothing and thresholding. 
Summers {\it et al.} \cite{Summers2005a} use a region growing scheme and obtain the final segmentation by thresholding.
Paik {\it et al.} \cite{paik} use thresholding as well, followed by morphological dilation.
Chen {\it et al.} \cite{Chen2000} use a modified adatptive vector quantization followed by region growing.
Sundaram {\it et al.} \cite{sundaram} start from manually selected seeds and segment the region by thresholding with marching cubes.
Franaszek {\it et al.} \cite{Franaszek2006} use a modified region growing, fuzzy connectedness, and Laplacian level set segmentation to obtain a smooth surface. 
Chen {\it et al.} \cite{Chen2009b} also use a level set approach, whose velocity is determined by a Bayesian classification of the pixels.

Assuming that the value of each voxel is a mixture of different tissue types, Wang {\it et al.} \cite{Wang2006} use the statistical expectation-maximization (EM) algorithm to estimate the parameters of these tissue types (they use air, soft tissue, muscle, and bone/tagged material). The voxels are classified as air, mixture of air with tissue, mixture of air with tagged materials, or mixture of tissue with tagged materials, and then dilation/erosion operations are performed.

\red{Yamamoto {\it et al.} \cite{yamamoto} use a vertical motion filter at the fluid level. Serlie {\it et al.} \cite{serlie} use a sophisticated method to address the T-junction artifact with promising results.}

Region growing and level sets seems to be the current chosen techniques for colon segmentation. However, not much work has been done in comparing the smoothing techniques (or the regularization term in the level set method) to see which one is more adapted to polyp detection. The segmentation technique here proposed is geared toward the subsequent step of polyp detection, and simultaneously segments and prepares the obtained surface for this task.

\noindent \textbf{Classification}

The main variations in this topic are both in the features used and in the classification method. 
Summers {\it et al.} \cite{summers2005} detect polyps larger than $10mm$ by computing mean curvatures and sphericity ratio, and present results over a large screening patient population.
Yoshida {\it et al.} \cite{yoshida} use the \textit{shape index} (defined later in this paper) and \textit{curvedness} as geometric features, applying fuzzy clustering and then using directional gradient concentration to reduce false positives.
Paik {\it et al.} \cite{paik} also use geometrical features, computing the Surface Normal Overlap (SNO) instead of calculating curvatures.
Wang {\it et al.} \cite{wang2005_medphys} compute a global curvature, extract an ellipsoid, and analyze morphological and texture features on this ellipsoid. They reach $100\%$ sensitivity with a relative low false positives (FP) rate, using heuristic thresholds and texture features. 
Hong {\it et al.} \cite{hong} map the 3D surface to a rectangle, use 2D clustering, and reduce false positives with shape and texture features.
Sundaram {\it et al.} \cite{sundaram} compute curvatures via the Smoothed Shape Operators method, and use principal curvatures and Gaussian curvatures to detect polyps. 
 All these described techniques based on local geometric computations suffer from a high dependence on the regularity of the polyp shape itself, ignoring how pronounced it is with respect to the surrounding area. Using geometry alone is also very sensitive to the accuracy of the colon segmentation.

More recently, van Wijk {\it et al.} \cite{vanwijk} proposed a Partial Differential Equation (PDE) motion that flattens only the polyp-like shapes, and then they consider the difference between the original and the processed images. The main drawback of this approach is that the PDE motion is not capable of flattening polyps which are already flat. Consequently, for flat polyps, the difference between the original surface and its smoothed version is too small to be detected by the algorithm. Ong {\it et al.} \cite{Ong2011} use the neighborhood to compute curvatures; instead of using only a 1-ring neighborhood to approximate second derivatives and compute the curvatures, they use a larger region to reduce the effect of noise. \red{Konukoglu {\it et al.} \cite{konu2} propose a heat diffusion field method to characterize polyps. }
Proprietary algorithms \cite{bogoni,taylor} have been reported as well, but with no better results than the methods mentioned above. 

\noindent \textbf{Complementary techniques}

In addition to these works that tackle either the segmentation, classification or both as a complete CAD system, some other complementary techniques were proposed. These algorithms are located in between the segmentation and the classification steps, or after classification as a false positive reduction technique. For example, G\"otk\"urk {\it et al.} \cite{gotkurk} proposed a technique to reduce the false positives based on features calculated from three random orthogonal sections, and then classifying with SVM.
Konukoglu {\it et al.} \cite{konukoglu2007} proposed to introduce a polyp enhancement stage, in between the segmentation and classification stages of the pipeline in \cite{paik}. The polyp enhancement is based on the heat equation, with the conductivity term adapted to the local geometric properties of the surface. 
Suzuki {\it et al.} \cite{suzuki} use artificial neural networks to reduce the false positives of the algorithm in \cite{yoshida} described above. 
The results are very promising, achieving $96.4\%$ sensitivity (over $28$ polyps) with $1.1$ FP per case. However, this evaluation does not take into account small lesions and polyps submerged in fluid.
\newpage

\noindent \textbf{General comments}


The list of references included in this section is by no means exhaustive. We do not intend here to present a full review of existing techniques, but to discuss a set of works that are representative of the diversity of the proposed approaches for polyp detection. Direct quantitative comparison of the classification results obtained by these methods is strictly speaking meaningless, since they were obtained using different databases.

Table \ref{tabla_review} summarizes the results reported by the methods discussed above. Notice that these results were obtained using databases containing polyps larger than $6mm$ in size (or larger than $10mm$ in some databases). 

To the best of our knowledge, no algorithm reported in the literature can detect small polyps properly. On the other hand, for polyps larger than $6mm$ in size, no algorithm can achieve $100\%$ sensitivity with less than one false positive per case. 
It is crucial to keep improving these rates, 
this is the reason why VC is still an active research field.

\input{tabla_review.tex}

\subsection{Overview of the Proposed System}
\label{pipeline}

The main goal we are addressing in this work is to highlight/flag all the candidate polyps, so the radiologist can quickly check them. It is crucial to minimize the false negatives, keeping a reasonable false positives number. We achieve this by a four-steps process that is completely automatic and constitutes the entire end-to-end algorithm, from data to candidate polyps flagging.


\begin{figure*}[!t]
\begin{center}
\includegraphics[width=\textwidth]{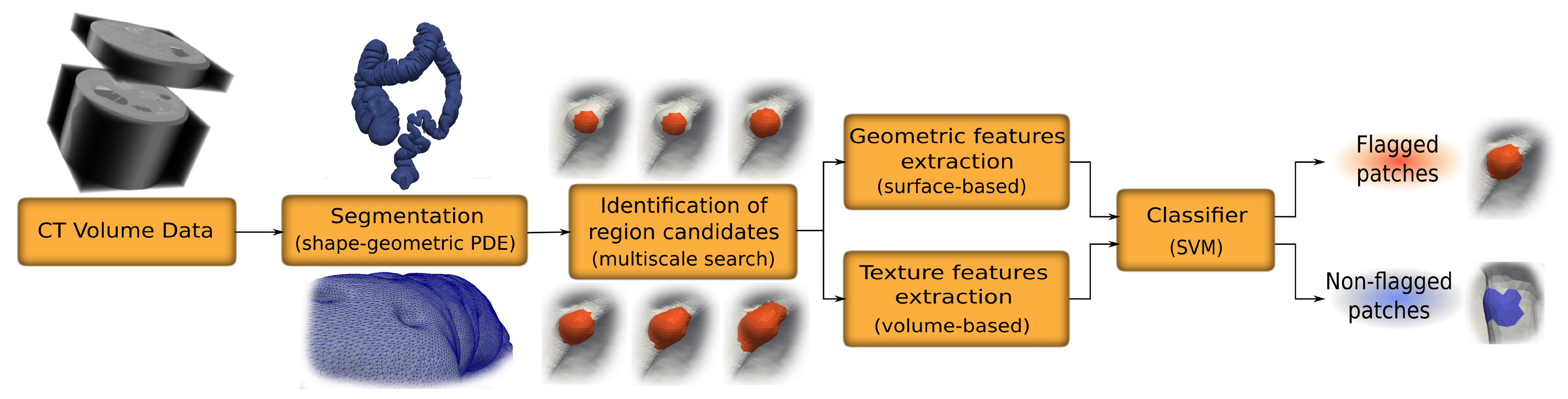}
\caption{Basic pipeline of the proposed polyp flagging system.\label{fig:pipeline}}
\end{center}
\end{figure*}

The proposed system with its four steps is illustrated in Figure \ref{fig:pipeline}. The first step is colon segmentation, which takes as input the CT volume data, and produces a $3D$ mesh representing the colon surface.
In the second step, from the segmented mesh we perform an adaptive-scale search of candidates in order to capture the appropriate polyp size, obtaining a set of candidate patches. 
In the third step, both the CT volume data and the segmented colon surface are used to compute geometrical and textural features for each candidate patch that was identified in the previous stage. 
The final step consists of a machine learning algorithm that uses the computed features to classify patches as polyps or normal tissue. In the following sections we describe each of these steps in detail.

\subsection{Paper contributions}

In this work we introduce several novelties at every stage of the polyps detection system. These novelties result from the consideration of several alternatives. As it will be clear from the following, the solutions that were kept outperform other alternatives that were considered in this work, and significantly contribute to the quality of the results presented in Section \ref{sect:results}. These contributions are:

\begin{itemize}
\item We propose a simple segmentation approach, specifically designed to reduce tagged fluid artifacts;
\item The smoothing PDE algorithm outperforms the classical curvature motions in terms of polyp enhancement and classification;
\item Our proposed \textit{adaptive-scale} candidate search allows to precisely delineate the polyp region;
\item We propose to characterize the texture in the tissue by \textit{Haralick} features, which have proved successful in other applications in the image processing and computer vision literature;
\item Instead of only measuring texture or geometric features within a candidate region, we consider measures that explicitly take into account the tissue properties in the surrounding area. We call these \textit{differential features}, and we show that their use improves the classification performance.
\item We explicitly deal with the class imbalance problem (polyp vs non-polyp) in the learning step. We consider three machine learning techniques that were specifically designed to deal with this problem. The best classification results were obtained with \textit{cost sensitive} learning.
\end{itemize}

All these contributions will be detailed in the next sections.

\section{Colon Segmentation}
\label{segment}

The segmentation of the colon surface, which is critical in particular to compute geometric features, is divided into two parts: a pre-processing stage for dealing with the air-fluid composition of the colon volume, and a second stage that consists on smoothing the pre-processed image and obtaining the final colon surface by thresholding the smoothed volume. The overall procedure here presented is very simple and computationally efficient, leading to the state-of-the-art classification results later reported.

\subsection{Classifying CT regions}

\begin{figure}[h!]
\begin{center}
\includegraphics[width=\columnwidth]{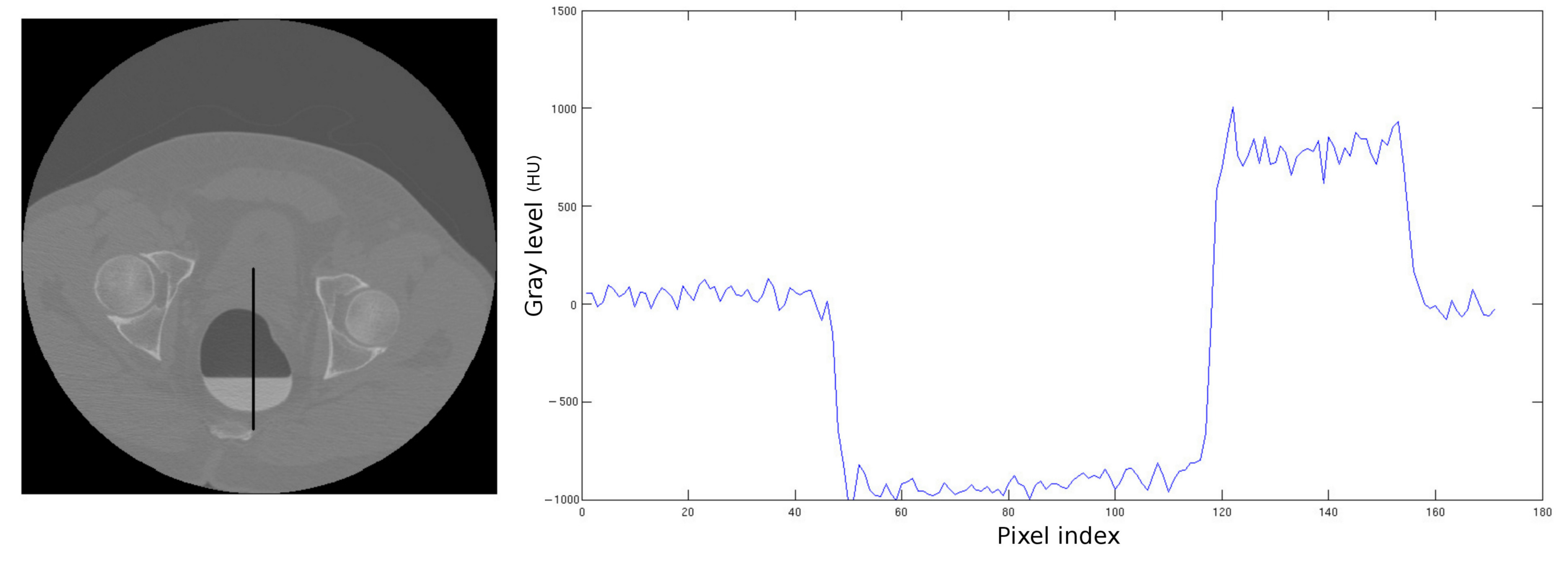}
\caption{CT slice and its different gray values for air, fluid and normal tissue, along the vertical profile.\label{fig:perfil}}
\end{center}
\end{figure}

All the cases from the used database have the same preparation, which includes solid-stool tagging and opacification of luminal fluid, and can be noticed as the white liquid in Figure \ref{fig:perfil}. One of the strongest difficulties concerning the segmentation of the colon from abdominal volumes in CT is the presence of this tagged fluid and its interfaces with air and tissue. Figure \ref{fig:perfil} shows a CT slice and its pixel values over the highlighted vertical profile. At first sight there are three clearly distinguishable classes: the lowest gray levels correspond to air, the highest levels correspond to fluid, and the middle gray values correspond to tissue. Nevertheless, there are around $6$ interface voxels between air and fluid whose gray values, due to continuity, lie within the normal tissue range. Therefore, a na\"{i}ve approach based on gray values only, ignoring the physical nature of the tissue and its environment, is not suitable for proper tissue classification and segmentation.

The goal of this first pre-processing stage is to obtain an initial segmentation, as close as possible to the actual colon boundary (border). This border is diffuse due to the CT resolution, and if we adopt a binary segmentation approach (interior/exterior) the border will be necessarily bumpy. We prefer not to lose information about the gray decay throughout the border, so we compute a volume $u_0$ intended to have homogeneous values in the colon interior and exterior, and a smooth transition between them.

\red{In order to do that, it makes sense to assign to each voxel the likelihood of being air, fluid or air-fluid interface. }
The air and fluid likelihoods are estimated as follows. We pick any CT study and manually segment the regions corresponding to air (class $w_1$) and tagged fluid (class $w_2$). Then, their gray level distributions are learned using standard kernel density estimation techniques \cite{kernelsmoothing}.
 This step is done just once.
Then, at runtime, for each new study, the gray level histogram is computed. These histograms are trimodal, with peaks corresponding to air, tissue, and tagged fluid. We assume that the distribution of the gray values, both for air and tagged fluid, are shifted versions of the previously learned ones (this assumption was experimentally validated). Then, the air and tagged fluid distributions for the considered study, are determined by localizing their peaks.  These shifted functions are used to assign the air and tagged fluid likelihood values to the voxels.  

Note that this assignment fails on the air-fluid and air-fluid-tissue interfaces. 
The basic idea for assigning a value to these voxels is presented next.

Here we take advantage of the physics of the problem, and in particular of the gravity and the position of the patient: the subject is laid horizontally
so the interface between the fluid and the air is a plane parallel to the floor. The voxels situated on the interface then have a large gradient in the
vertical direction. 
However, the transition is about $6$ voxels wide for the standard data resolution used in this work, so the computations should be done taking this into account. 
Additionally, if a given voxel belongs to the interface layer, it is expected that at least half of the neighbor voxels at the same horizontal plane also belong to the interface layer. 

The implementation of these criteria is as follows. A cubic neighborhood around each voxel $\mathbf{x}$ is considered, and for each one of the ``columns'' that result of fixing the $x$ and $y$ coordinates, the air-likelihoods of the upper voxels and the fluid-likelihoods of the lower voxels are accumulated. If the tested voxel belongs indeed to the interface layer, then all these air and fluid likelihoods will be high. The value $IC(\mathbf{x})$ that represents the confidence level of $\mathbf{x}$ being an interface voxel is then an increasing function of this accumulated measures. 
Algorithm \ref{algo} provides a pseudo-code that represents this procedure. 

In order to guarantee that a high value is associated to every notoriously interior voxel, we assign to the initial segmentation $u_0$ the maximum of these three values, namely, the air and fluid likelihoods and the interface confidence level: $u_0(\mathbf{x}) = \max\left(p(x|w_1),p(x|w_2),IC(\mathbf{x})\right)$. This way, internal voxels (air, fluid and interface) will have values close to one, external voxels will have values close to zero, and the transition will be smooth across the actual colonic wall.

\begin{algorithm}
\For{each voxel (x,y,z)}{
	sum=0\;
	\For{$i$=$-1$ \KwTo $1$}{
		\For{$j$=$-1$ \KwTo $1$}{
			\For{$k$=$1$ \KwTo $2$}{
			sum += $p$($($x+i,y+j,z+k$)|w_1$)\;
			sum += $p$($($x+i,y+j,z-k$)|w_2$)\;
}
}
}
$IC$($x,y,z$) = sum/18\;
}
\caption{Computation of interface confidence level. \label{algo}}
\end{algorithm}

After the computation of the initial segmentation $u_0$, some spurious (isolated) voxels may have high values (bones for example, or simply noise), so we clean the initial segmentation by keeping the connected components\footnote{Actually, since the initial segmentation $u_0$ is not binary, a (conservative) threshold of $0.6$ is considered to separate the connected components.} containing some chosen voxels used as seeds. 
It is important for the system to be automatic, so no human intervention should be needed to choose these seeds. In order to do that, the seeds are automatically detected by choosing the voxels with largest values of $IC(\mathbf{x})$, since these high values only occur at the interface between air and fluid. Voxels with the largest values of the air likelihood can be used too. After the first seed has been used and a connected component has been obtained, we look for a new seed outside this component, apply the filter with this seed, and append the new corresponding connected component to the previous ones. This iterative procedure continues until no seed outside the extracted connected components is found or a maximum number of connected components is reached. Usually, only two iterations are sufficient. This way, the segmentation step is able to handle the lumen discontinuities problem and to obtain the multiple pieces of the colon that might be disconnected.

\begin{figure}[h!]
\begin{center}
\includegraphics[width=0.8\columnwidth]{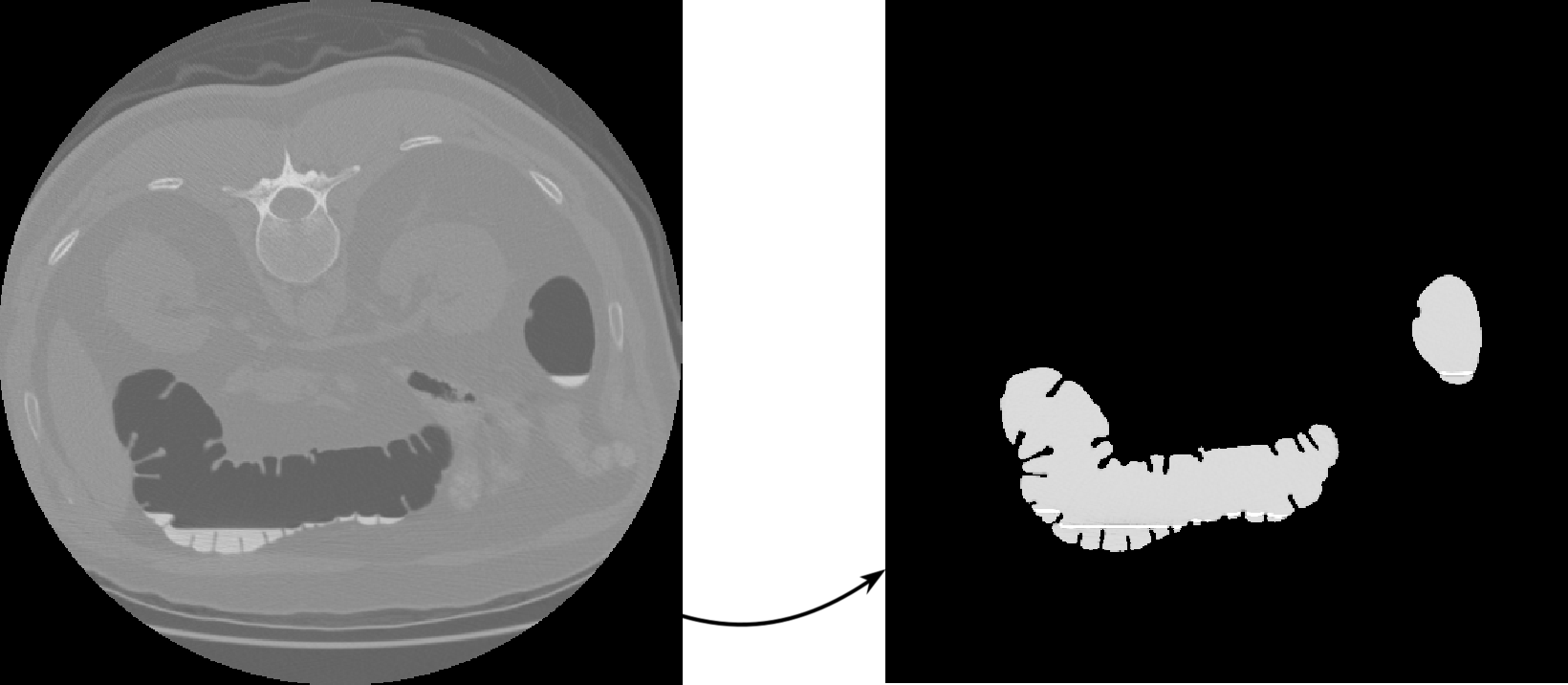}
\caption{ CT slice and its corresponding slice on the initial segmentation $u_0$. Patient in prone position.\label{fig:prmap}}
\end{center}
\end{figure}

Figure \ref{fig:prmap} shows a slice of the original volume data and the same slice of the initial segmentation $u_0$.

\begin{figure}[h!]
\begin{center}
\includegraphics[width=0.48\columnwidth]{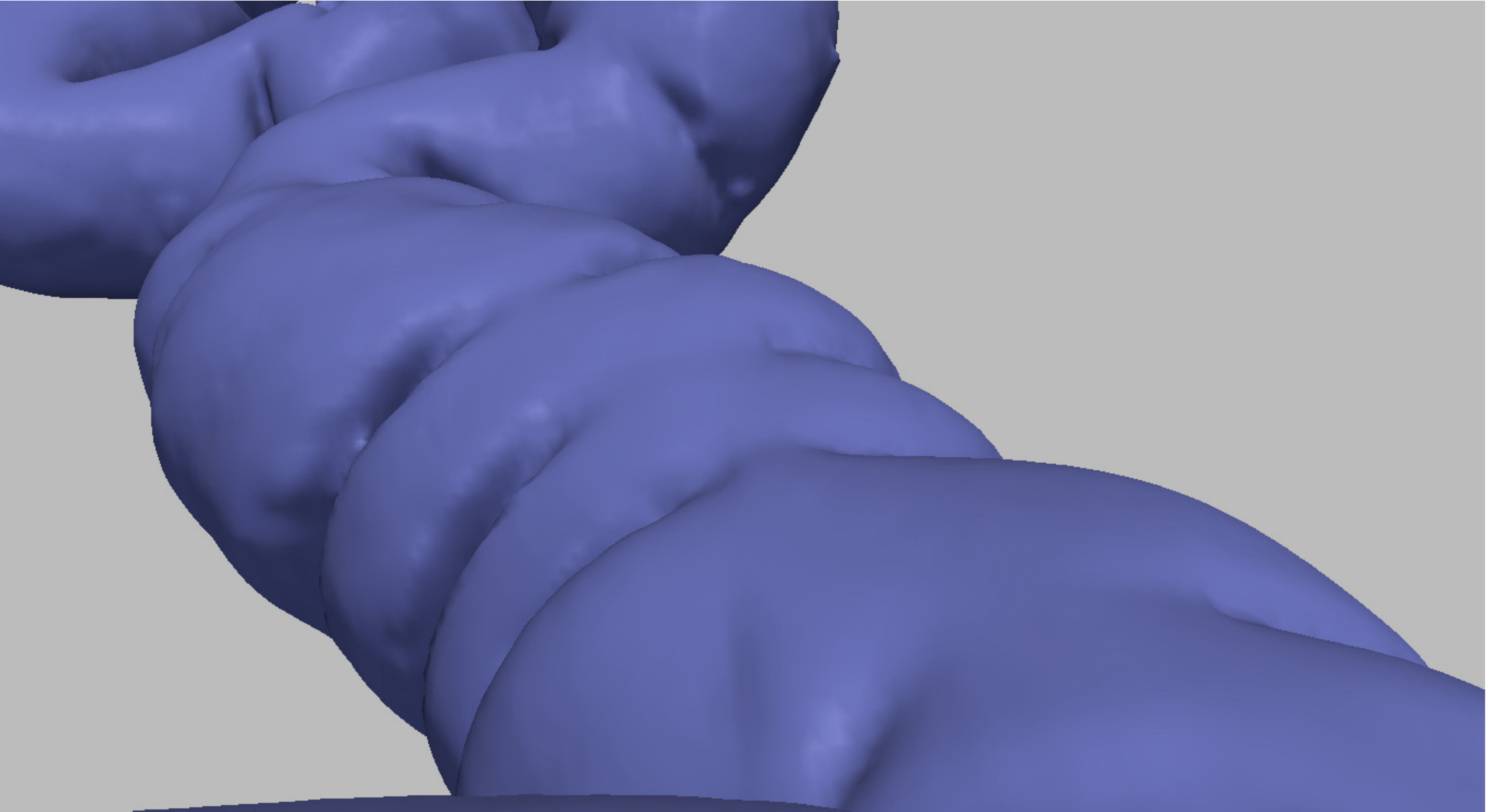}
\includegraphics[width=0.48\columnwidth]{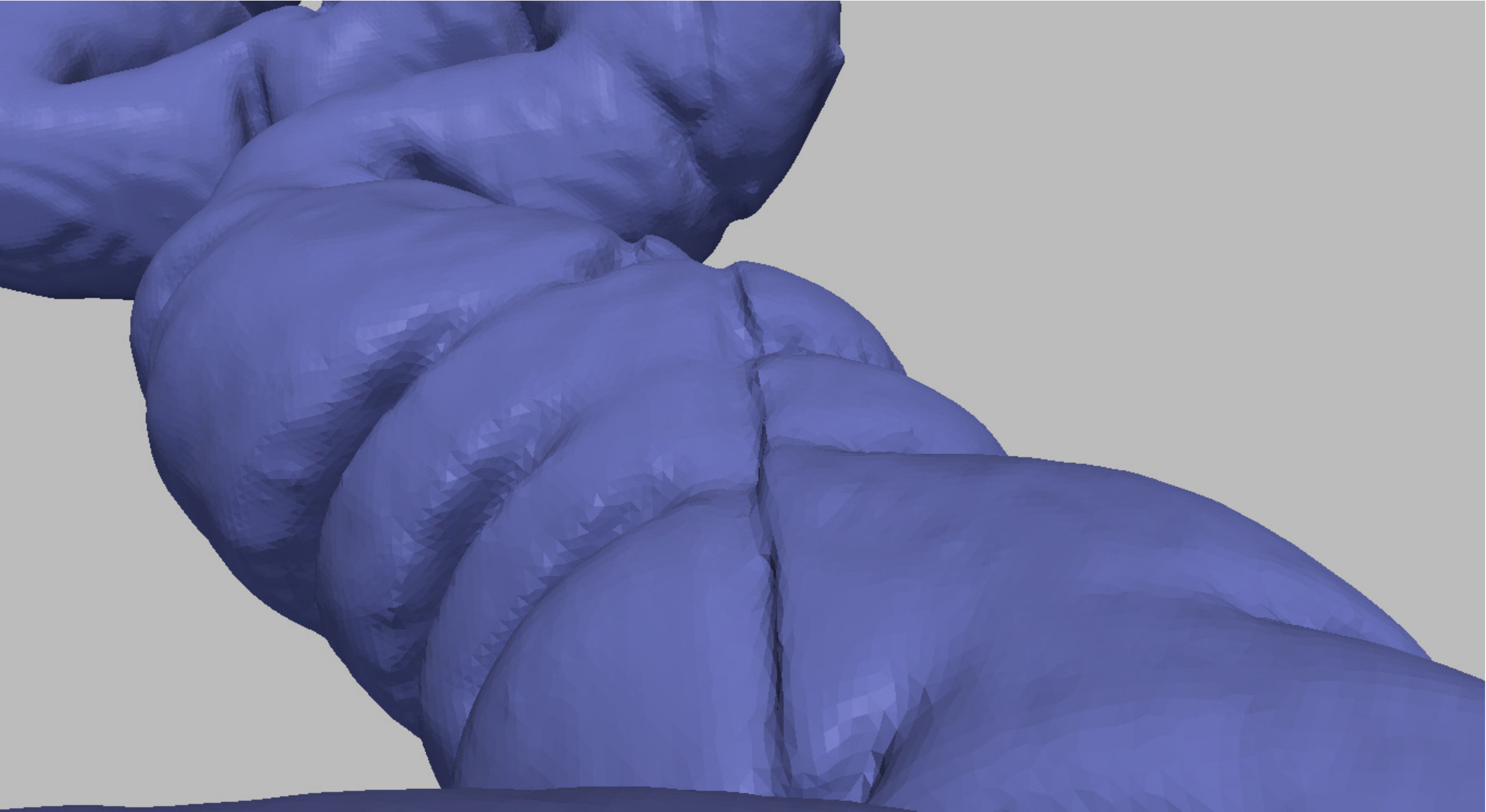}
\caption{Comparison of reduced artifacts in our segmentation (left) with a previously tested more standard version. \label{fig:canaletas}}
\end{center}
\end{figure}

The air-fluid-tissue joint may create artifacts in the segmentation. This is a critical point, not only because of the quality of the segmentation, but mainly for the potential of yielding several false positives in the polyps detection step. 
 It is not rare that segmentation algorithms result in ``gutter-like'' shapes along this interface. This improvement is crucial, since if small oscillations occur along the ``gutter'' (which is expectable due to the resolution of the CT image), artifacts
with polyp-like shape are produced, and this, of course, degrades the performance of the whole CAD system. We paid particular attention to this issue while designing the segmentation algorithm: the $IC$ computation allows to avoid these artifacts. 
Figure \ref{fig:canaletas} illustrates the performance of our segmentation method compared to another version that presents some problems along this interface. The aim of the comparison is not to discuss which segmentation is better, but to show that our algorithm presents a smooth surface along the place where the gutter is expected to be.

\subsection{Smoothing and colon surface computation}

In order to eliminate noise and to obtain a smoother colon surface after the segmentation stage, we proceed to smooth the initial segmentation $u_0$. 
In this section we derive a PDE-driven smoothing technique that preserves the shape of the polyps, while obtaining a smooth enough surface to reliably compute local geometric features such as curvatures. Of course, the ultimate goal of the method is to simplify and to improve the polyp/non-polyp classification system. The effectiveness of the proposed approach will be assessed with experiments both qualitatively and quantitatively in Section \ref{sect:results}, where ROC curves obtained with the proposed PDE and other filtering alternatives will be compared.


We concentrate here on a family of smoothing PDEs of the form
\begin{equation}
\frac{\partial u(\mathbf{x},t)}{\partial t} = \beta |\nabla u| \quad , \quad u(\mathbf{x},0) = u_0(\mathbf{x}) \,\, ,
\label{upde}
\end{equation}
where the initial volume $u_0$ results form the preprocessing technique described in the previous section. After a few iterations of this PDE evolution, the inner colonic wall will be extracted as a suitable iso-level surface of the resulting $3D$ image $u(\mathbf{x},T)$. The choice of the number of iterations and the iso-level are not arbitrary and will be discussed in detail at the end of this section. 

We recall that the Level Set Method \cite{oshersethian} states that if $u(\mathbf{x},t)$ evolves according to \eqref{upde}, then its iso-levels (level sets) satisfy
\begin{equation}
\frac{\partial \S}{\partial t} = \beta \Nm,
\label{spde}
\end{equation}
where $\S$ is any iso-level surface and $\Nm$ its unit normal. This geometric view enables to design $\beta$ to fulfill a set of requirements we will impose to the surface evolution. In particular, we are interested in motions driven by the principal curvatures $\k_{max}$ and $\k_{min}$. 

\begin{figure}[h!]
\begin{center}
\includegraphics[width=0.3\columnwidth]{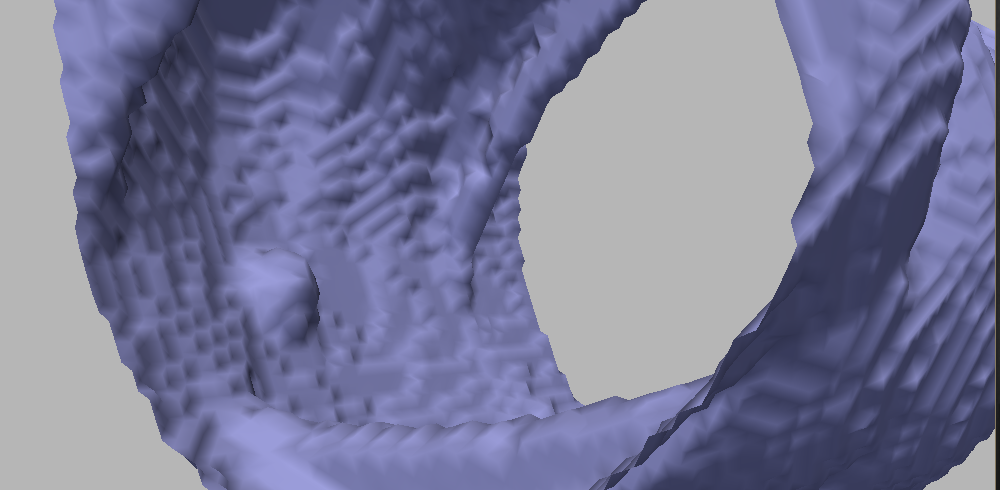}
\includegraphics[width=0.3\columnwidth]{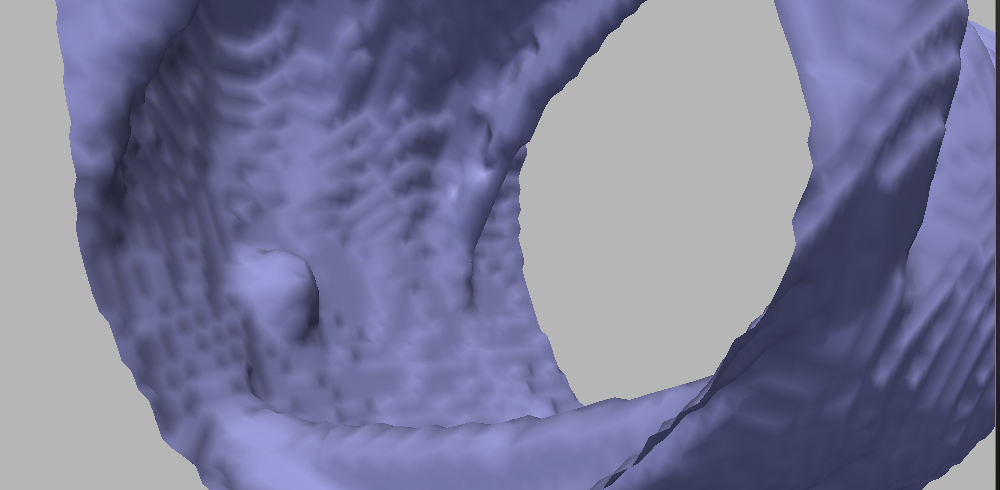}
\includegraphics[width=0.3\columnwidth]{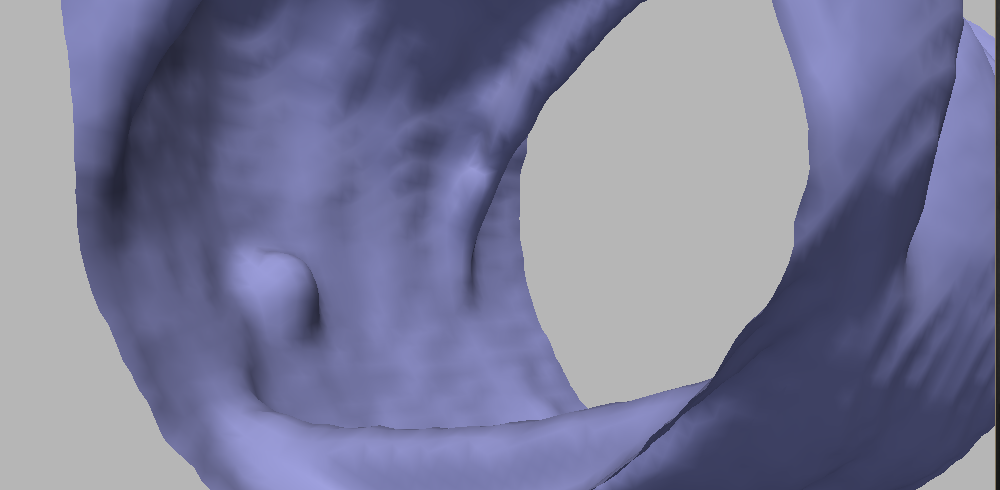}
\includegraphics[width=0.3\columnwidth]{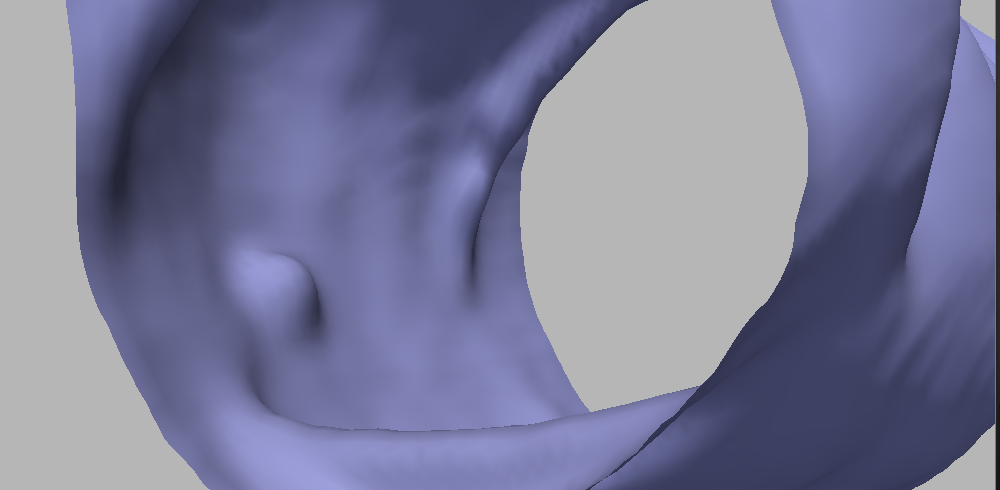}
\includegraphics[width=0.3\columnwidth]{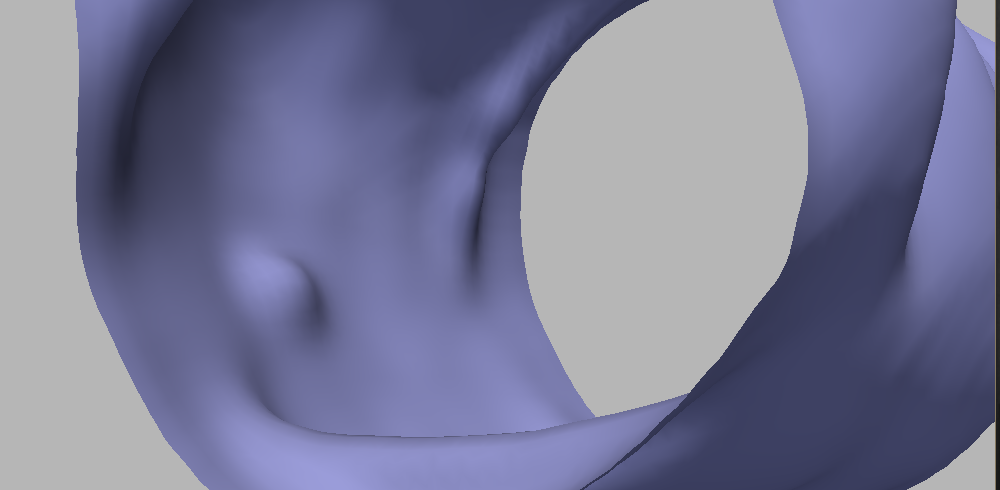}
\includegraphics[width=0.3\columnwidth]{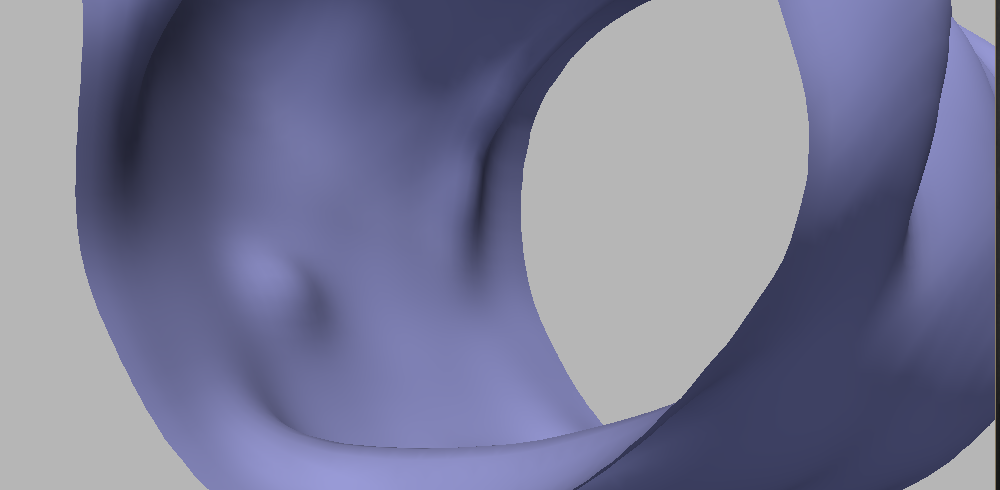}
\caption{Mean Curvature Motion: original surface and the result after $2$, $8$, $15$, $30$ and $50$ iterations. Note how both the surface (as desired) and potential polyps (undesired) are smoothed and flattened.\label{fig:mcmcolon}}
\end{center}
\end{figure}

\red{With the mean curvature motion ($\beta = \mathcal{H}$), the polyps are flattened too fast, as shown in in Figure \ref{fig:mcmcolon}. 
A suitable variation of the motion by Gaussian curvature,\footnote{The motion by Gaussian curvature $\beta = \K$ has several problems with surfaces containing non-convex parts \cite{chopp}.} namely the affine motion
\[ \frac{\partial \S}{\partial t} = (\K^+)^{1/4}\Nm \quad \mathrm{where} \quad \K^+ = \max(\K,0) \,\, ,\]
has a better behavior in general, but the results regarding polyp flattening are comparable with the mean curvature.}

%
%

\begin{figure}[h!]
\begin{center}
\includegraphics[width=0.6\columnwidth]{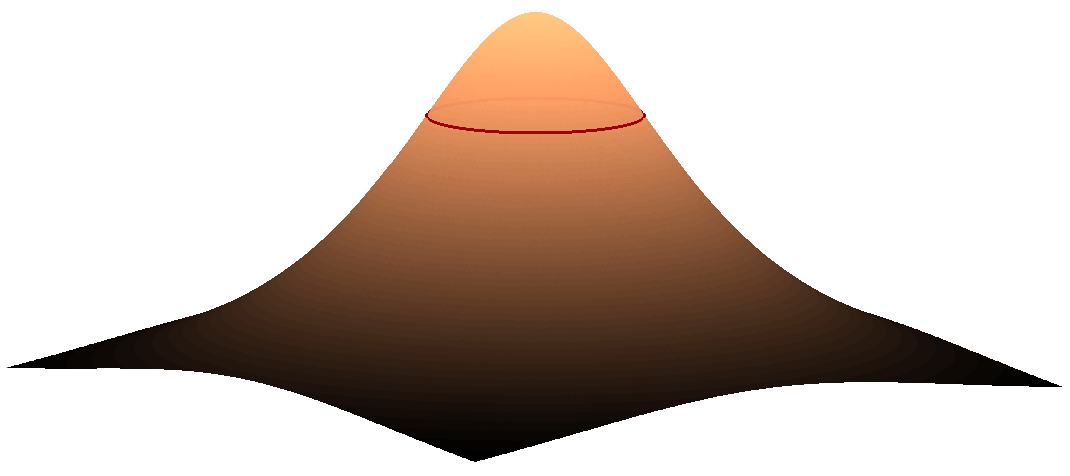}
\caption{Polyp with a curve of inflection points (in red), where $\k_{min}=0$.\label{fig:inflec}}
\end{center}
\end{figure}

A classical motion that appears to be well suited for our problem is the motion by minimal curvature \cite{Caselles1996}. Indeed, polyps have a curve of inflection points all around it, separating its upper and lower sections (see Figure \ref{fig:inflec}). Along this curve, the minimal curvature is $\k_{min}=0$, and therefore this section of the polyp does not move (or moves very slowly), so intuitively under this motion the polyps should persist longer. In our application, this evolution yields very good results in terms of both surface smoothing and polyp enhancement. 

This PDE can be modified to obtain better results in terms of polyp detection. We now propose a set of modifications that lead us to the proposed smoothing evolution equation, and we show qualitative results to support this claim. The improvement in terms of polyp detection performance is discussed in Section \ref{sect:results}. 

\begin{figure}[h!]
\begin{center}
\includegraphics[width=0.3\columnwidth]{0.png}
\includegraphics[width=0.3\columnwidth]{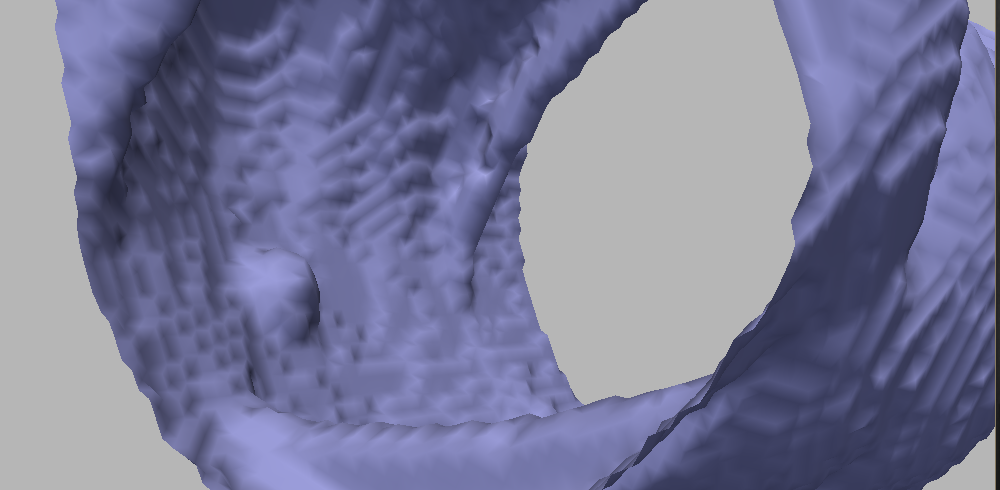}
\includegraphics[width=0.3\columnwidth]{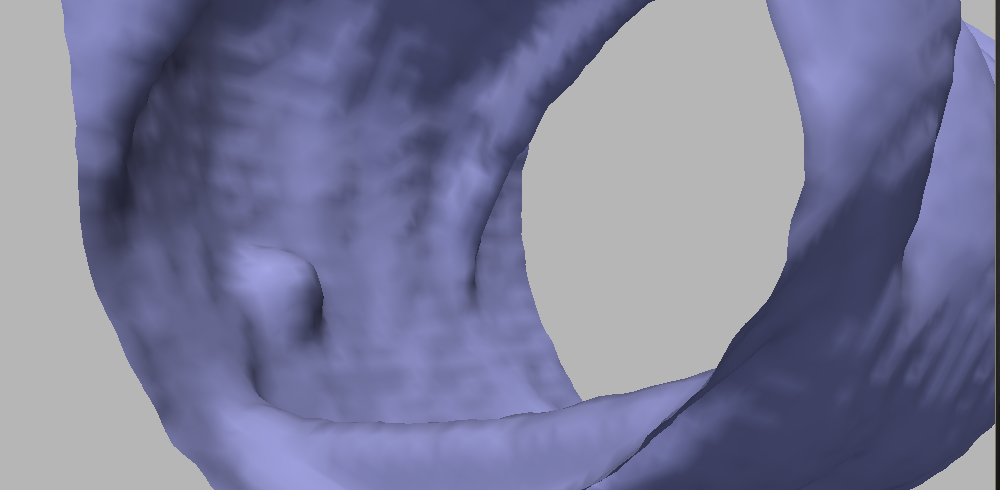}
\includegraphics[width=0.3\columnwidth]{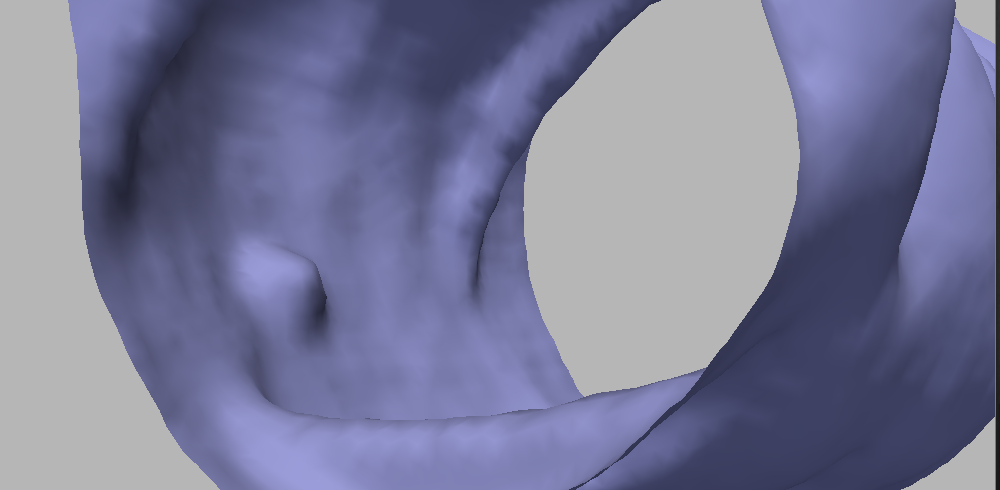}
\includegraphics[width=0.3\columnwidth]{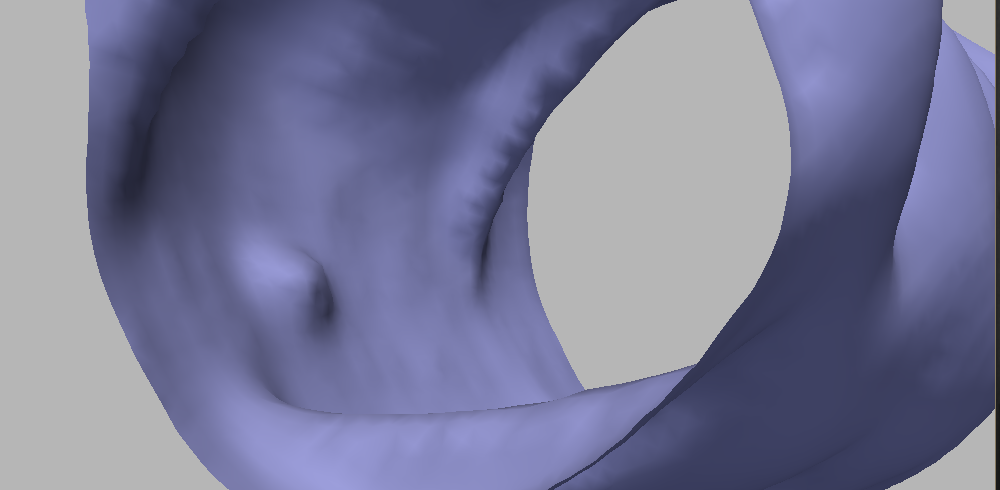}
\includegraphics[width=0.3\columnwidth]{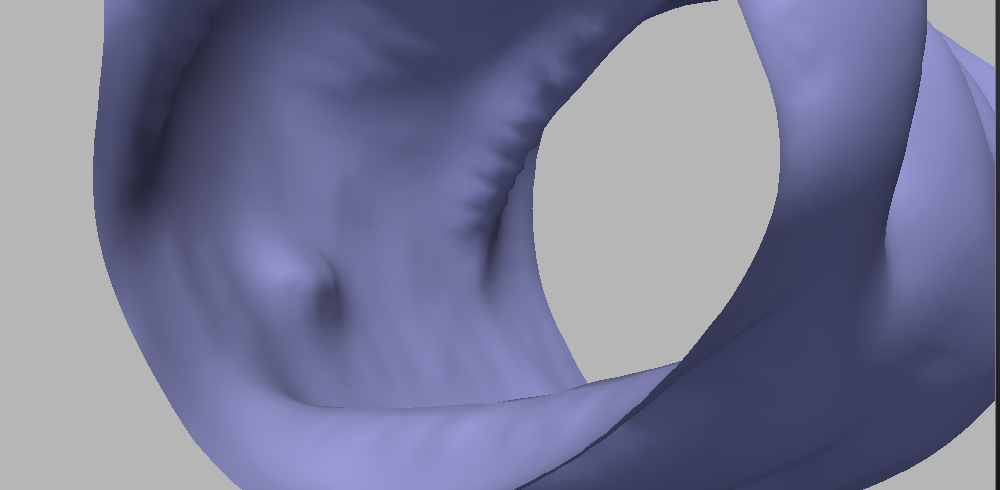}
\caption{Evolution by $\k_{min}^{1/4}$: original surface and the result after $2$, $8$, $15$, $30$ and $50$ iterations.\label{fig:kmin14}}

\end{center}
\end{figure}

\begin{figure}[h!]
\begin{center}
\includegraphics[width=0.9\columnwidth]{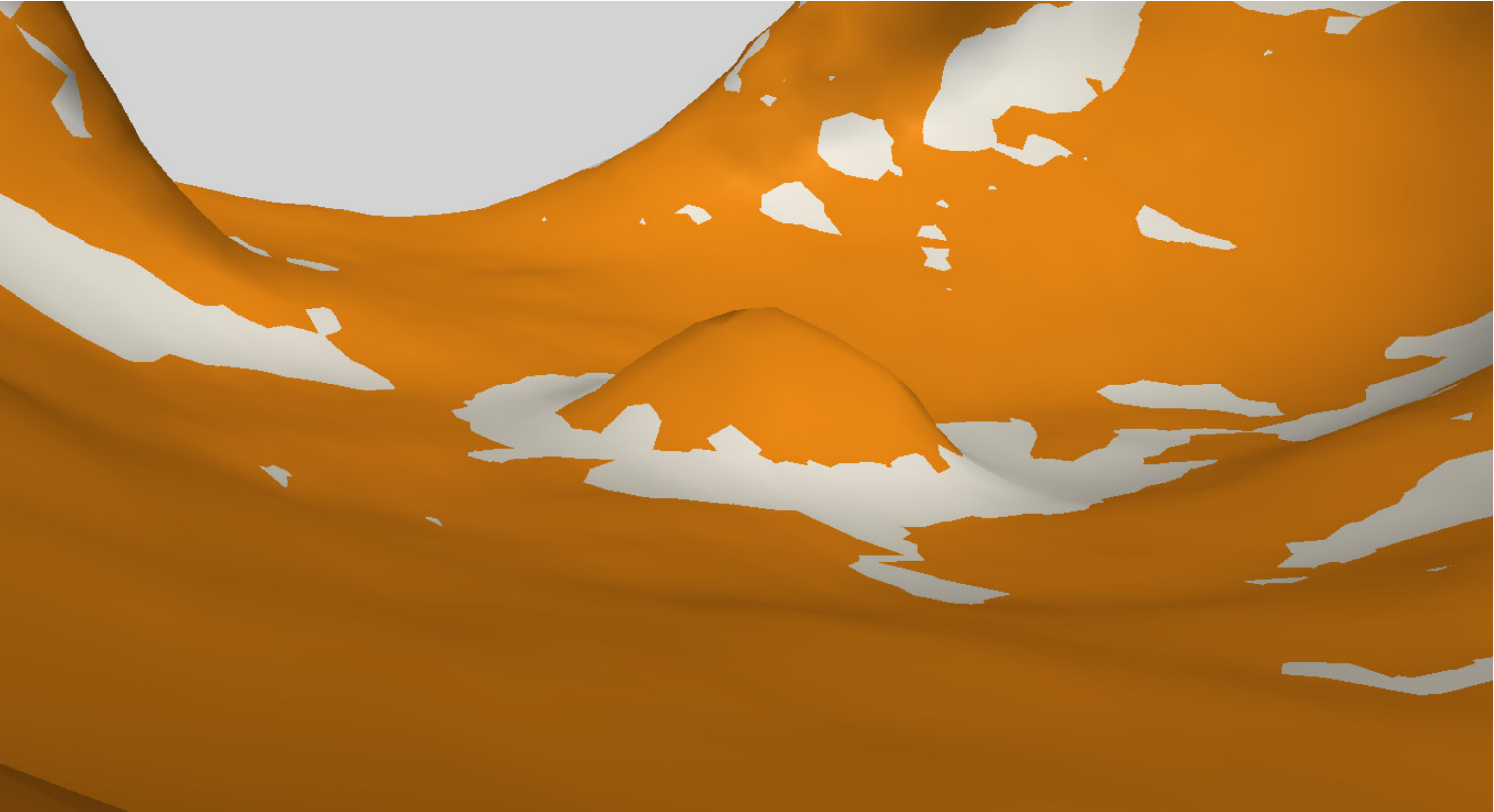}
\caption{Comparison between evolutions. Motion by $k_{min}$ in light gray vs. motion by $k_{min}^{1/4}$ in dark gray. \red{Both surfaces are overlaid, so sections that are not visible are hidden below the other surface.}\label{fig:kmin14vskmin}}
\end{center}
\end{figure}

The first modification is inspired by the exponent $1/4$ of the affine motions in dimension $3$, and yields to the following curvature motion equation:
 \[ \frac{\partial \S}{\partial t} = \k_{min}^{1/4}\Nm \,\, .\]
%
Figure~\ref{fig:kmin14} shows the result after a few iterations; comparing to Figure~\ref{fig:mcmcolon}, it can be readily seen that this motion achieves a better trade-off in terms of noise reduction and polyp preservation. Figure \ref{fig:kmin14vskmin} evidences the difference with a comparative image: the result of the motions by $\k_{min}$ and $\k_{min}^{1/4}$ are shown in gray and in orange, respectively. On the polyp protrusion, the orange surface is above the gray surface, while the opposite is observed in the surrounding area. This shows that the evolution by $\k_{min}^{1/4}$ leads to better polyp enhancement. 

The second modification that we introduce is based on the idea of preserving the polyps qualities that we later use to identify them. Towards this aim, we take into account a surface property that will be used in the feature extraction stage as well. A measure of the local shape of a surface is the so-called \textit{shape index} \cite{koenderink},
$$\si := -\frac{2}{\pi}\arctan \left( \frac{\k_{max}+\k_{min}}{\k_{max}-\k_{min}}\right).$$
A complementary measure called \textit{curvedness} $C$, is defined as
$$R := \sqrt{\frac{\k_{max}^2+\k_{min}^2}{2}} \quad , \quad C := \frac{2}{\pi}\ln R \,\, .$$

\begin{figure}[h!]
\begin{center}
\includegraphics[width=0.75\columnwidth]{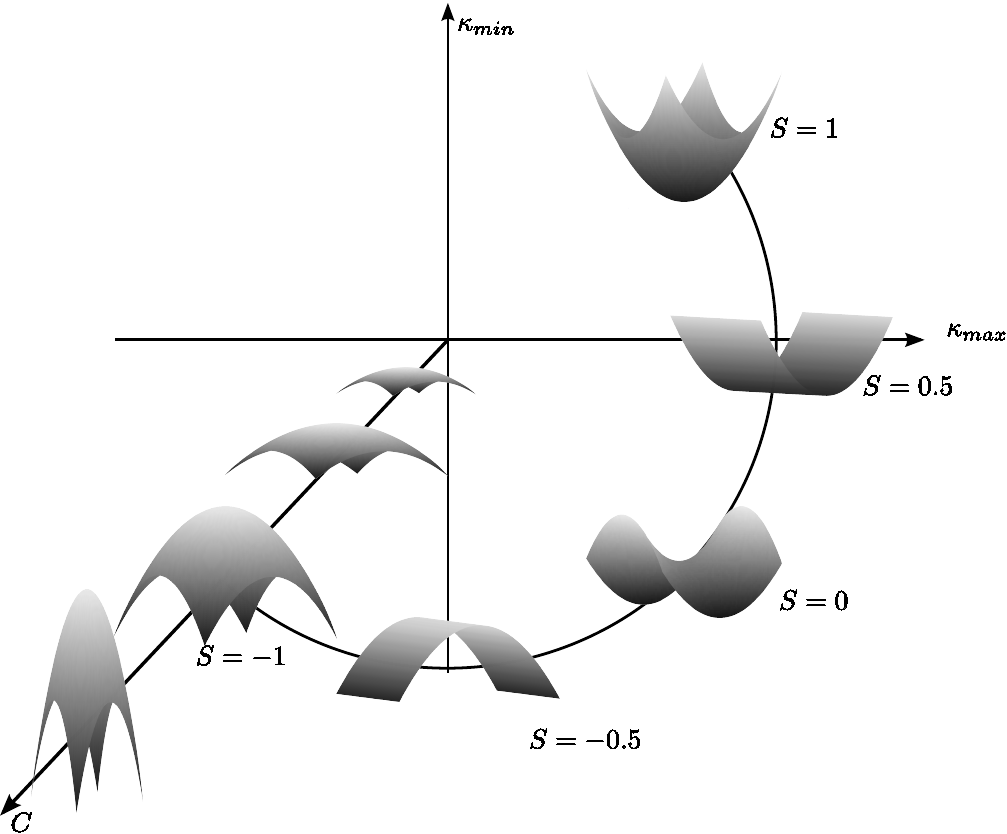}
\caption{Some shapes and their corresponding shape index values.\label{shape1}}
\end{center}
\end{figure}

The $\left( \k_{max},\k_{min}\right)$ plane is then transformed into the $(\si,C)$ plane. While the value of $\si$ is scale-invariant and measures the local shape of the surface, the value of $C$ indicates how pronounced it is.
Figure \ref{shape1} shows different shapes and their corresponding shape index. Due to the chosen orientation, shape index values close to $-1$ (protrusions) are of special interest for polyp detection.

Special care must be taken when computing the shape index. Principal curvatures are usually derived from $\H$ and $\K$, however, as shown by \cite{Monga}, this may lead to inaccuracies that can be strongly amplified when computing the shape index. A much more stable estimation can be obtained by computing $\k_{min}$ and $\k_{max}$ directly from $u$ (see \cite{Monga}).

Back to the PDE motion, the next step is to include this information concerning the shape of the surface in order to make potential polyps evolve differently than the rest of the colon surface.
More precisely, we modify the best motion so far ($\beta = \k_{min}^{1/4}$), in such a way that the resulting motion further enhances the potential polyps. In
order to achieve this, we first need to characterize the potentially polyp points, and then modify the deformation function accordingly.

\begin{figure}[h!]
\begin{center}
\includegraphics[width=0.95\columnwidth]{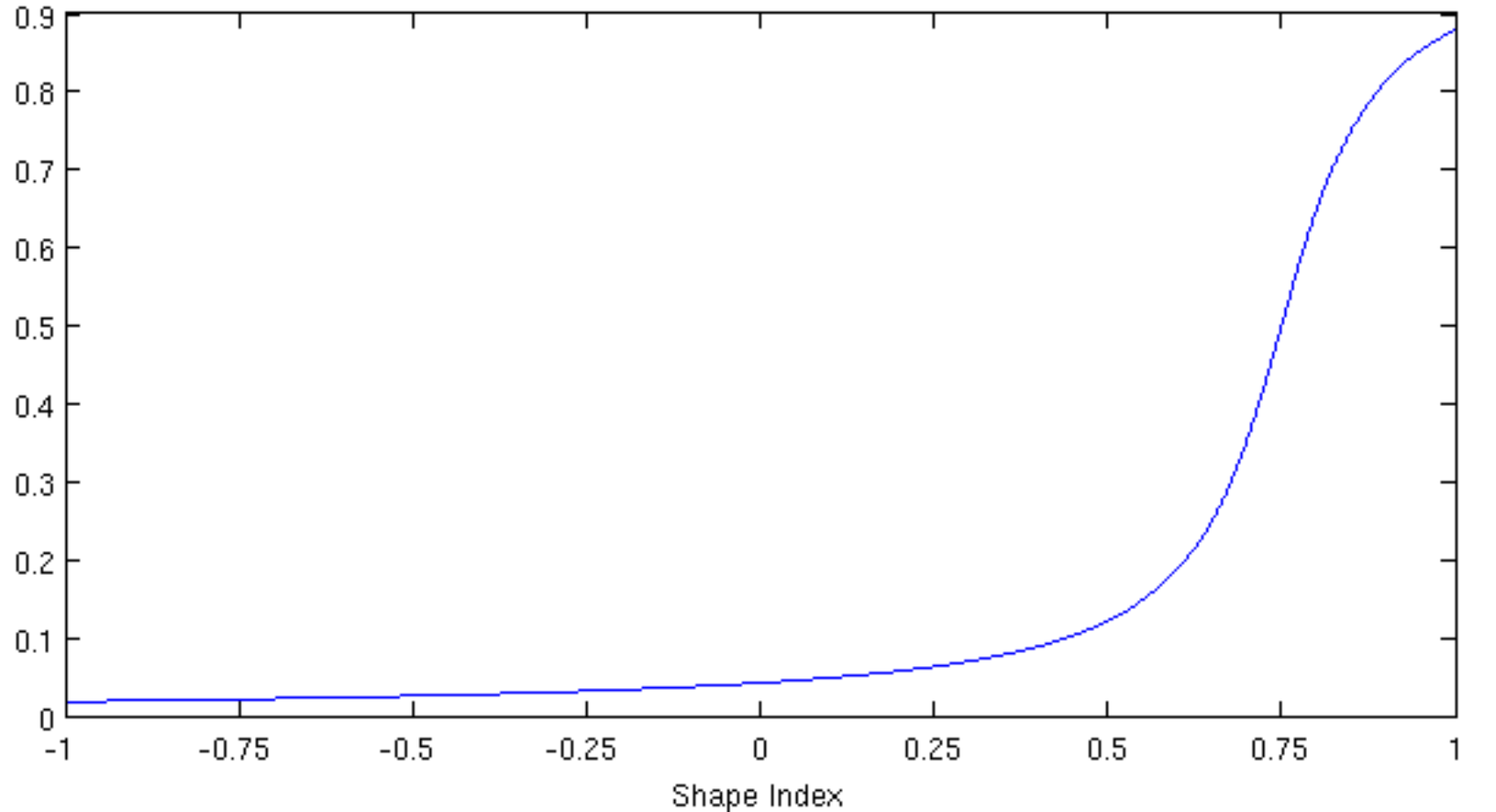}
\caption{Function $g(\si) = \frac{1}{\pi}\arctan \left((\si-0.75)\cdot 10\right) +\frac{1}{2}$, multiplying factor for PDE curvature evolution. \label{fig:fc}}
\end{center}
\end{figure}

 We define a function that acts as a multiplying factor to the term $\k_{min}^{1/4}$,
making the surface evolve slower at the interest points. One option is to choose this function to depend on the shape index only, assigning low values to shape index near
$-1$, and values close to unity to other points. A smooth function $g(\si)$ verifying these constraints is shown in Figure \ref{fig:fc}.

The final motion then becomes
\[ \frac{\partial \S}{\partial t} = g(\si) \, \k_{min}^{1/4}\Nm \,\, .\]

This proposed evolution keeps all the advantages of the motion by $\k_{min}^{1/4}$ and in addition, the polyps are flattened slower, so at the end the obtained
surface is smooth and the polyps are still outstanding.

As discussed in Section \ref{review}, the aim of the work by \cite{konukoglu2007} is to enhance polyps, although they present their work as a preprocessing stage and not a complete CAD system. However, there are several differences with our approach. We use a function of both minimum curvature and shape index, while they use (modified) mean curvature; we use the same evolution for the whole surface, while they use a different function for each polyp candidate; our algorithm smooths the whole surface (needed in order to compute curvatures properly), where their approach is thought to modify the polyp regions only. As our evolution algorithm is part of the segmentation step, while theirs is a stage between segmentation and classification, it is not clear how to quantitatively compare the performances. In any case, as our proposed PDE is part of the segmentation, these two techniques are complementary.


The number of iterations can be set by choosing the value that maximizes the overall performance of the system, measured in terms of the free-response ROC curve (FROC), so the number of iterations is set to obtain the best FROC curve. This can be done by trying with several values and keeping the one which maximize this performance, using cross validation over the training dataset. Alternatively, we can consider a sphere of the size of the CT resolution and compute analytically the number of iterations that are needed to make it vanish (see Appendix \ref{apendice}). The idea behind this procedure is to smooth the surface up to the resolution limit. These two approaches led to the same result, namely $15$ iterations, and therefore this is the chosen value for the experiments in this paper.

\begin{figure}[h!]
\begin{center}
\includegraphics[width=0.8\columnwidth]{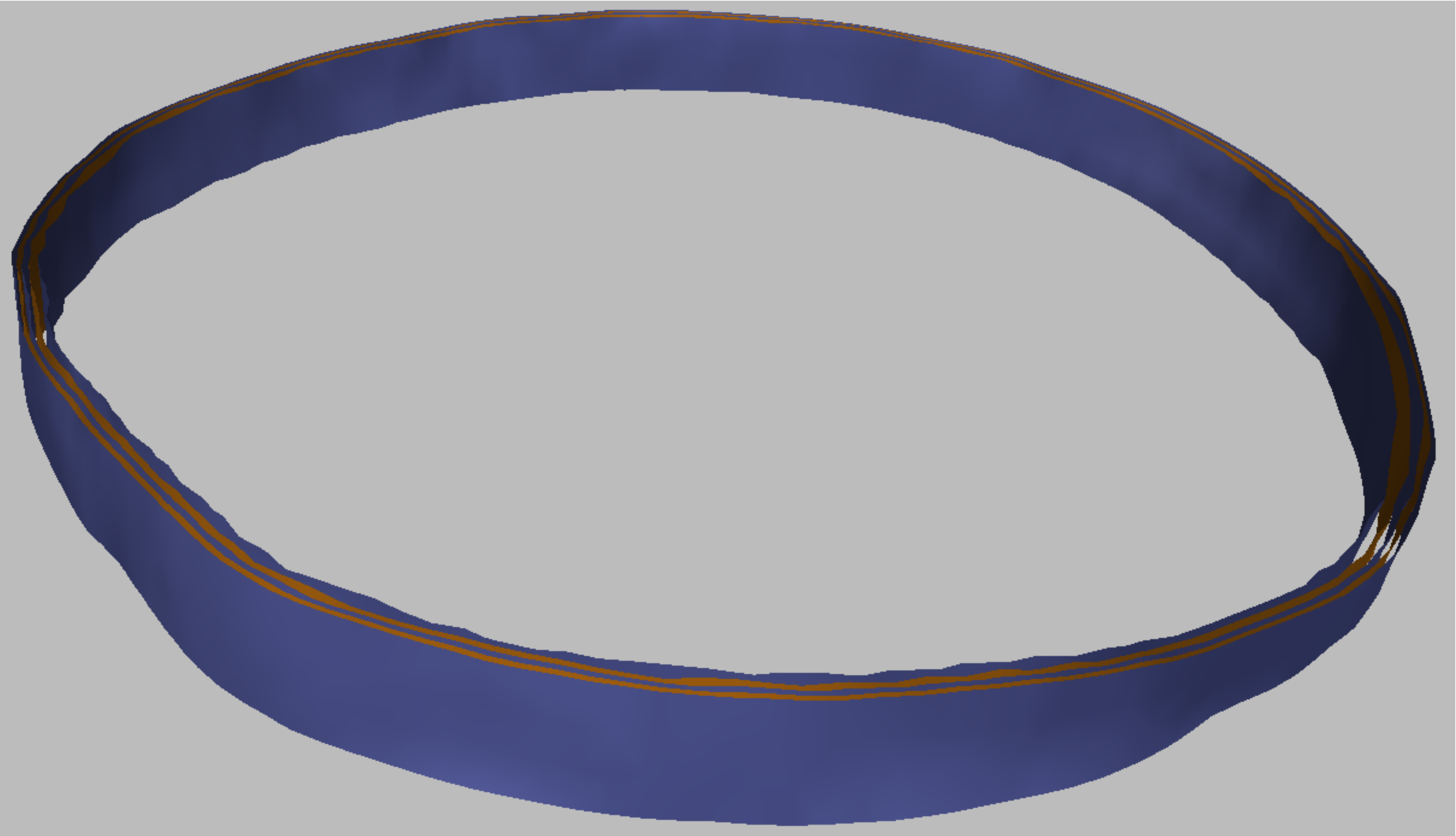}
\caption{Different iso-level surfaces (a thin section of the colon wall): $\alpha$ values $0.5$, $0.6$, $0.7$, $0.8$, and $0.9$. Recall that the whole variation range for $\alpha$ is $(0,1)$. Note that, although the $\alpha$ values are very different, the obtained iso-level surfaces are very close to each other.\label{fig:capas}}
\end{center}
\end{figure}

\begin{figure}[h!]
\begin{center}
\includegraphics[width=0.4\columnwidth]{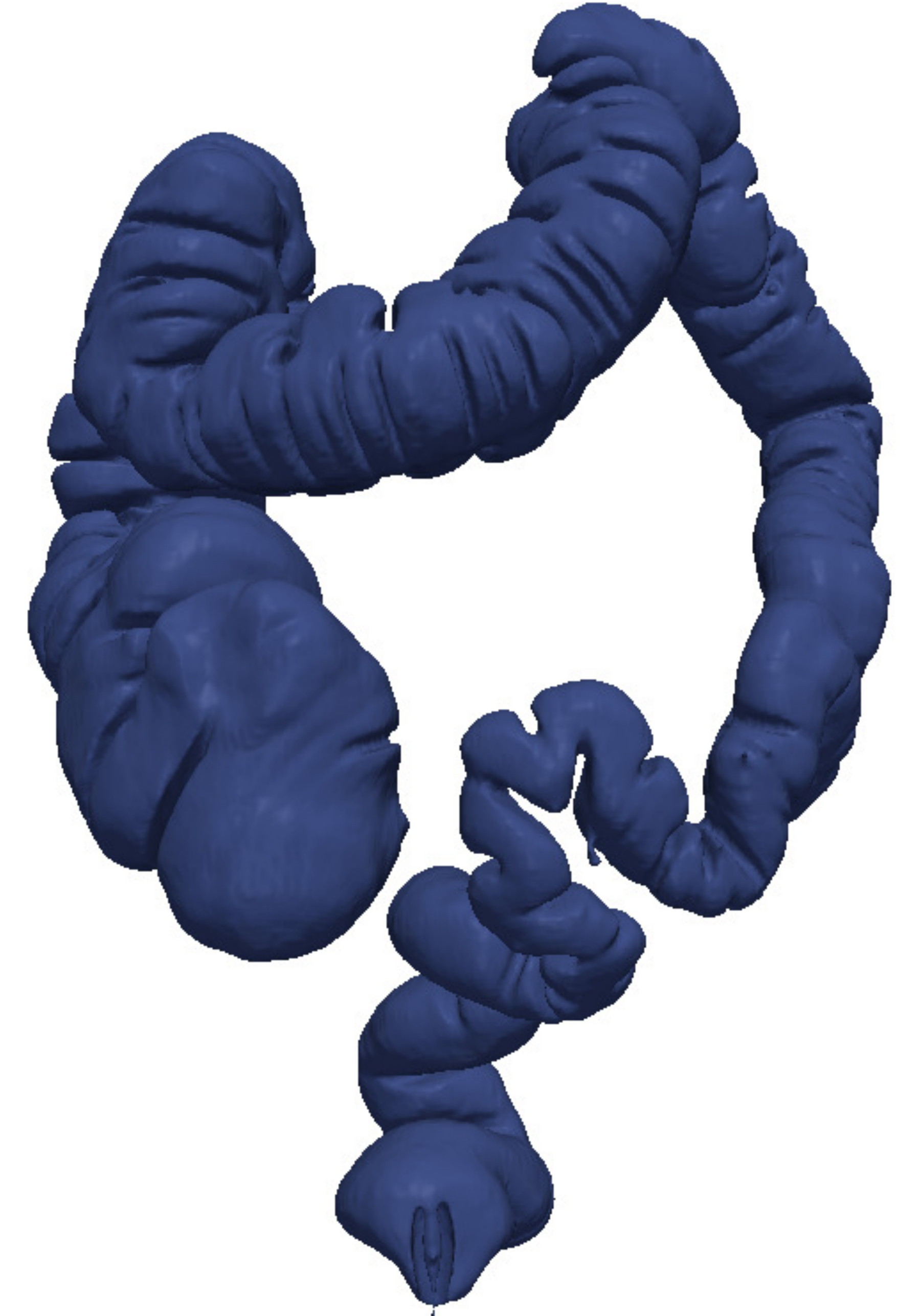}
\caption{Example of a segmented colon. \label{fig:segm3d}}
\end{center}
\end{figure}

At this point, after choosing the appropriate diffusion and the number of iterations, we have a smoothed volume $u(\mathbf{x},T)$ indicating the volume inside of the colon.
We then extract the surface of the colon, using the marching cubes algorithm \cite{marching}, obtaining the iso-value surface of level $\alpha\in [0,1]$. The choice of the value $\alpha$ can be made by maximizing some criteria, in order to obtain the most contrasted surface in a given sense
 \cite{enric08}. This optimization-oriented method was tested, and we observed that in our particular application all the consistent surfaces are very close to each other (see Figure \ref{fig:capas}), and all of them are reasonable segmentations of the colon.
 Therefore, the computational effort is not justified 
 and we simply kept the iso-level surface $\alpha=0.7$. 
Note that this choice can be safely made once for all the data. 
The result of this stage is then a triangulated surface $\S$ representing the colon wall, Figure \ref{fig:segm3d}.


Due to the lack of ground truth, we simply evaluated the segmentation visually and as a component of the polyp detection pipeline (for which we do have ground truth). Visual inspection indicated that the segmentation only missed two very small parts in the entire dataset, and no extra regions (like small bowel) were included. We estimate that only $<1\%$ of all the database colon surfaces was missed. We have also observed that earlier versions of our segmentation algorithm, e.g., with different velocity functions, do negatively affect the polyps detection. We did not test our algorithm with another bowel preparation due to the lack of such data, and this is an important subject of future research.

\section{Polyp Delineation and Feature Extraction}
\label{features}

For each case study, we consider as input both the raw CT images and the segmentation of the colon volume as obtained following the procedure described in the previous section. 
The first stage of the proposed polyps flagging algorithm consists in detecting surface patches that are candidates of being polyps.
The complete set of connected points that constitutes the candidate patch is found by progressively growing the patch and keeping the one that maximizes the geometric dissimilarity with the surrounding area, in the sense of the features presented below (the starting point for this growth is also detailed below). The detection system is therefore based on differential (non-absolute) features, thereby better adapting to intrinsic variabilities both of the colon region and of the potential polyps, as further explained next.

All the polyp detection methods reported in the literature try to detect or classify the polyps from properties defined only within the candidate region, without considering
the data surrounding the region (of course, the fact that the polyp is raised respect to the surrounding area is always used, but no features measured explicitly taking into account context information were proposed yet). However, it is important to analyze the context in which the candidate patch is located, not only because different sections of the
colon present different characteristics, but also because polyps can be situated over different structures such as folds or plain colonic wall. A good feature
including the shape of the neighborhood for example, can help in the discrimination between irregular folds and polyps over folds. In addition, looking for significant differences in the gray level imitates the human-based inspection, which
highlights zones that contrast with their vicinity.

In this regard, most of the features described in this section take into account the local information of the area surrounding the candidate patch. Polyps (actually
all the candidate patches) are then characterized not only by their intrinsic geometry and structure, but also by their relationship with the surrounding area.
This makes the features more robust to the particular local phenomena, in a context where the natural variations of
the properties of the colon tissue impact the measures and make absolutes thresholds or decision rules impractical. 
The normal tissue of different cases may vary (due to different biological properties of the subjects or to different conditions of the studies), so absolute thresholds in texture features lack meaning; while texture patterns differ from study to study, what does not vary is the fact that polyps have different properties than normal tissue.


\subsection{Candidate detection and geometrical features}

\begin{figure}[h!]
\begin{center}
\includegraphics[width=0.7\columnwidth]{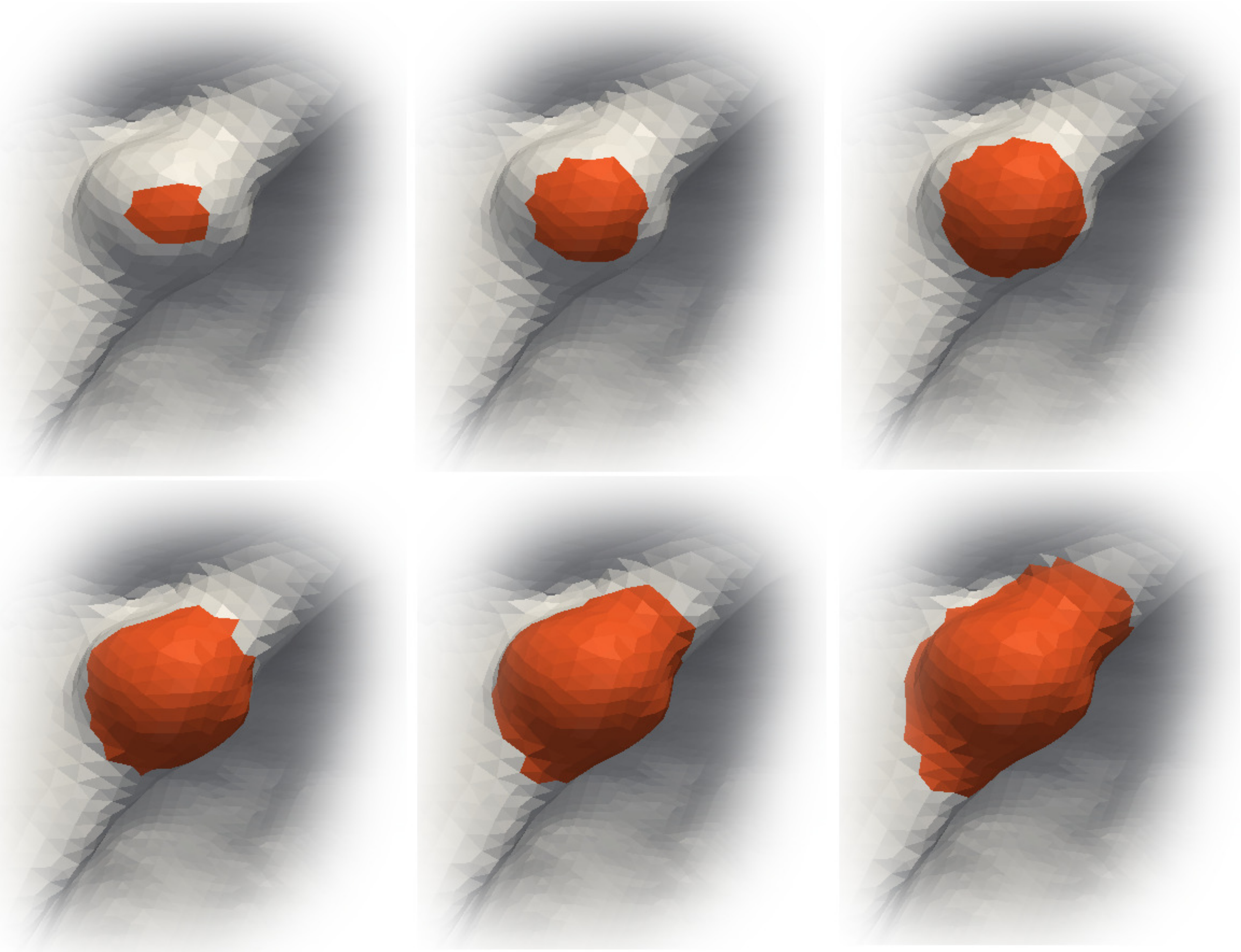}
\caption{Sets $\P_1\ldots \P_n$: different sizes are tested in order to select the most appropriate patch. $\P_1\ldots \P_n$. \label{fig:patchgrow}}
\end{center}
\end{figure}

The starting point for the geometric features described in this section is the segmented surface $\S$. 
Let us consider the shape index as a function $\si:\S \to [-1,1]$, and recall that the polyps have shape index values close to $-1$. Therefore, it is expected that a region (patch) of the surface that corresponds to a polyp contains at least one local minimum of the shape index function. The detection of the candidate patches starts from this observation, and follows an adaptive-scale search.
For each local minimum $x_0 \in \S$ of the function $\si$, several level sets of $\si$ ($\P_1\ldots \P_n$) around $x_0$ are tested, and the level set $\P_i$ that maximizes the distances between the histograms described below, is the considered candidate patch, which we simply denote by $\P$, Figure \ref{fig:patchgrow}. A total of $n=7$ level sets are tested, corresponding to the shape index values from $-0.8$ to $-0.5$ with a $0.05$ step. 
The following description is given for the final chosen patch $\P$, but the ring and histogram computations are made for all the level sets $\P_i$ in order to select the most appropriate of them (the most appropriate scale).

\begin{figure}[h!]
\begin{center}
\includegraphics[width=0.3\textwidth]{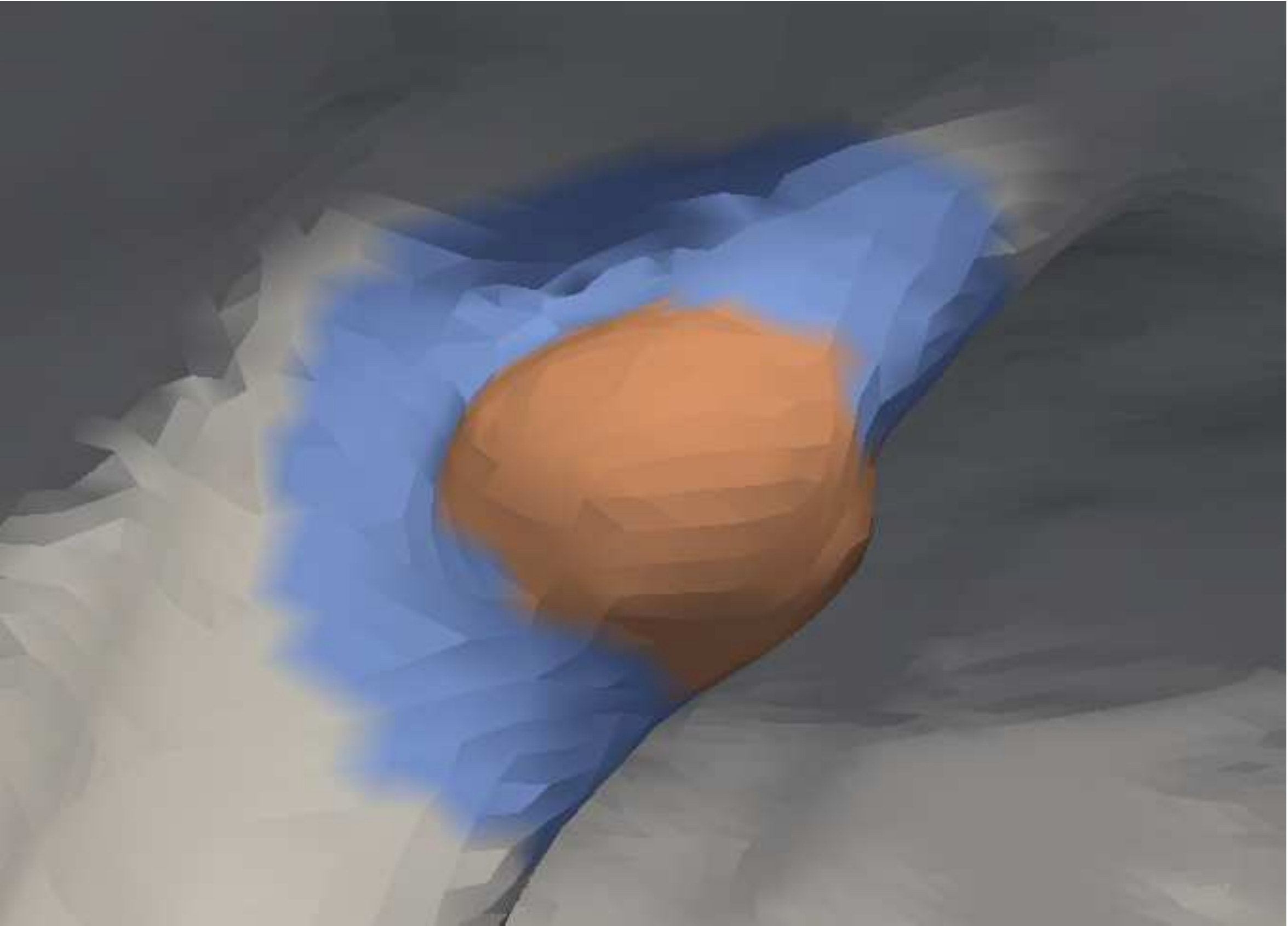}
\caption{Ring (in blue) surrounding a candidate polyp (in orange).\label{ring}}
\vspace{-0.4cm}
\end{center}
\end{figure}

Given a candidate patch $\P$, a ring $\Ri$ around $\P$ is computed, in order to consider geometrical measurements with respect to the area surrounding the patch. The ring is calculated by dilating the patch $\P$ a certain geodesic distance, such that the areas of $\P$ and $\Ri$ are equal. The geodesic distance computation is performed using the algorithm by \cite{computinggeodesic}. Figure \ref{ring} shows a candidate patch (actually a true polyp), and its corresponding ring.

Histograms of the shape index values are then computed for the patch $\P$ and the ring $\Ri$, and two different distances between them are computed: the $L_1$ distance \red{(defined as $\sum_i{|x_i-y_i|}$, where $x_i$ and $y_i$ are the histogram values)} and the symmetric Kullback-Leibler divergence \cite{cover}. If the patch corresponds to a polyp-like shape then the values of the histogram $\P$ will be
concentrated around the $-1$ extrema, on the other hand, the histogram $\Ri$ will be inclined to the other extreme in case of a polyp on a normal colon wall
(concave), or with tendency to values near $-0.5$ if the polyp is on a fold. These two features give a measure of the geometric local variation of the
candidate patch $\P$.  We assume that there are no other polyps in $\Ri$ or that they do not significantly affect the statistics on the ring.


Although these two distances are the most discriminative features\footnote{\red{Several feature selection techniques confirmed this. For instance, the following methods available in \textit{Weka} \cite{weka}: \textit{Information Gain, Gain Ratio, and Relief-F.} Using the data described below, these techniques sort the features according to their discriminative power.}}, we also consider the following additional ones since they still help to discriminate some typical false positives:
\begin{itemize}
\item The mean value of the shape index over the patch $\P$, which
describes the shape of the selected patch.
\item The area of the patch, since we want to detect polyps in a certain
range of size. 
\item The growth rate of the areas at the adaptive-size stage, meaning the ratio between the area of the chosen patch $\P=\P_i$ and the area of the immediately smaller tested patch $\P_{i-1}$; this feature measures how fast the shape of the patch is changing, in a context where it is difficult to quantize the variation of the shape. 
\item And finally the \textit{shape factor},
 $$ SF = \frac{4\pi \cdot
Area}{Perimeter^2} \quad , $$ which
measures the shape of the patch border, how efficiently the perimeter is used in order to gain area,\footnote{The maximum value for the
shape factor is $1$ and it is achieved only by the circle.} and it favors circle-like patches (like the polyp patch in Figure \ref{ring}), avoiding
elongated patches (like the false positives in folds).
\end{itemize}
We then end-up with a total of $6$ geometric features for detecting candidate polyps, namely: $L_1$ and Kullback-Leibler distance between shape index histograms
of patches and corresponding rings, the mean shape index over $\P$, the area of the candidate patch, the growth rate of the areas and the shape factor.


\subsection{Texture features}

Due to the differences in biological activity of polyp cells, the gray-level of the CT image and its texture can be very helpful for detecting polyps. This is in particular useful for flat or small polyps, where the geometric information is limited. 
Some work has been done on the inclusion of texture features (inside the candidate polyps only), in order to reduce false positives \cite{wang2005_medphys}. According to the reported results, there is a lot of room for improvement in texture features. \cite{Nappi2002} also propose a texture feature, but it relies on structures located at the interface between the polyp and the interior of the colon (tagged fluid or air). It does not make explicit use of the information of the normal tissue surrounding the polyp. 
We propose both the use of new texture features and the inclusion of the information on the candidate's surrounding area.

First, for each polyp candidate $\P \subset \S$, a volume $V_1$ is calculated, containing the patch $\P$ and a portion of the inner tissue next to the patch. The volume $V_1$ is obtained by dilating (in 3D) the patch $\P$ towards the inner colon tissue (we discard the air or fluid voxels). A second volume $V_2$ surrounding $V_1$ is calculated by dilating $V_1$. Volume $V_2$ is intended to contain normal tissue in order to compare it with the polyp candidate tissue. \red{In order to choose how much dilation is performed, we use a technique similar to the one in the previous section: several dilation distances are tested, and we keep the distance that makes the differential features most discriminative.}


The features chosen are a subset of the classical Haralick texture features \cite{haralick}, namely, entropy, energy, contrast, sumMean, and homogeneity. Seven co-occurrence matrices (considering seven directions in $\R^3$, $(1,0,0),(0,1,0),(0,0,1),(1,0,1),(1,1,0),(0,1,1)$ and $(1,1,1)$) are calculated with the voxels of $V_1$, and all the five features are averaged over the seven directions. The analogous computation is made for $V_2$, and the differences between the two volumes, for each texture feature, is considered.
Additionally, the mean gray levels of the voxels in both volumes is computed, and their difference is considered as a feature. In this way, six texture features are considered. The rest of the Haralick features were discarded because the overall performance was not as good as the performance with this particular subset. \red{We have confirmed this choice with several feature selection techniques, with the same procedure described above.}
This approach for computing the texture features, measuring differences with the surrounding area, leads to better discrimination than the features computed just for $V_1$, as demonstrated next.

\section{Classification}
\label{sect:classif}

After the candidates detection with the adaptive-scale approach, the number of true polyps was much lower than the number of non-polyps patches, a relation on the order of $500$:$1$, which is a significant problem for the learning stage of the classifier. Three techniques were considered to overcome this shortcoming.


The MetaCost approach \cite{metacost} consists of combining several instances of the classifier instead of stratification (modify the proportion of classes in the training data according to the costs). 
This method does not work with ``stable'' classifiers (those that produce similar models with slightly different training sets) like SVM or Naive Bayes.

The Cost Sensitive Learning approach, unlike the MetaCost, tries to balance the classes before the learning stage. The implementation we used from \textit{Weka} \cite{weka} simply takes as input the re-balance parameters (cost matrix) and replicates instances of the minority class.
One of the advantages of this approach is that no assumptions are made about the behavior of the classifiers (unlike the previous method) nor the distribution of the data (unlike the next method).


Finally, the Synthetic Minority Over-sampling TEchnique (SMOTE) is a method to generate artificial instances of the minority class, in order to get a balanced data to learn from. The new artificial instances are created as a convex combination of the existing instances of the minority class. Therefore, there is an underlying assumption that the optimal partition in the feature space gives convex sets, which may not be the case in several applications.

We tested all these options and the best results were obtained using Cost Sensitive Learning, as expected from the comments above.


The numerical results listed below were obtained by classifying with SVM using Cost Sensitive Learning, after normalizing the data; Na\"{i}ve Bayes performed similarly.

\section{Results}
\label{sect:results}

\begin{figure}[h!]
\begin{center}
\includegraphics[width=0.9\columnwidth]{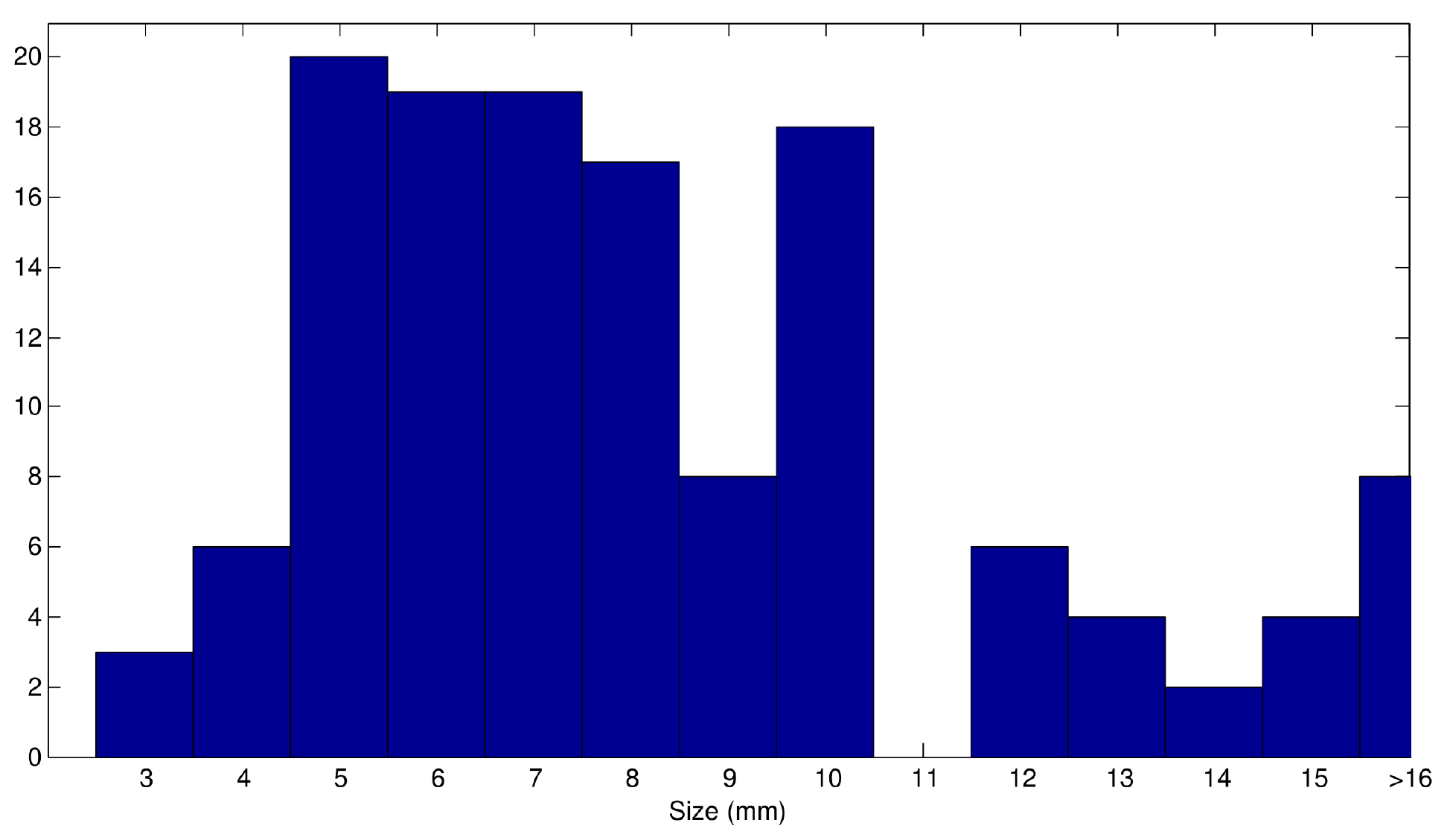}
\caption{Histogram of the polyps' sizes.\label{fig:histo_size}}
\vspace{-0.4cm}
\end{center}
\end{figure}

\red{A total of $150$ patients of the Walter Reed Army Medical Center (WRAMC) database \cite{paper_wramc} were used to test the proposed CAD algorithm\footnote{Data provided courtesy of Dr. Richard Choi, Virtual Colonoscopy Center, Walter Reed Army Medical Center.}. Most of these patients have two sets of CT images, one for supine and one for prone position. Taking precautions not to train the classifier with, for instance, the prone images set and test it with the supine set (i.e. one cannot use a supine study of a given patient for training, an the corresponding prone study for testing, or vice versa), we can consider the $300$ images sets as independent. From now on, we refer to each of these $300$ images sets as a \textit{case}. The database contains $134$ polyps detected by optical colonoscopy, including $12$ flat polyps.  Among these $134$ polyps, $86$ are larger than $6mm$ in size, and the other $48$ are between $3mm$ and $6mm$ in size. Figure \ref{fig:histo_size} shows the distribution of polyps' sizes in the database. The size and shape classification of these polyps was taken from the WRAMC database description. These descriptions were provided by the physicians involved in the OC examinations. Taking these examinations as the ground truth, patches classified as polyp were considered TP if the distance to a ground truth polyp was less than $3mm$.}

\begin{figure}[h!]
\begin{center}
\begin{footnotesize}
\def\svgwidth{220pt}
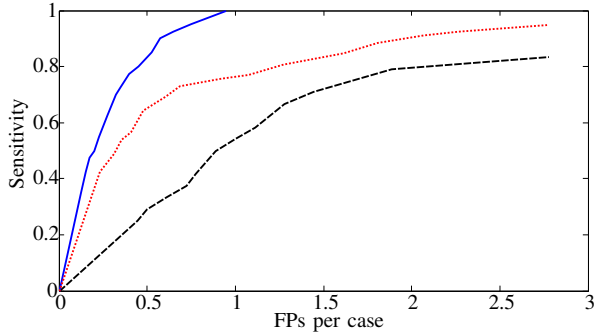
\end{footnotesize}
\caption{\red{FROC curve of the proposed system for different polyps sizes: larger than $6mm$ (solid), smaller than $6mm$ (dashed), and all polyps (dotted).}\label{fig:froc}}
\vspace{-0.4cm}
\end{center}
\end{figure}

\red{The evaluation was carried out by splitting the dataset into two halves, training and testing.}
\red{Under this setting, we obtained the FROC in Figure \ref{fig:froc}, which shows the performance for different polyps sizes. About $40\%$ of the polyps were covered by tagging, but the classification results do not vary depending on this fact, the performance is the same for covered and for non-covered polyps.}

These values are comparable with the state-of-the-art results \cite{wang2005_medphys,suzuki}, but our database includes very small polyps. 
A more precise comparison of results is 
not necessarily meaningful, since in general each work considers its own database.  

\begin{figure}[h!]
\begin{center}
\begin{scriptsize}
\def\svgwidth{110pt}
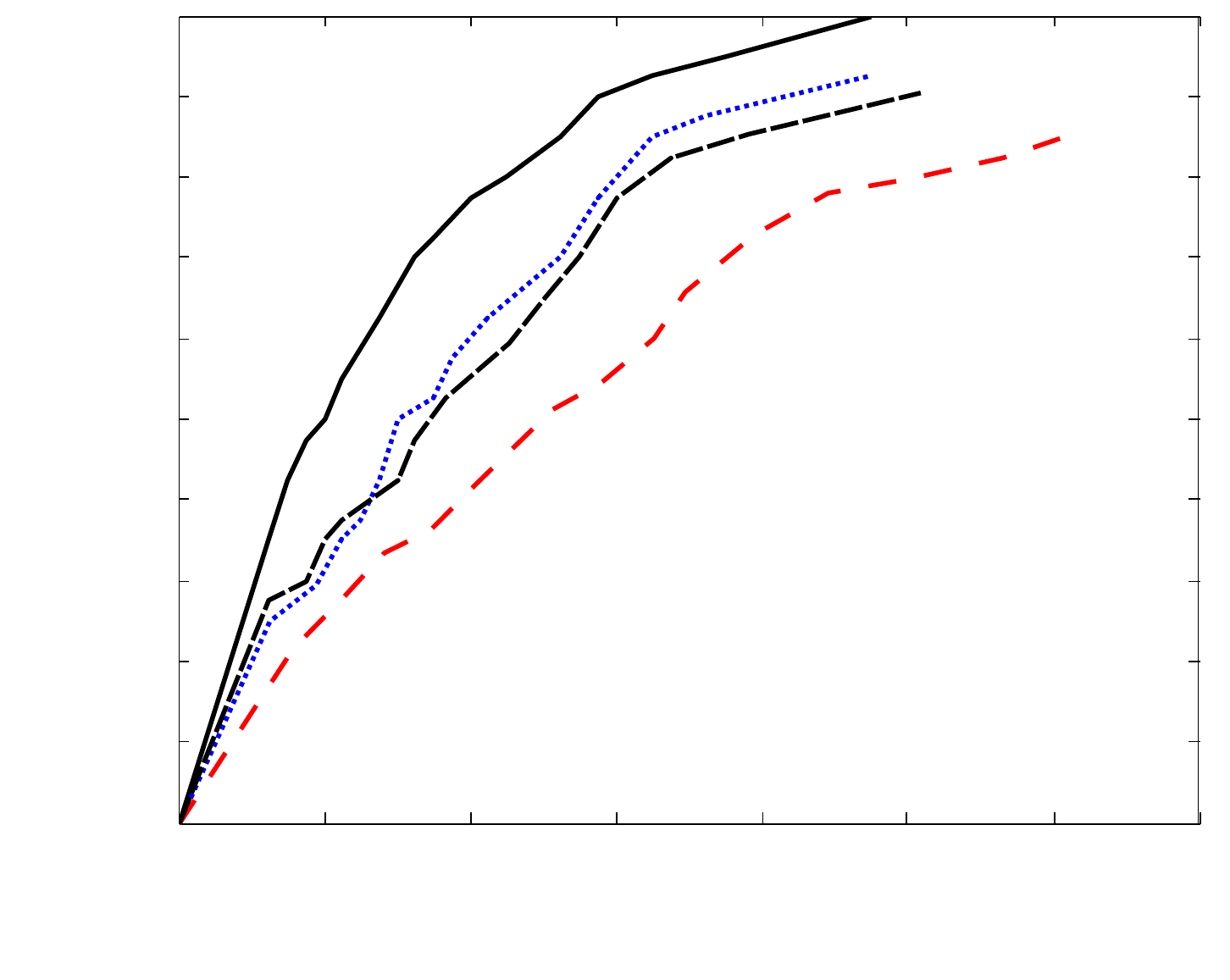
\def\svgwidth{110pt}
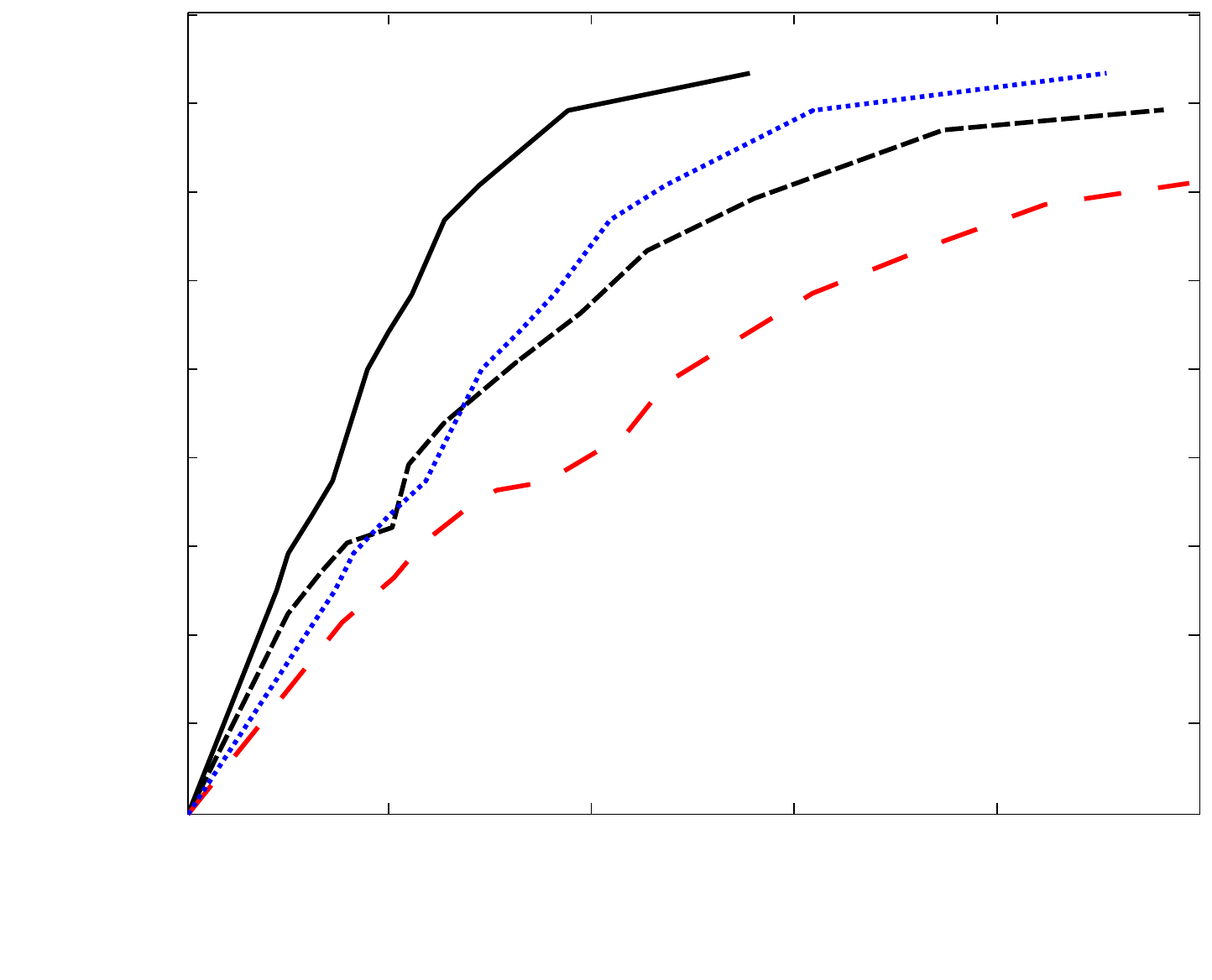
\end{scriptsize}
\caption{\red{FROC curve comparing the performances using the different smoothing methods, classifying large polyps (left) and small polyps (right). The curve for the proposed evolution is shown in solid line, the results for the evolution by $\H$ and $\k_{min}$ are shown in dotted and dashed lines respectively, and the lower curve is the result when no smoothing is performed.}\label{fig:roc_segm}}
\end{center}
\end{figure}

The FROC curve in Figure \ref{fig:roc_segm} compares the performance of the system with the different smoothing methods discussed in Section \ref{segment}. The proposed smoothing technique achieves better results than the other discussed methods.

\begin{table}[h!]
\begin{center}
\begin{tabular}{|c|c|c|}
\hline
 & \multicolumn{2}{c|}{Texture features}\\
& Absolute & Differential\\
\hline
Sensitivity & $96\%$ & $100\%$  \\
\hline
FP per case & $3.1$ & $2.2$  \\
\hline
\end{tabular}
\end{center}
\caption{Comparison of absolute and differential texture features, with polyps larger than $3mm$ in size.}
\vspace{-0.4cm}
\label{tablaresult}
\end{table}

\begin{table}[h!]
\begin{center}
\begin{tabular}{|c|c|c|}
\hline
 & \multicolumn{2}{c|}{Polyp sizes}\\
& $>3mm$ & $>6mm$\\
\hline
Sensitivity & $93\%$ & $100\%$  \\
\hline
FP per case & $2.8$ & $0.9$  \\
\hline
\end{tabular}
\end{center}
\caption{Comparison of performance by polyp size.}
\vspace{-0.4cm}
\label{tablaresult6mm}
\end{table}

\begin{figure}[h!]
\begin{center}
\begin{scriptsize}
\def\svgwidth{110pt}
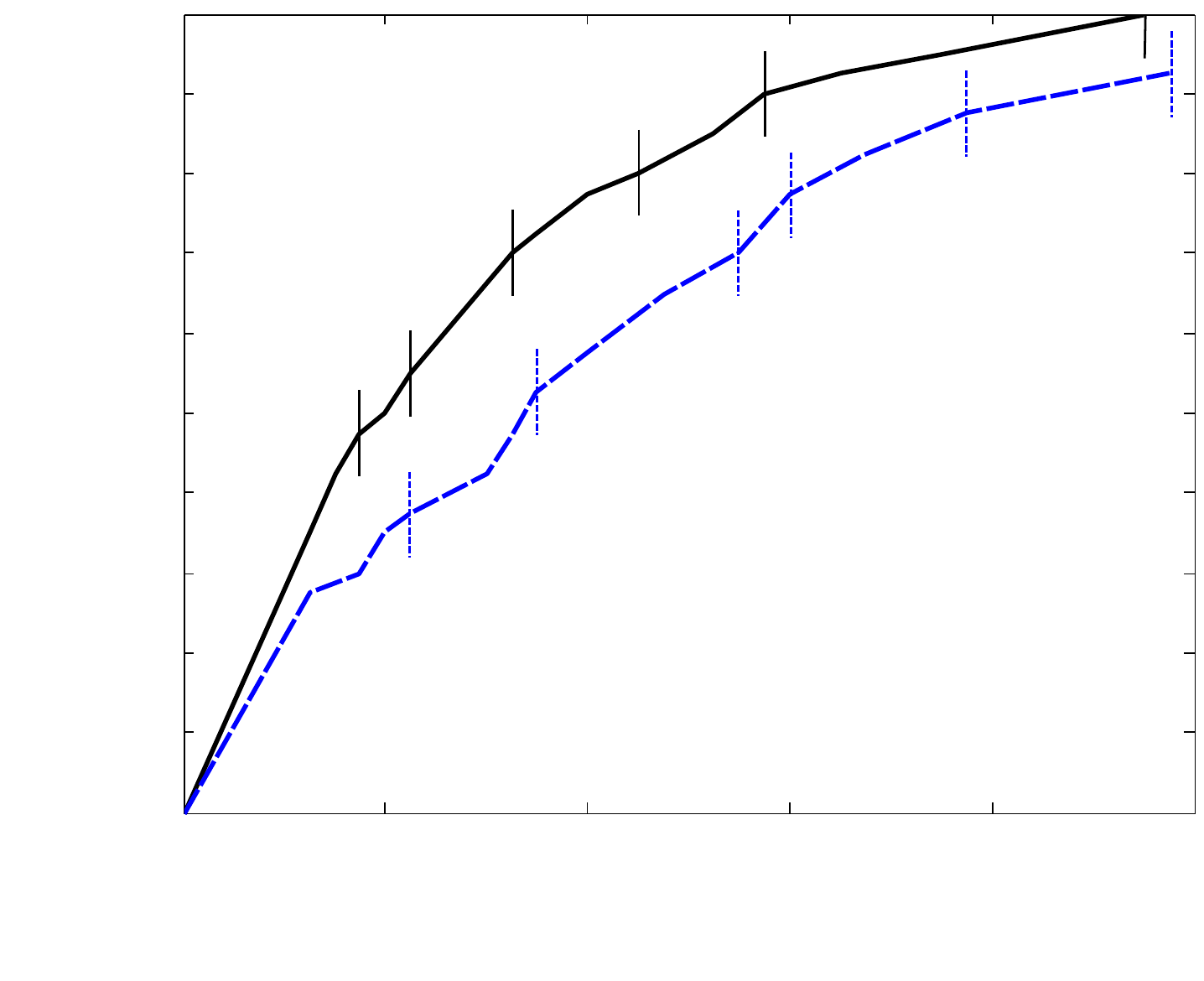
\def\svgwidth{110pt}
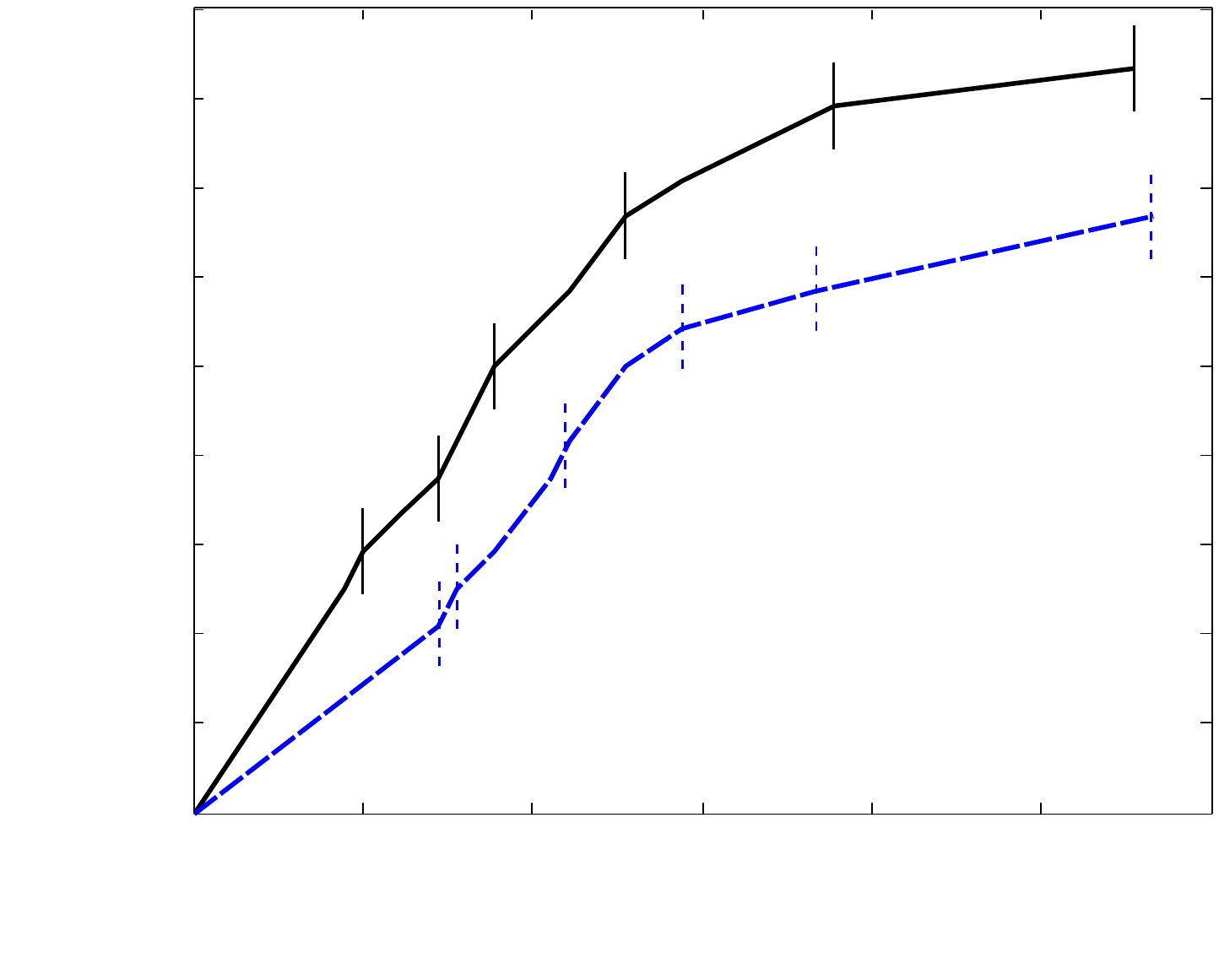
\end{scriptsize}
\caption{\red{FROC curve with $95\%$ confidence intervals, comparing the performances with differential (solid) and absolute (dashed) texture features, classifying polyps larger than $6mm$ in size (left) and smaller than $6mm$ in size (right). Confidence intervals computed according to \cite{macskassy}.}\label{fig:roc_diff}}
\end{center}
\end{figure}

Table \ref{tablaresult} shows the comparison between absolute and differential texture features. The classification was performed using all the geometric features and either the absolute texture features (computed just for $V_1$), or the differential texture features, using the standard leave-one-patient-out strategy. 
The results show that, when combined with the differential geometric features, differential texture features are more discriminative than the absolute ones. The FROC curve in Figure \ref{fig:roc_diff} extends the results in Table \ref{tablaresult} and compares the performance of the classifier when using differential or absolute features.
 Table \ref{tablaresult6mm} compares the classification results according to the polyps size. Again, the work with such small, as well as flat polyps, is unique to the framework here presented, as will be discussed in Section \ref{disc_small}.

\begin{figure}[h!]
\begin{center}
\begin{scriptsize}
\def\svgwidth{240pt}
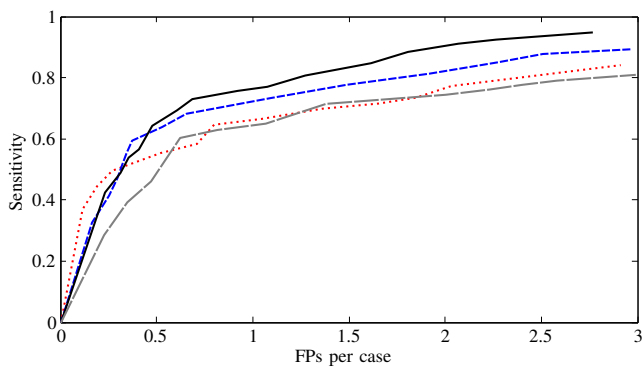
\end{scriptsize}
\caption{\red{FROC curve comparing the performances of different classification approaches. SVM with Cost Sensitive (solid), SVM with SMOTE (dashed), C4.5 trees with AdaBoost (dotted) and plain SVM (long-dashed). } \label{fig:roc_cost}}
\end{center}
\end{figure}

Finally the FROC curve in Figure \ref{fig:roc_cost} compares the results of the different classification approaches. Cost Sensitive, SMOTE, and MetaCost were used as a pre-processing stage for SVM, AdaBoost was used with C4.5 trees. Parameters in all classifiers were optimized via cross validation. Other tested classifiers were not included due to their lower performance.

\vspace{-0.2cm}

\section{Discussion}
\label{discussion}

\subsection{Small, big and flat polyps}
\label{disc_small}

It is clear that both the small and flat polyps are much more difficult to detect than the other polyps. What is not clear is if the same kind of algorithm and features are suitable for detecting all the range of polyp types and sizes. 
We showed that the proposed combination of features, although it may not be optimal for every specific type of lesion, is able to correctly detect all of them.

All the stages in the pipeline, specially the segmentation and the features, contribute to the good classification results for the whole database. 
However, it would be interesting to study which pre-processing techniques and features are better for each type and size of polyp, and to possibly propose different CAD systems for each class of polyp. Nevertheless, the $93\%$ sensitivity together with the $2.8$ FP rate for polyps larger than $3mm$ in size is as remarkable as the $0.9$ FP rate for polyps beyond $6mm$ in size with $100\%$ detection.


\subsection{Geometric and texture importance}

Although geometrical features are the most discriminative ones (see Table \ref{table:geomtext}), texture features still play a fundamental role in the classification. Adding the texture features to the geometric ones, the sensitivity reaches $93\%$, and at the same time the false positives rate decreases by $30\%$.

\begin{table}[h!]
\begin{center}
\begin{tabular}{|c|c|c|c|c|}
\hline
 & \multicolumn{4}{c|}{Features}\\
& All & Geometric & Texture & Excl. $L_1$ \& KL\\
\hline
Sensitivity & $93\%$ & $88\%$ & $68\%$  & $83\%$\\
\hline
FPs p/case & $2.8$ & $6.5$ & $19$  & $12.8$\\
\hline
\end{tabular}
\end{center}
\caption{\red{Comparison of performance using only geometric vs only texture features, and excluding the histogram $L_1$ and KL distances, for polyps larger than $3mm$.} \label{table:geomtext}}
\end{table}

\begin{figure}[h!]
\begin{center}
\includegraphics[width=0.18\textwidth]{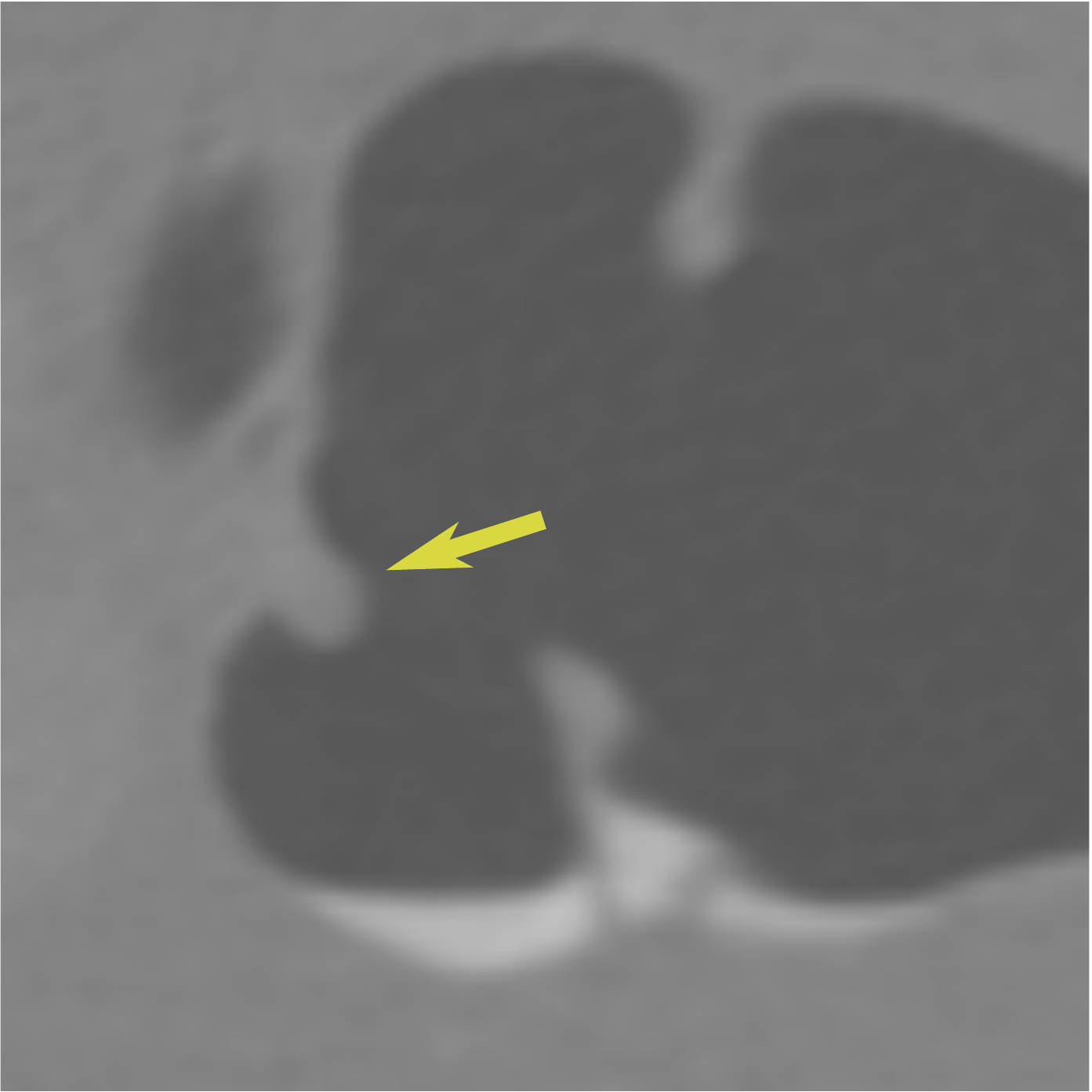}
\includegraphics[width=0.2\textwidth]{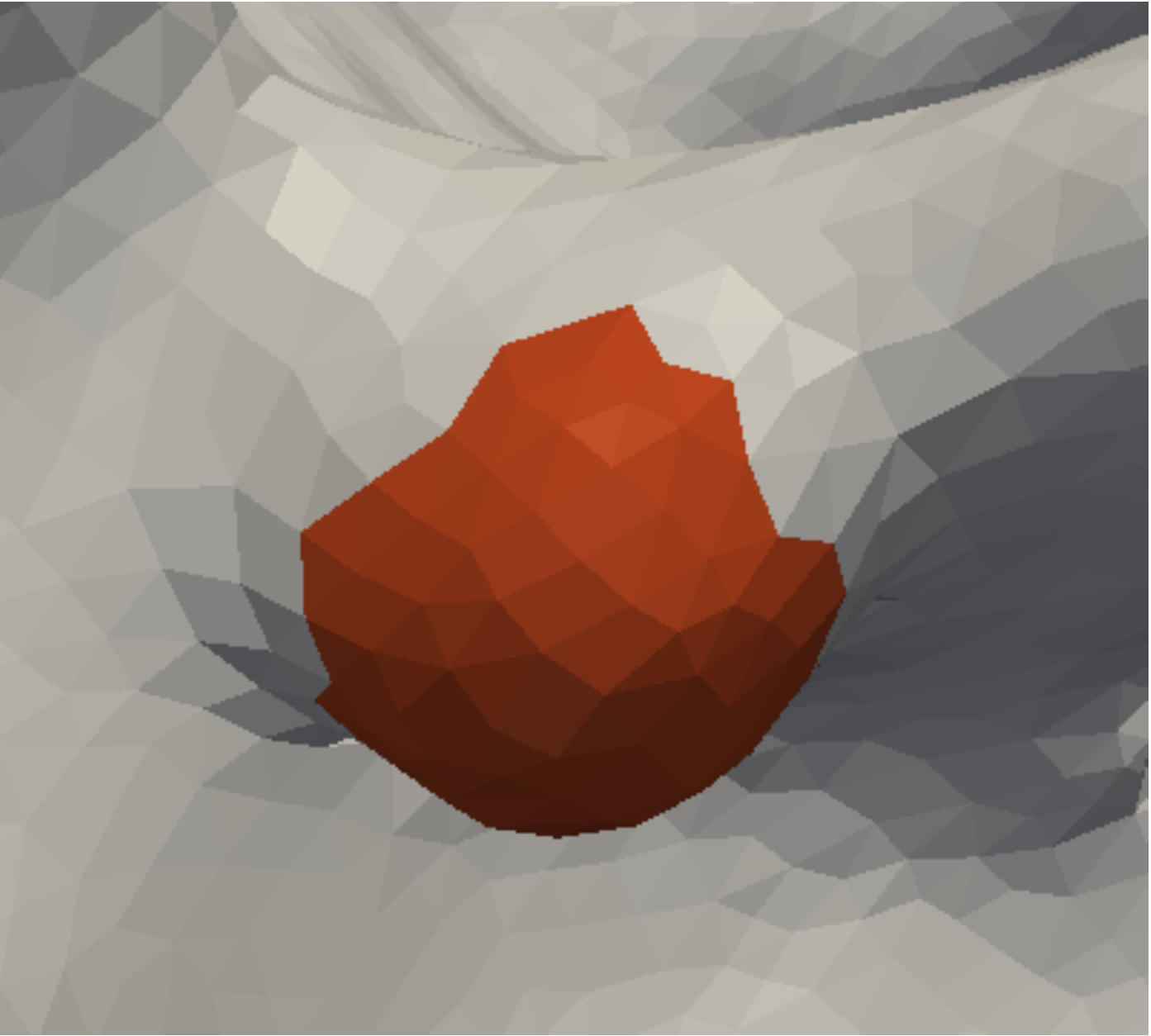}
\caption{Polyp with no texture information.\label{geopol}}
\end{center}

\begin{center}
\includegraphics[width=0.18\textwidth]{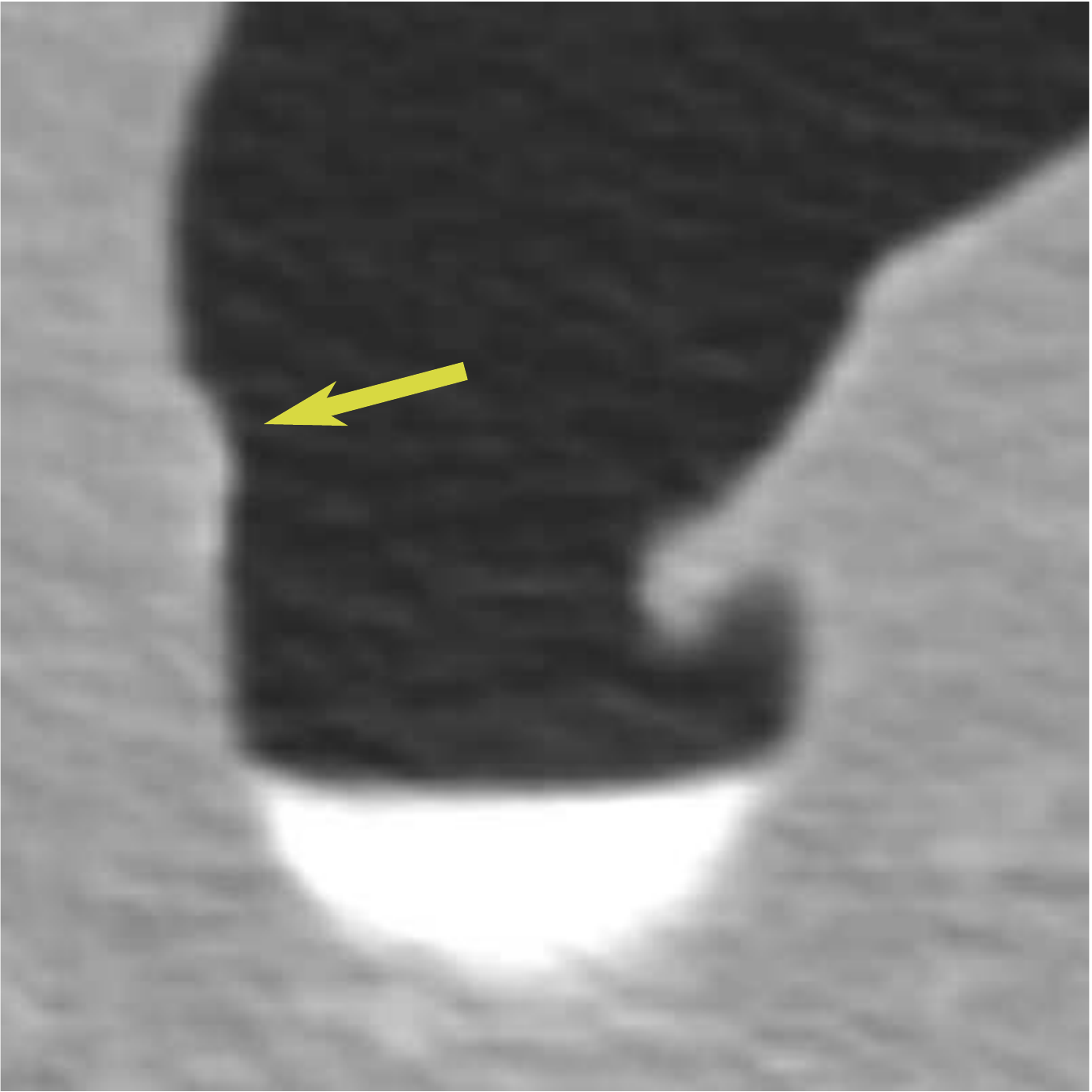}
\includegraphics[width=0.2\textwidth]{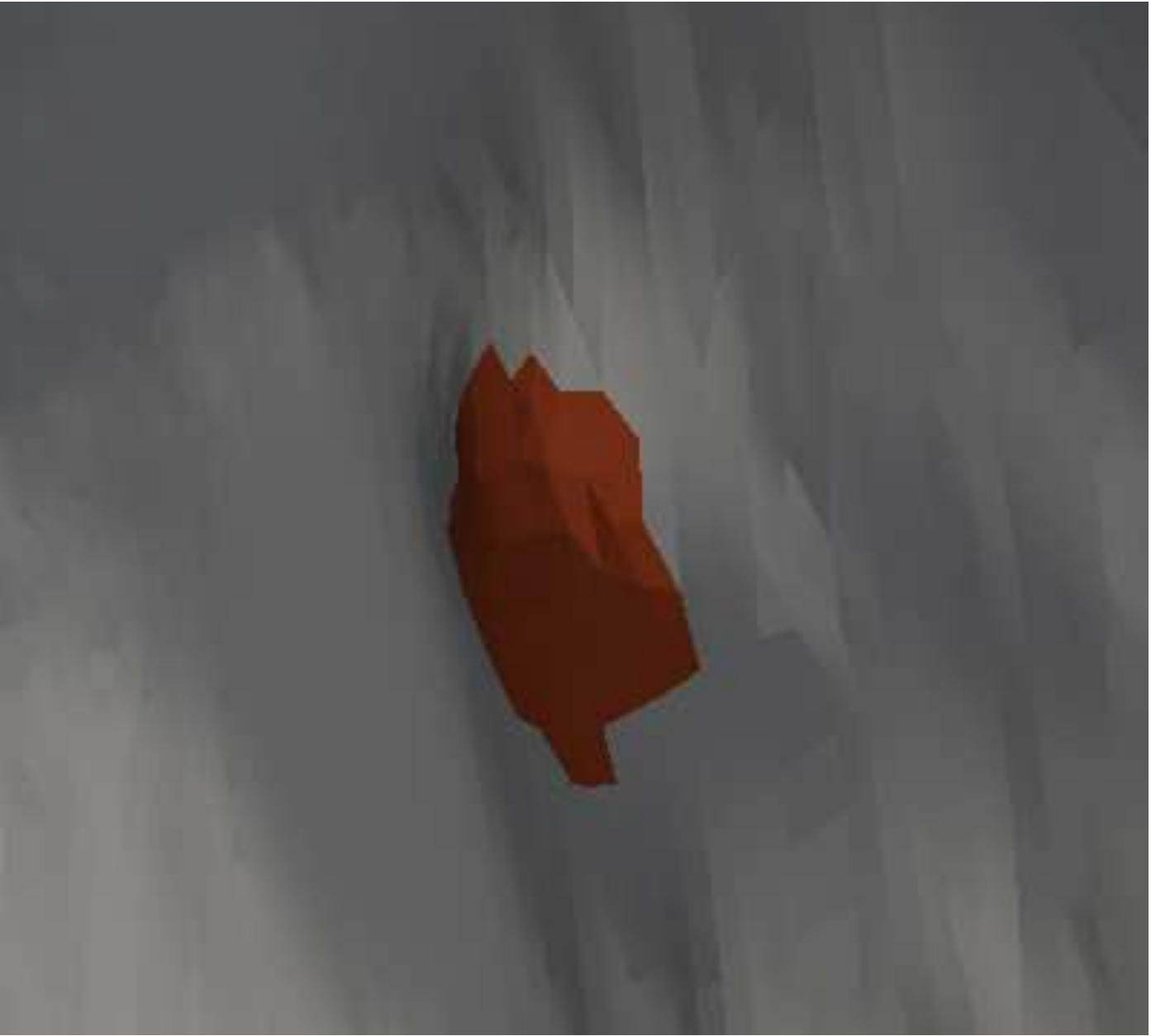}
\caption{Polyp with texture information, but weak geometric information.\label{textpol}}
\end{center}
\end{figure}

Figure \ref{geopol} shows a detected polyp, where the geometry is crucial, because the gray-level does not present considerable local variations. This is specially true for polyps submerged in tagged material. On the other hand, in the flat polyp of Figure \ref{textpol}, the geometry is weakly discriminative (although the measure considering the ring enhances the detectability), and the texture features lead to a correct classification.

Texture information is very important also because it is more robust to segmentation errors,
as the texture features are computed by integrating over the volumetric data. 
Moreover, the differential texture features (the differences between $V_1$ and $V_2$) outperform the absolute texture features (just computed in $V_1$), as shown in Table \ref{tablaresult}. 

\vspace{0cm}

\subsection{Qualitative analysis of false positives}


In addition to the number of false positives, it is very important to study how these FP patches look like, since some of them can be quickly ruled out by the expert and some can be avoided by improving some aspects of the segmentation step.

\begin{figure}[h!]
\begin{center}
\includegraphics[width=0.48\columnwidth]{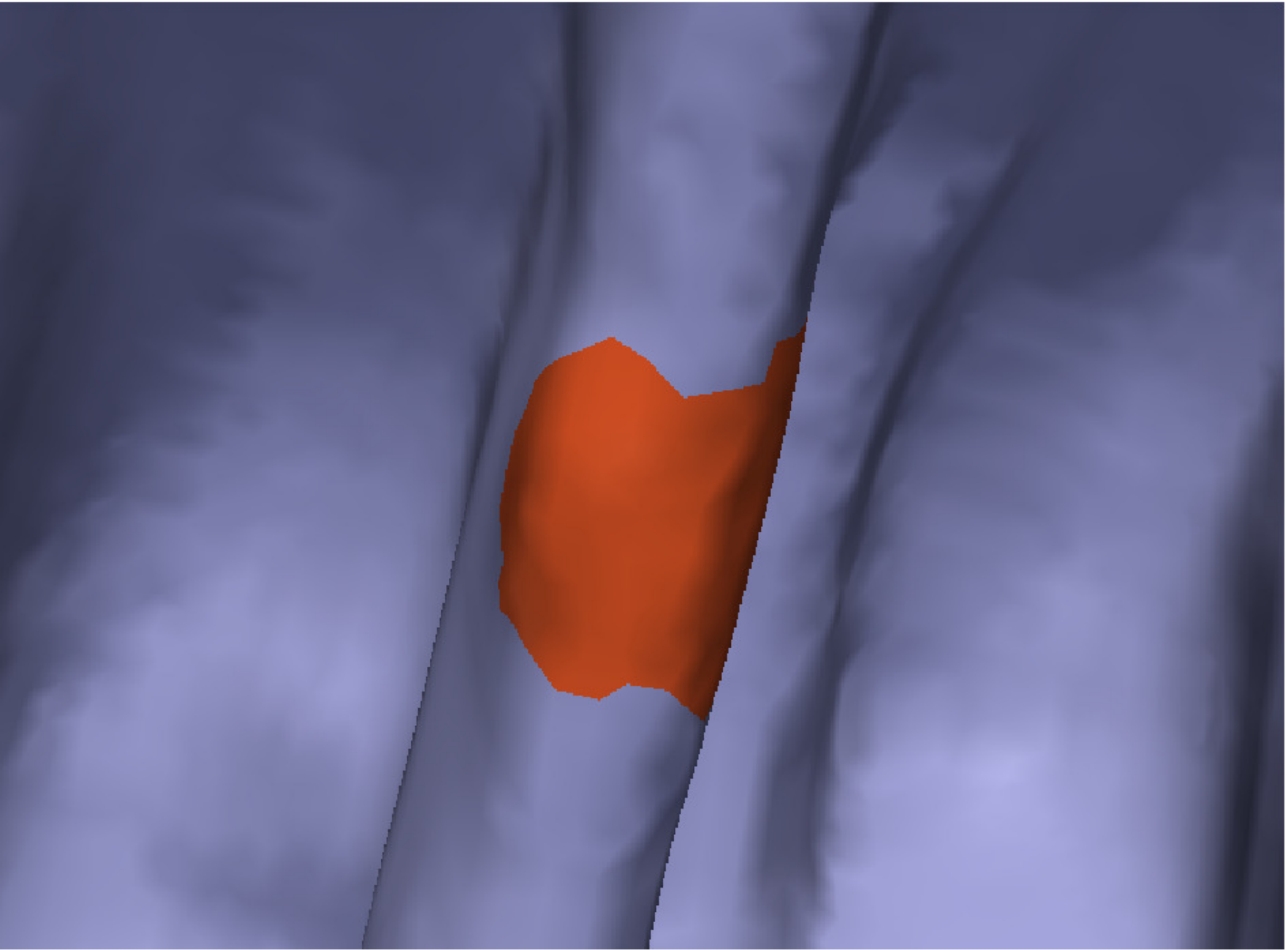}
\includegraphics[width=0.48\columnwidth]{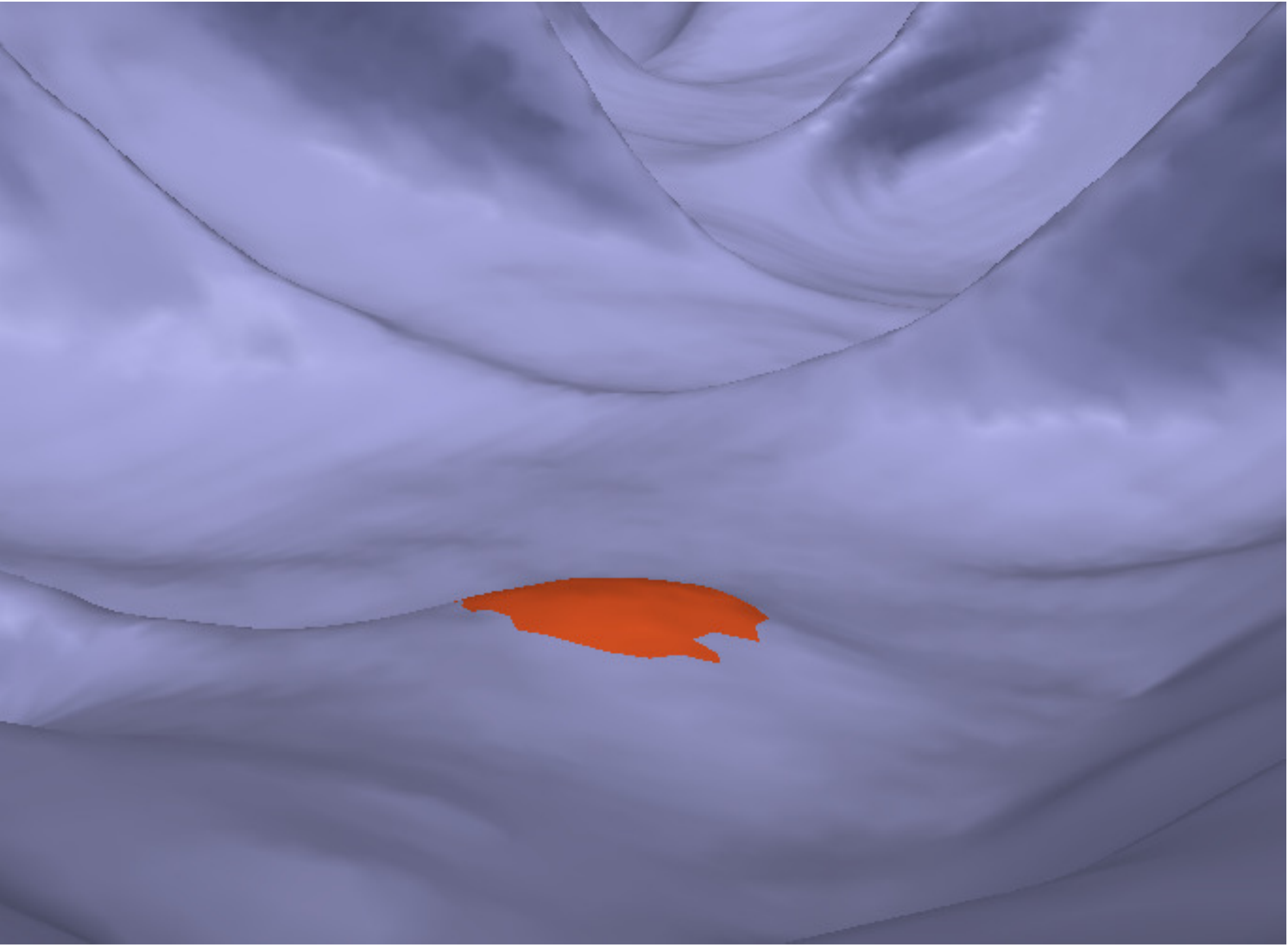}
\caption{False positives: fold and patch similar to polyp.\label{foldpolike}}
\end{center}
\end{figure}

\begin{figure}[h!]
\begin{center}
{\includegraphics[width=0.2\columnwidth]{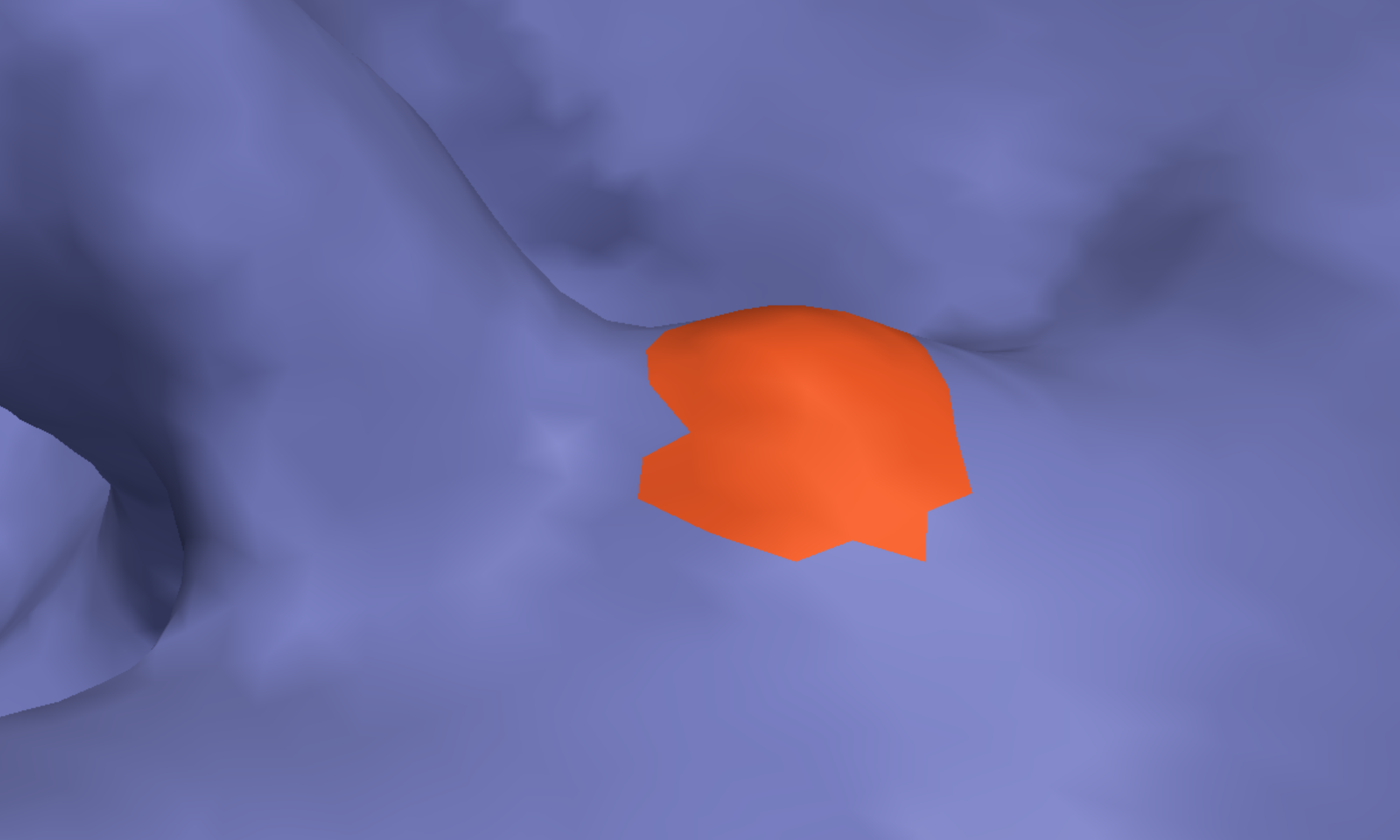}}
{\includegraphics[width=0.2\columnwidth]{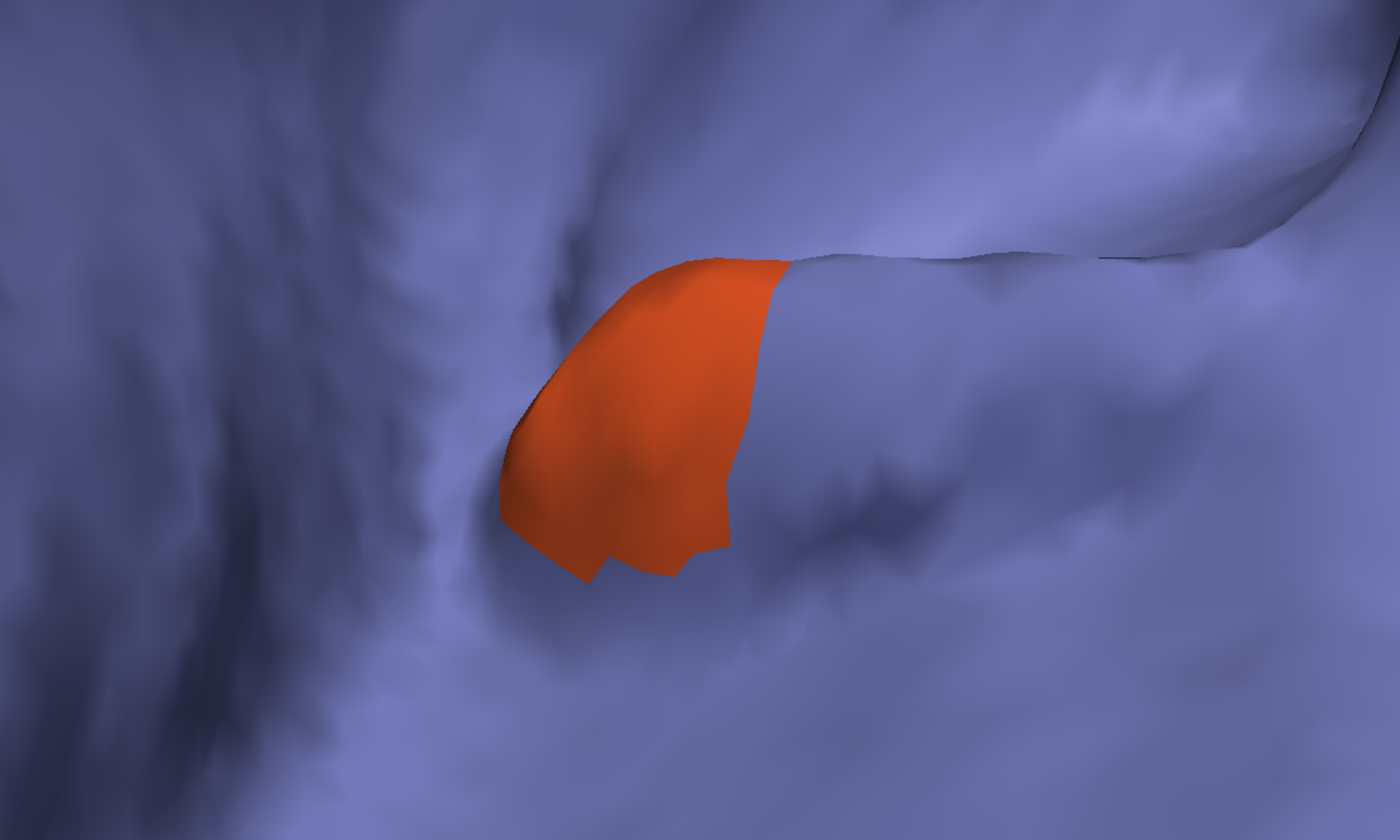}}
{\includegraphics[width=0.2\columnwidth]{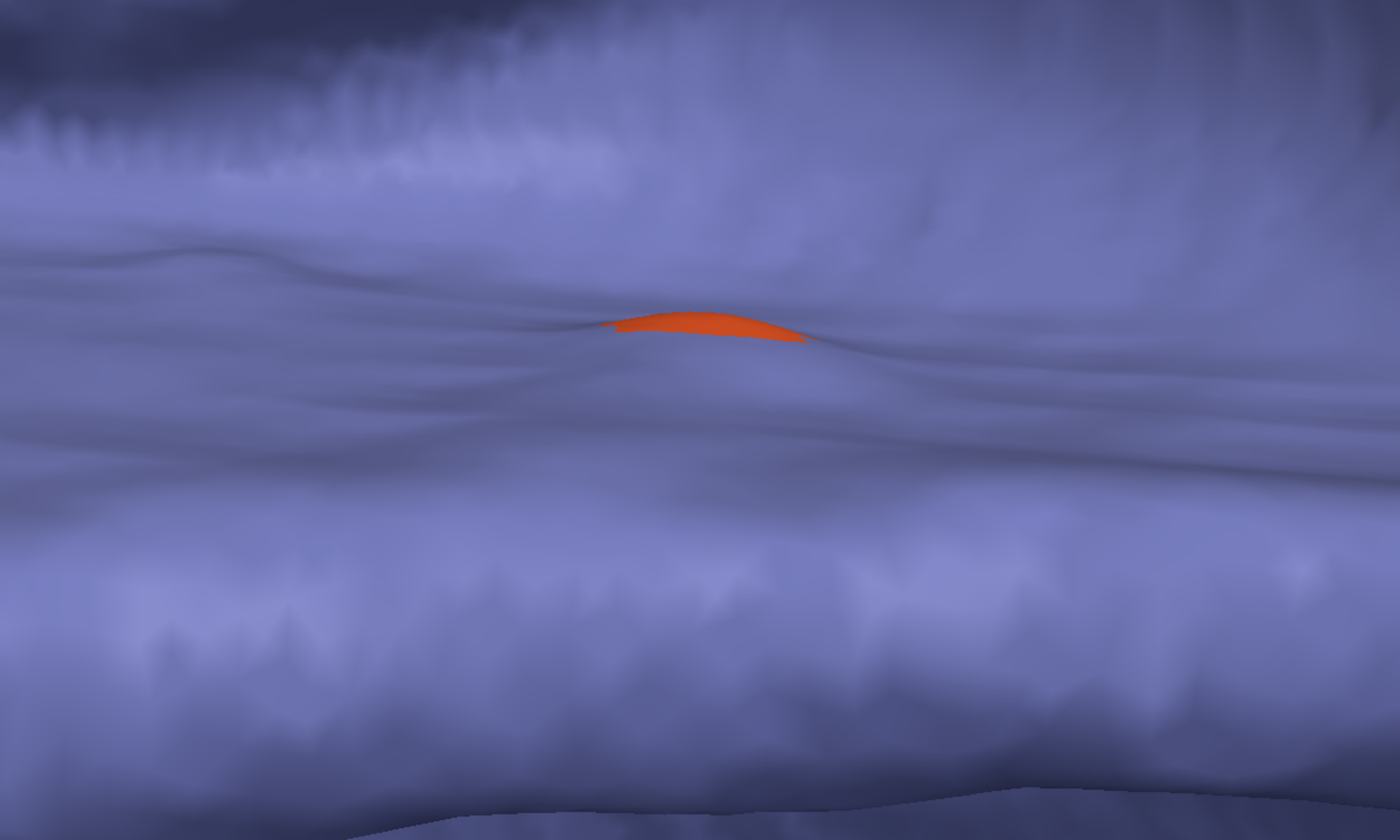}}
{\includegraphics[width=0.2\columnwidth]{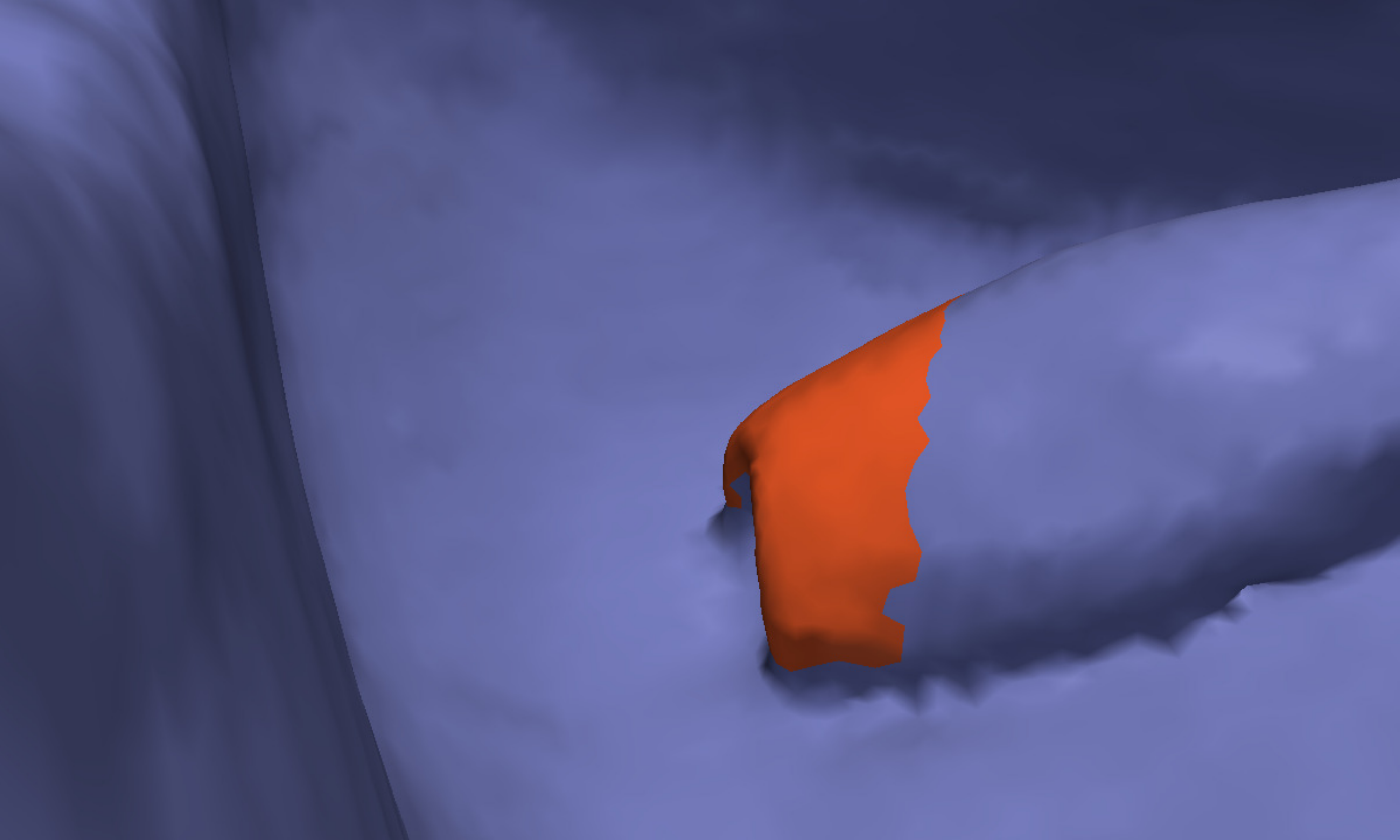}}

\vspace{0.1cm}
{\includegraphics[width=0.2\columnwidth]{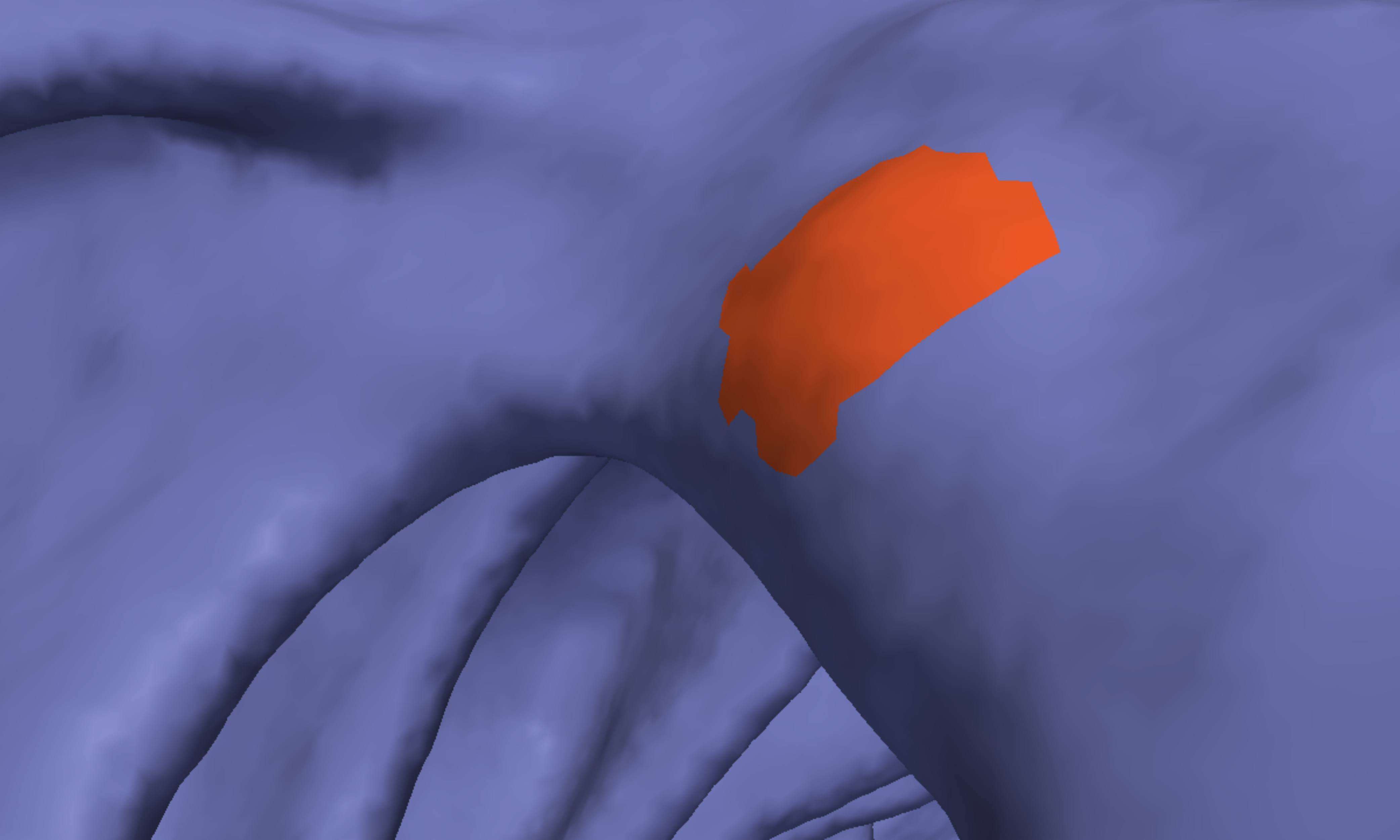}}
{\includegraphics[width=0.2\columnwidth]{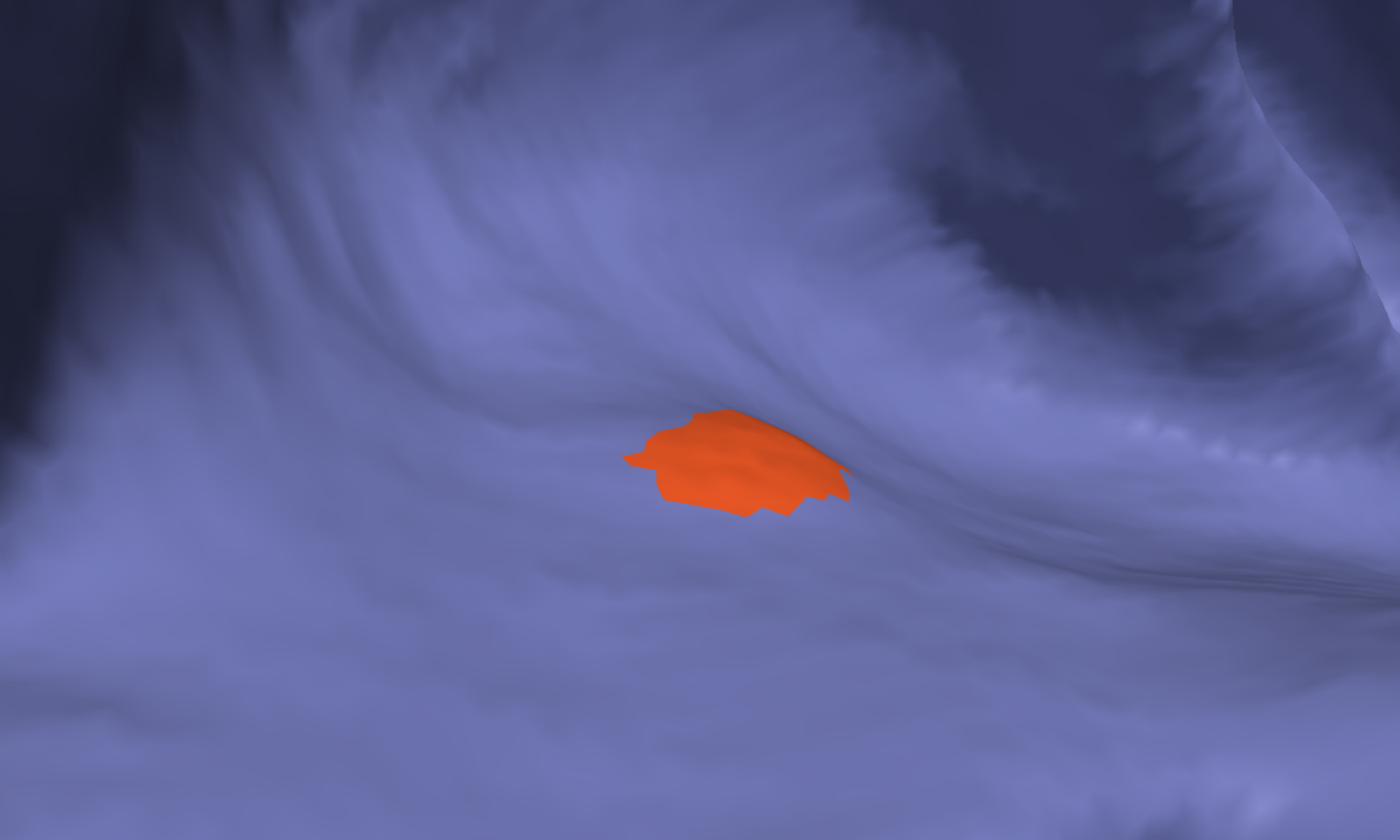}}
{\includegraphics[width=0.2\columnwidth]{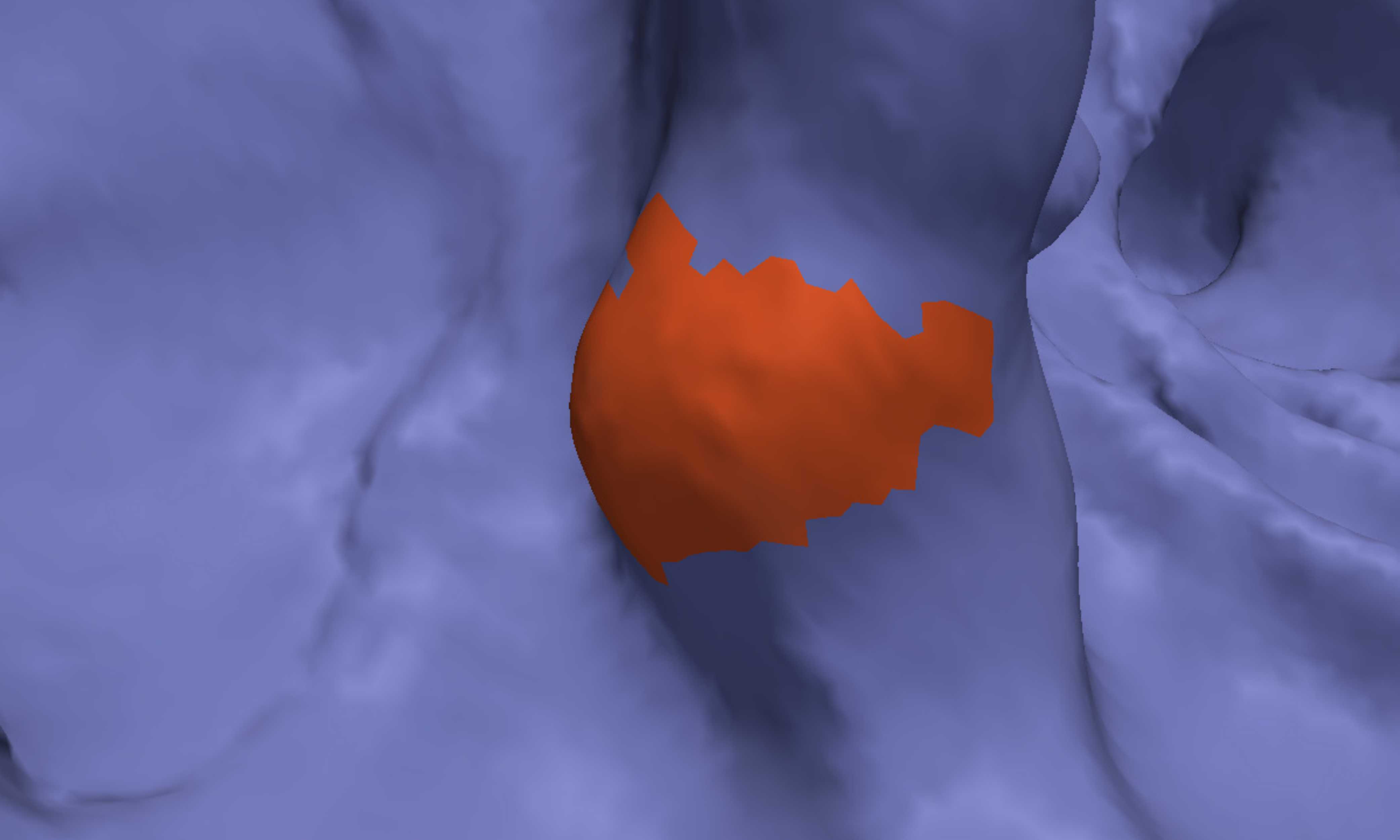}}
{\includegraphics[width=0.2\columnwidth]{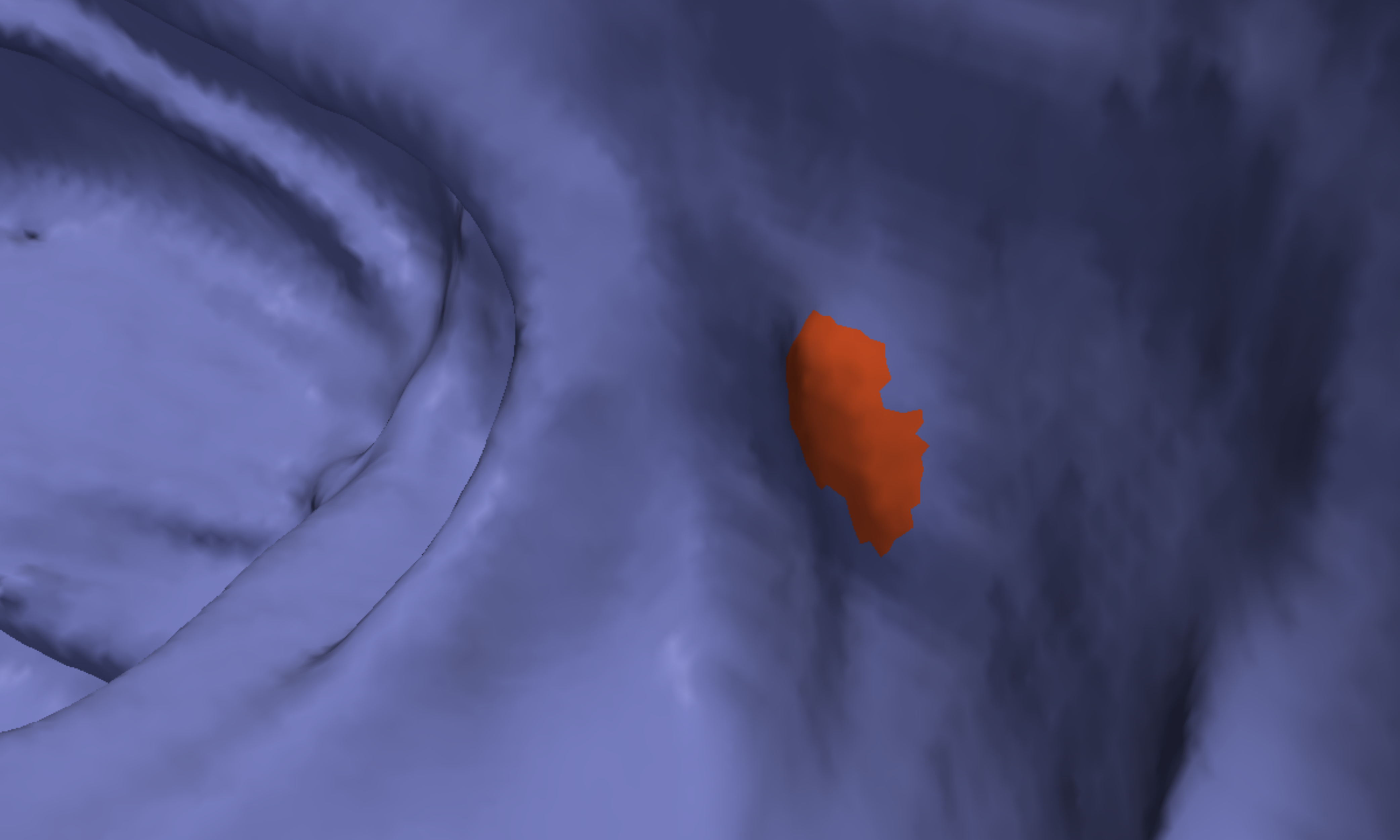}}

\vspace{0.1cm}
{\includegraphics[width=0.2\columnwidth]{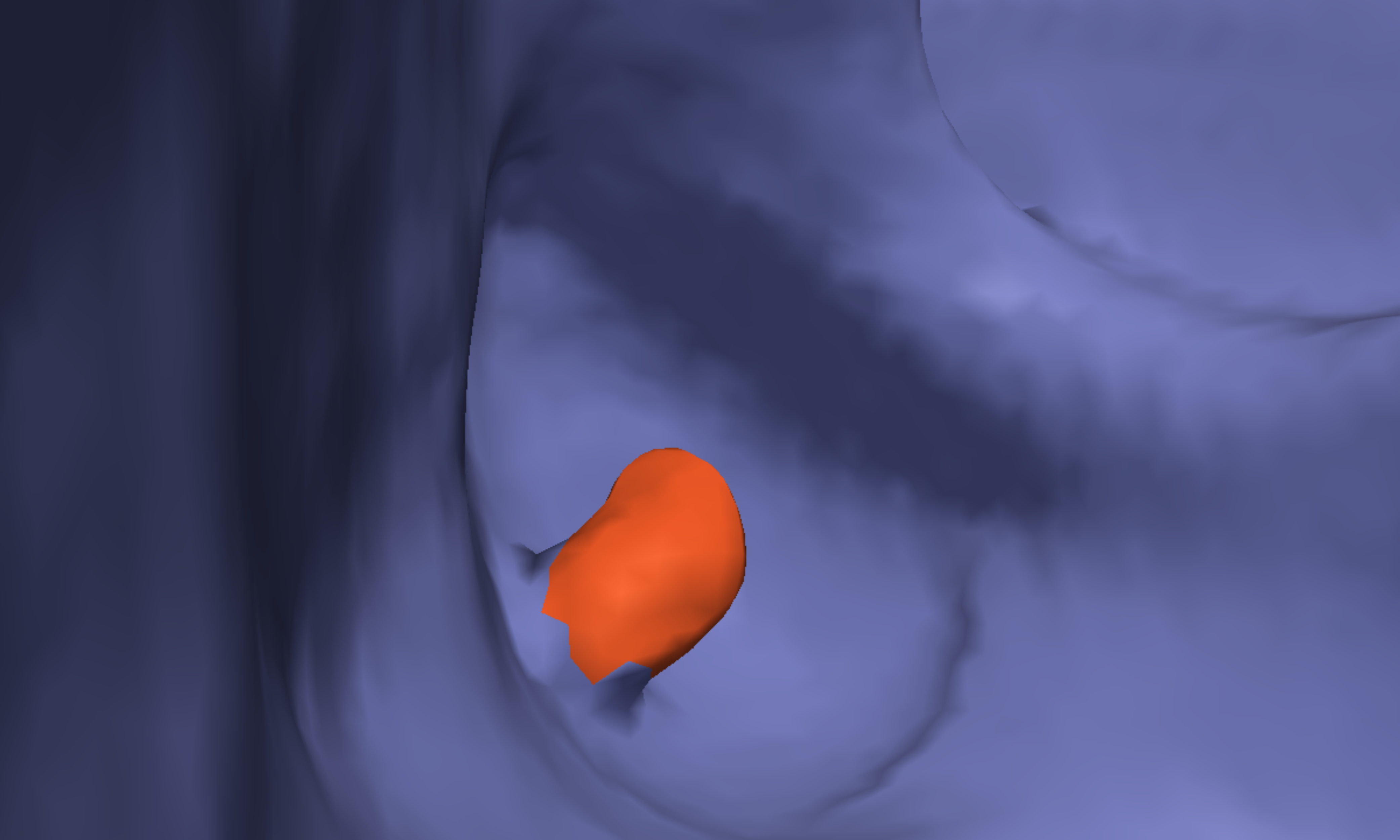}}
{\includegraphics[width=0.2\columnwidth]{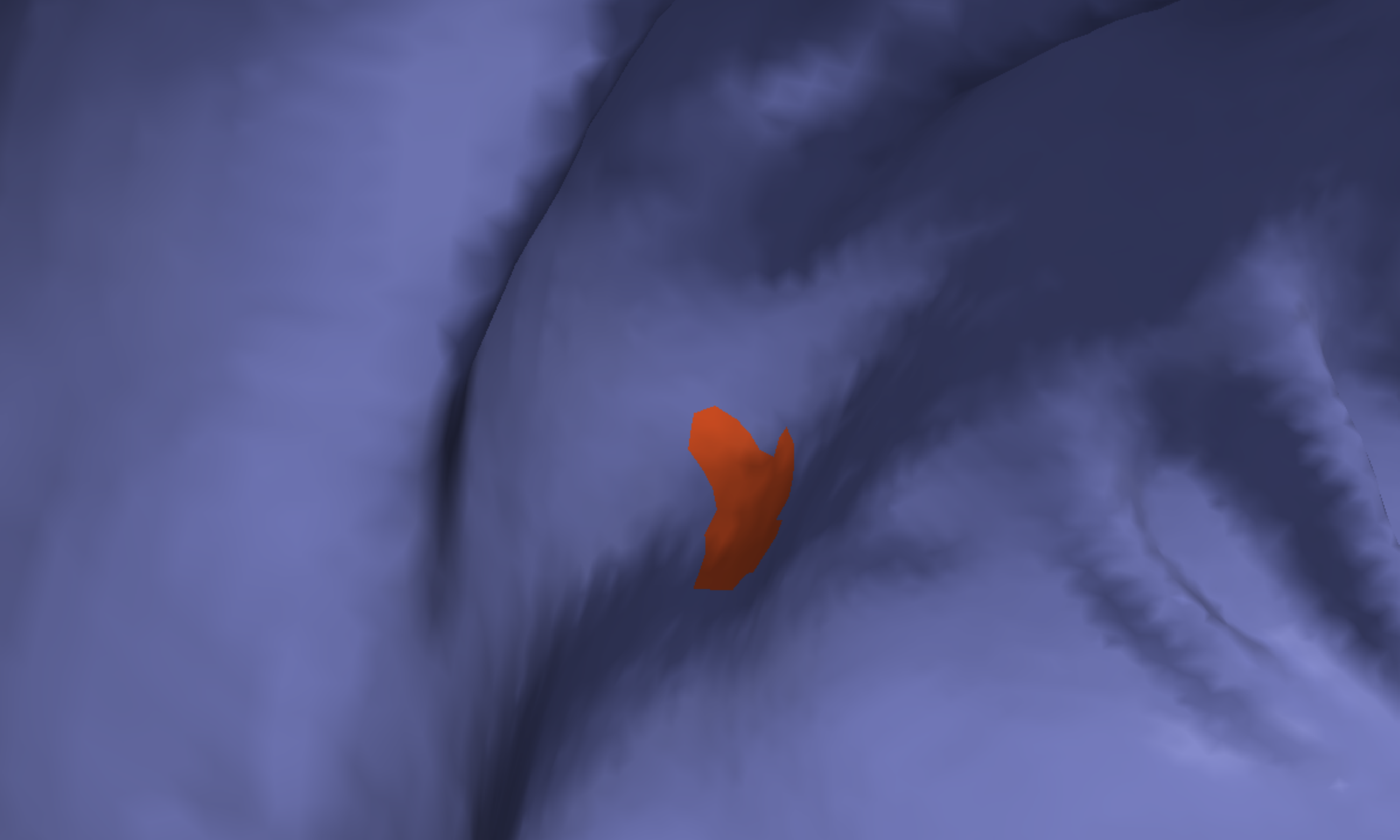}}
{\includegraphics[width=0.2\columnwidth]{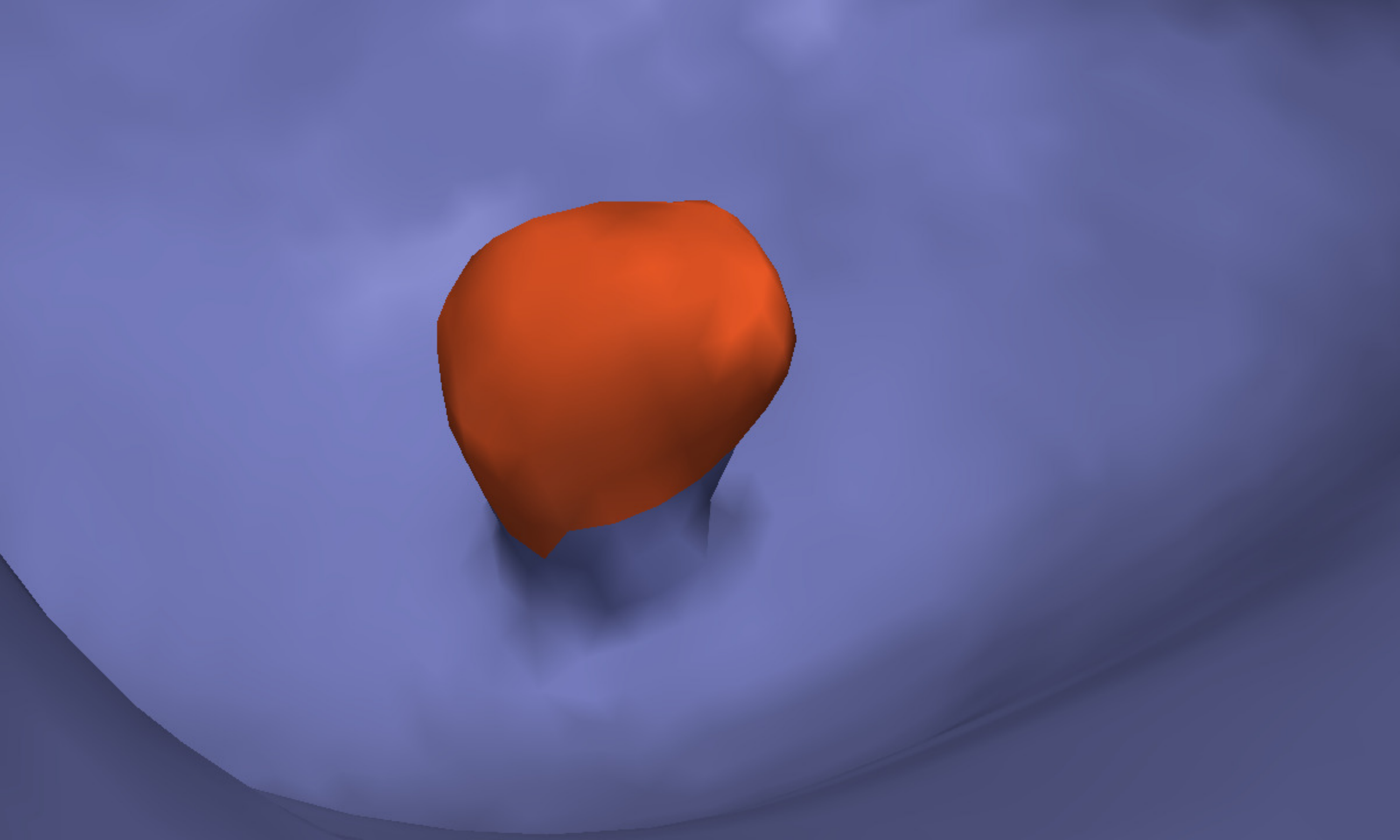}}
{\includegraphics[width=0.2\columnwidth]{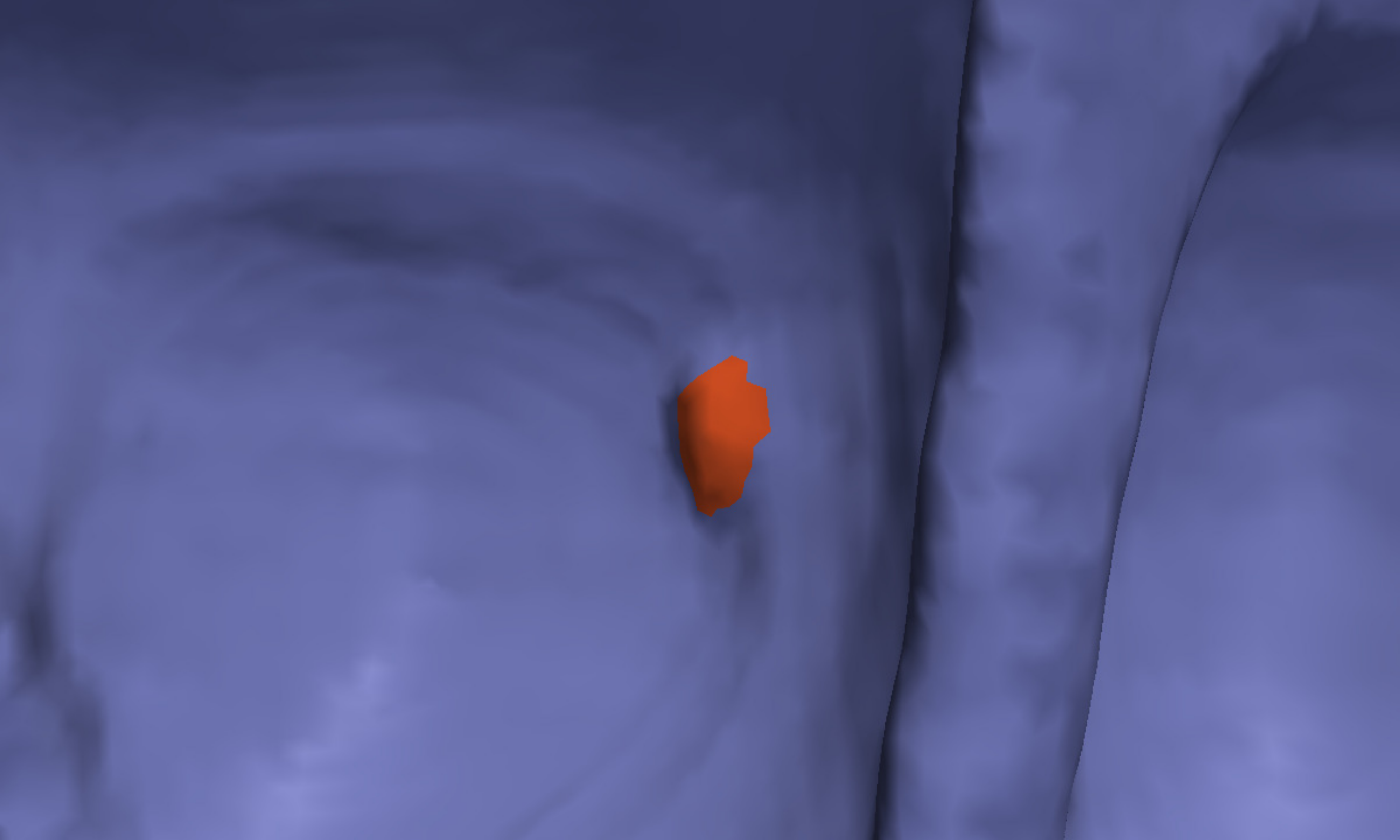}}

\vspace{0.1cm}
{\includegraphics[width=0.2\columnwidth]{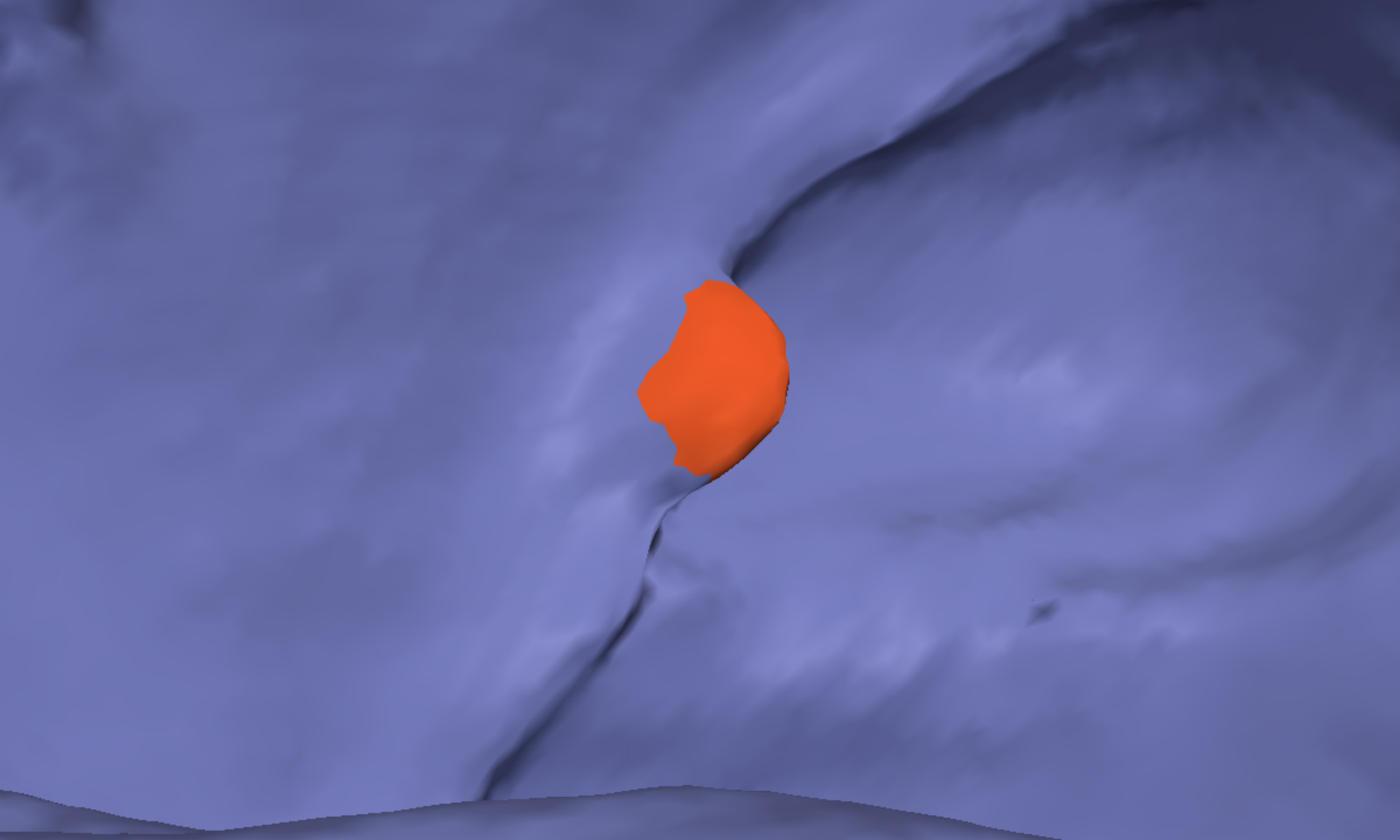}}
{\includegraphics[width=0.2\columnwidth]{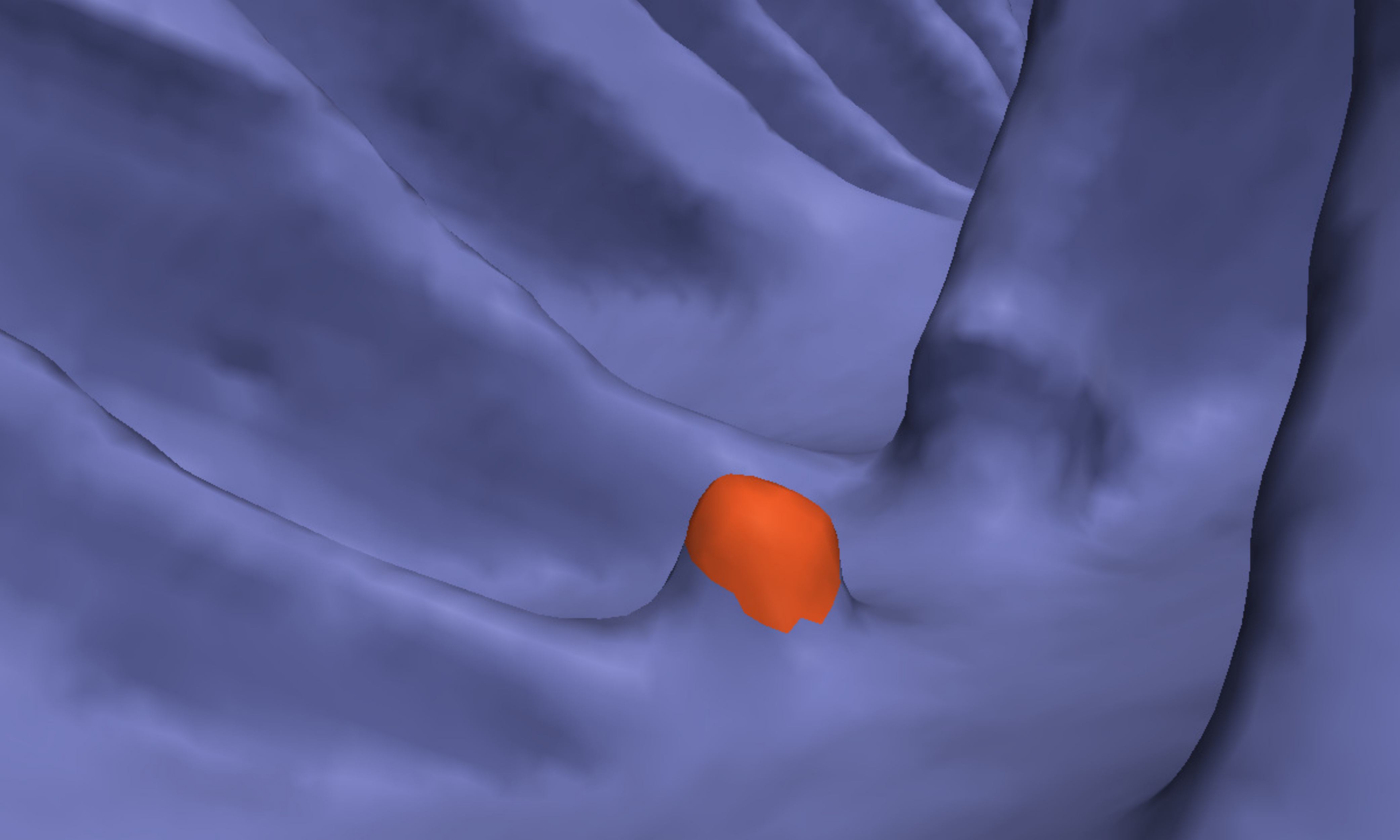}}
{\includegraphics[width=0.2\columnwidth]{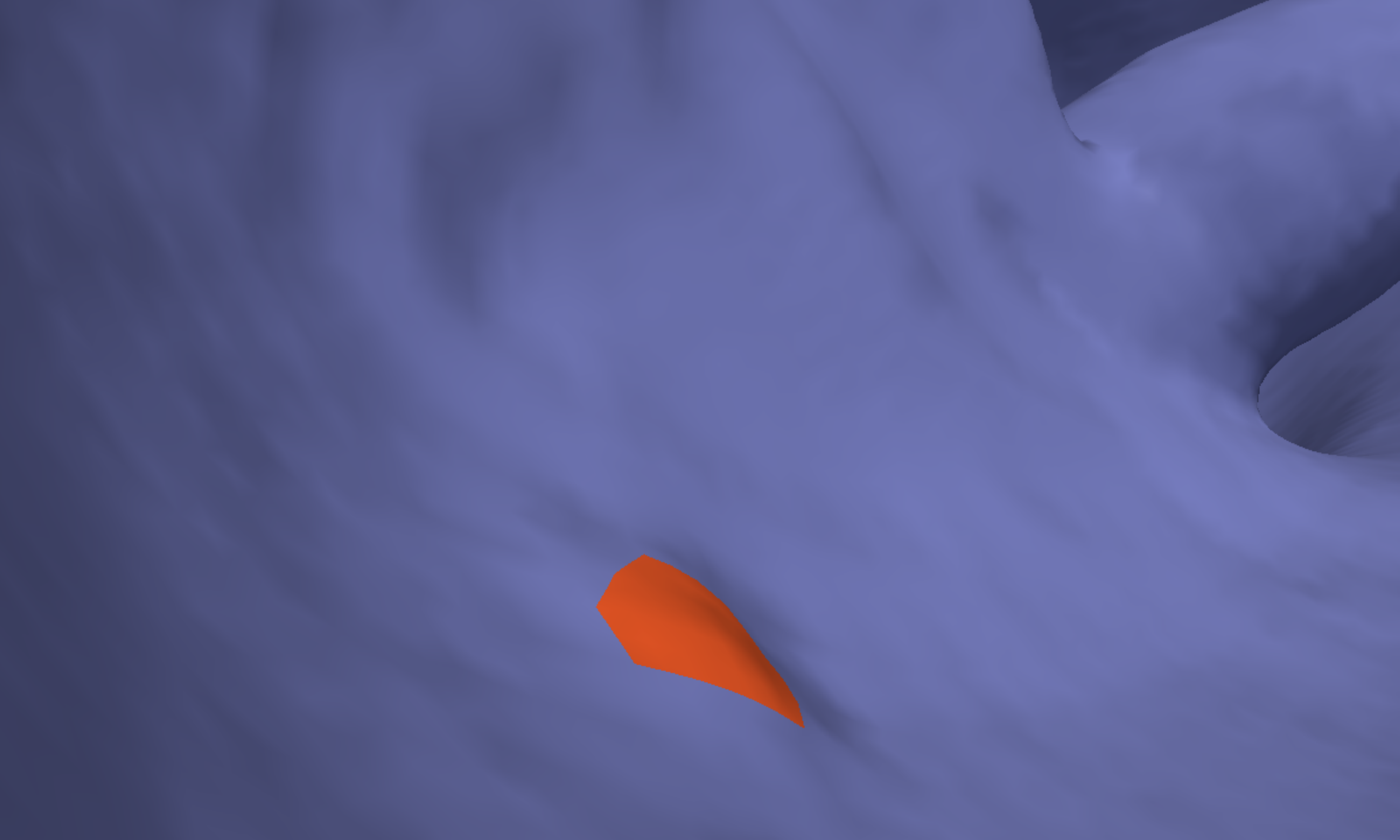}}
{\includegraphics[width=0.2\columnwidth]{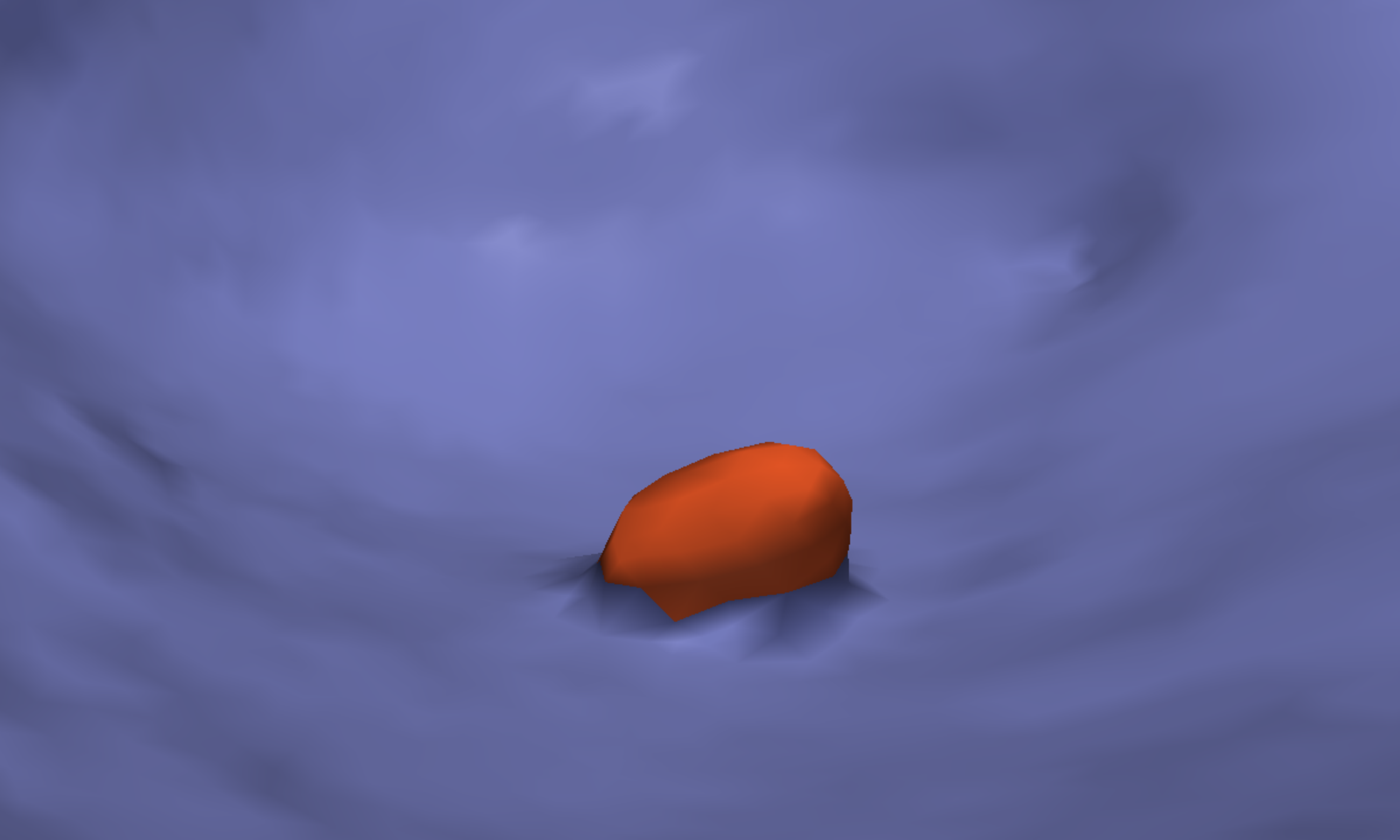}}
\caption{Examples of false positives according to the available labeled data, with some segmentation errors, parts of the insufflation tube, and
some patches with polyp-like shape.\label{fig:fps}}
\end{center}
\end{figure}

About half of the false positives are quite reasonable, in the sense that they are (usually small) sections of the colon that are polyp-like shaped, (see Figure \ref{foldpolike}), specially taking into account that we designed the system to also detect small and flat polyps. Among these false positives, $40\%$ are very small patches that may be avoided by incorporating some new features to the classification, or by adding a size threshold if very small polyps are not considered of interest (which was not our case). 
On the other hand, about $10\%$ of the FPs occurred in fold sections of the wall, Figure \ref{foldpolike}, and another $10\%$ occurred in parts of the insufflation tube. All these patches (from folds and the insufflation tube), are easily ruled out by visual inspection. Another $20\%$ of the FPs were caused by colon segmentation errors, this number having been significantly reduced thanks to the texture features as discussed above. \red{About half of these false positives due to segmentation errors are caused by bad quality original CT slices in the region (generally due to the partial volume effect), like the bottom-left example in Figure \ref{fig:fps}. Another fraction of FP due to segmentation are protuberances caused by some fluid voxels near to the colon wall, where the gray value is away from standard values. This is the case of the bottom-right example in Figure \ref{fig:fps}.}
An additional representative set of false positives is shown in Figure \ref{fig:fps}.

\begin{figure}[h!]
\begin{center}
{\includegraphics[width=0.48\columnwidth]{fp14.pdf}}
{\includegraphics[width=0.44\columnwidth]{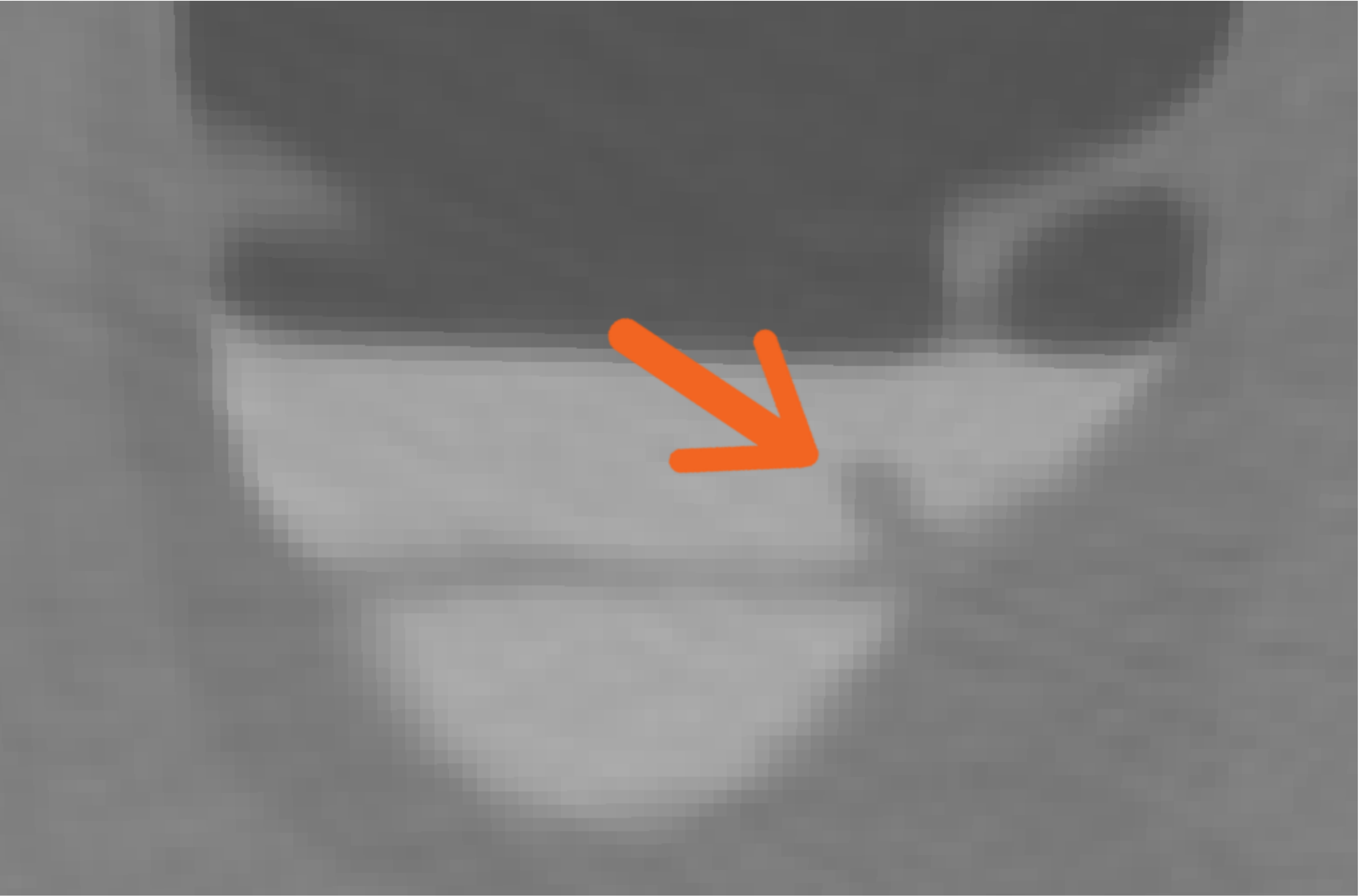}}
\caption{Visual inspection of the CT slices seems to support that the geometry is well reconstructed by the segmentation procedure. Nevertheless, since this structure was not labeled as a polyp in the ground truth database, here we consider this detection to be a false positive.\label{fig:polyp_nolabeled}}
\end{center}
\end{figure}

The region in Figure \ref{fig:polyp_nolabeled} was classified as polyp by our system. However it was not labeled as such in the database, so we have decided to count it as a false positive. We have first included this false positive in the ``segmentation error'' category, as suggested by visual inspection. Nevertheless, a more careful examination using the original CT slices shows that the structure is indeed a tissue protuberance, and it is not a segmentation artifact. It may be a part of the colon not explored in the OC. Unfortunately we have no additional information that would allow us to clarify whether this region corresponds to a polyp or not.

\subsection{Contribution of each novelty}

In this work we have presented a complete system for polyp flagging, with several novelties. Each one of these novelties, presented in the introduction and detailed along the paper, contributes to different aspects and in different ways to reach the final system performance: 

\begin{itemize}
\item Artifacts due to the tagged fluid are usually a very important source of false positives. We have minimized these false positives with our region pre-processing approach.
\item The smoothing PDE algorithm helps to detect small polyps, and it has proven to outperform the classical curvature motions.
\item The \textit{adaptive-scale} candidate search provides a correct delineation of the polyp region, as illustrated in the figures presented along the paper.
\item The new \textit{differential features} improve the performance with respect to the features measured on the polyp region only, as shown by the corresponding FROC curves.
\item The use of \textit{Cost Sensitive} learning is crucial to overcome the class imbalance problem. It has proven successful, allowing to reach higher performance than other techniques also specifically designed to deal with the class imbalance problem.
\end{itemize}

\section{Conclusion}
\label{conclusions}

We introduced a complete pipeline for a Computer Aided Detection algorithm that flags candidate polyp regions.
The segmentation stage is very simple and fast, and its main novelty is the smoothing PDE which enhances the polyps, leading to a better detection.
In addition to the incorporation of the Haralick texture features, the main yet simple novelties of the proposed features and classification stages are twofold. First, the surrounding area of candidate polyps are explicitly taken into account. Indeed, the proposed (so-called differential) features are computed by comparing properties in the central and surrounding regions of the polyps.
We show that differential features are more discriminative than the absolute ones, as they emphasize local deviations of the geometry and texture over the colon.
The other novelty is a adaptive-scale strategy that test regions of different sizes and automatically selects the region that best delineates each candidate polyp.
The obtained quantitative results are very promising, detecting $100\%$ of the true-polyps, including small and flat ones, with a low false positives rate.
Additional improvement of the segmentation and, in collaborations with radiologists,
finding features that are tailored to polyp-like geometries, can further improve these results.

\section*{Acknowledgments}

Work supported in part by NSF, NGA, ARO, AFOSR and ANII-Uruguay. We thank Sergio Aguirre from Echopixel for important feedback regarding VC.

\bibliographystyle{IEEEtran}
\bibliography{biblio}

\appendix

\section{Number of iterations: analytical solving}
\label{apendice}

We want to compute the number of iterations needed to make a certain sphere (of the size of the CT resolution) vanish, according to our proposed $PDE$.
For a sphere, the shape index is constant, so the $PDE$ becomes:
 \[ \frac{\partial \S}{\partial t} = \k_{min}^{1/4}\Nm \,\, .\]
Then, as $k_{min}=1/r$, the sphere radious satisfies the following differential equation:
\[r'(t) = \frac{-1}{r(t)^{1/4}} \,\, .\]
Therefore,
\[\int_0^T r'(t)r(t)^{1/4}dt=-\int_0^T1dt \Rightarrow \frac{4}{5}\left(r(T)^{4/5}-r(0)^{4/5} \right) = -T \,\, .\]
As we want to find the value of $T$ so that $r(T)=0$,
\[T=\frac{4}{5}r(0)^{4/5} \,\, .\]
The resolution in the $z$ direction is $1mm$ in our examples, and the time step considered for the numerical method was $t_s=0.055$. That gives a value $N$ for the number of iterations $N=\lceil14.54\rceil = 15$.



\end{document}

%% file: tabla_review.tex

%
%

\begin{table}[h!]
\begin{center}
\begin{tabular}{|l|c|c|c|c|}
\hline
Technique & 				\# Polyps & 		Size & 		Sensit. & 		FPs  \\
\hline
Summers {\it et al.} \cite{summers2005} & 		$\approx 30$ &		$>10mm$ &		$89.3\%$ &		$2.1$ \\
Summers {\it et al.} & 		$\approx 120$ &	$>6mm$ &		$60\%$	 &		$8$  \\
\hline
Yoshida {\it et al.} \cite{yoshida} &		$12$ &			$>5mm$ &		$100\%$	 &		$2$  \\
\hline
Paik {\it et al.} \cite{paik} &			$7$ &			$>10mm$ &		$100\%$	 &		$7$ \\
\hline
Wang {\it et al.} \cite{wang2005_medphys} &		$61$ &			$>4mm$ &		$100\%$	 &		$2.68$ \\
\hline
Hong {\it et al.} \cite{hong} &	$\approx 120$ &			$>5mm$ &		$100\%$	 &		$3$ \\ 
\hline
Sundaram {\it et al.} \cite{sundaram} & $20$ &		$>10mm$ &		$100\%$ &		$18$  \\
Sundaram {\it et al.} & $122$ &		$>2mm$ &		$80\%$ &		$24$ \\
\hline
van Wijk {\it et al.} \cite{vanwijk} &	$57$ &			$>6mm$ &		$95\%$ &		$5$ \\
van Wijk {\it et al.} &	$32$ &			$>10mm$ &		$95\%$ &		$4$ \\
\hline
Suzuki {\it et al.} \cite{suzuki} &	$28$ &			$>5mm$ &		$100\%$ &	$1.1$ \\
\hline
Bogoni {\it et al.} \cite{bogoni} &	$21$ &			$>5mm$ &		$90\%$ &		$>3$ \\
\hline
Taylor {\it et al.} \cite{taylor} &	$32$ &			$>6mm$ &		$81\%$ &		$13$ \\
\hline
N\"{a}ppi {\it et al.} \cite{Nappi2002} & $12$ & $>5mm$ & $100\%$ & $2.4$ \\ 
\hline
\end{tabular}
\end{center}
\vspace{-0.2cm}
\caption{Numerical comparison of the reviewed methods, indicating the number and size of the polyps in the database, and the achieved sensitivity with the false positives (FPs) per study.}
\label{tabla_review}
\end{table}


%% file: roc_1.pdf_tex
\begingroup%
  \makeatletter%
  \providecommand\color[2][]{%
    \errmessage{(Inkscape) Color is used for the text in Inkscape, but the package 'color.sty' is not loaded}%
    \renewcommand\color[2][]{}%
  }%
  \providecommand\transparent[1]{%
    \errmessage{(Inkscape) Transparency is used (non-zero) for the text in Inkscape, but the package 'transparent.sty' is not loaded}%
    \renewcommand\transparent[1]{}%
  }%
  \providecommand\rotatebox[2]{#2}%
  \ifx\svgwidth\undefined%
    \setlength{\unitlength}{494.258bp}%
    \ifx\svgscale\undefined%
      \relax%
    \else%
      \setlength{\unitlength}{\unitlength * \real{\svgscale}}%
    \fi%
  \else%
    \setlength{\unitlength}{\svgwidth}%
  \fi%
  \global\let\svgwidth\undefined%
  \global\let\svgscale\undefined%
  \makeatother%
  \begin{picture}(1,0.56039154)%
    \put(0,0){\includegraphics[width=\unitlength]{roc_1.pdf}}%
    \put(0.07602103,0.03692808){\makebox(0,0)[lb]{\smash{0}}}%
    \put(0.21360099,0.03692808){\makebox(0,0)[lb]{\smash{0.5}}}%
    \put(0.37869696,0.03692808){\makebox(0,0)[lb]{\smash{1}}}%
    \put(0.51789551,0.03692808){\makebox(0,0)[lb]{\smash{1.5}}}%
    \put(0.68137289,0.03692808){\makebox(0,0)[lb]{\smash{2}}}%
    \put(0.82057144,0.03692808){\makebox(0,0)[lb]{\smash{2.5}}}%
    \put(0.9856674,0.03692808){\makebox(0,0)[lb]{\smash{3}}}%
    \put(0.06145373,0.05796972){\makebox(0,0)[lb]{\smash{0}}}%
    \put(0.03717492,0.155085){\makebox(0,0)[lb]{\smash{0.2}}}%
    \put(0.03717492,0.25220027){\makebox(0,0)[lb]{\smash{0.4}}}%
    \put(0.03717492,0.34769695){\makebox(0,0)[lb]{\smash{0.6}}}%
    \put(0.03717492,0.44481222){\makebox(0,0)[lb]{\smash{0.8}}}%
    \put(0.06145373,0.54030891){\makebox(0,0)[lb]{\smash{1}}}%
    \put(0.45153341,0.00617491){\makebox(0,0)[lb]{\smash{FPs per case}}}%
    \put(0.02098904,0.24410733){\rotatebox{90}{\makebox(0,0)[lb]{\smash{Sensitivity}}}}%
  \end{picture}%
\endgroup%

%% file: roc_smooth_grandes_1.pdf_tex
\begingroup%
  \makeatletter%
  \providecommand\color[2][]{%
    \errmessage{(Inkscape) Color is used for the text in Inkscape, but the package 'color.sty' is not loaded}%
    \renewcommand\color[2][]{}%
  }%
  \providecommand\transparent[1]{%
    \errmessage{(Inkscape) Transparency is used (non-zero) for the text in Inkscape, but the package 'transparent.sty' is not loaded}%
    \renewcommand\transparent[1]{}%
  }%
  \providecommand\rotatebox[2]{#2}%
  \ifx\svgwidth\undefined%
    \setlength{\unitlength}{418.09377125bp}%
    \ifx\svgscale\undefined%
      \relax%
    \else%
      \setlength{\unitlength}{\unitlength * \real{\svgscale}}%
    \fi%
  \else%
    \setlength{\unitlength}{\svgwidth}%
  \fi%
  \global\let\svgwidth\undefined%
  \global\let\svgscale\undefined%
  \makeatother%
  \begin{picture}(1,0.79682728)%
    \put(0,0){\includegraphics[width=\unitlength]{roc_smooth_grandes_1.pdf}}%
    \put(0.13953275,0.05642575){\makebox(0,0)[lb]{\smash{0}}}%
    \put(0.24285885,0.05642575){\makebox(0,0)[lb]{\smash{0.2}}}%
    \put(0.36149252,0.05642575){\makebox(0,0)[lb]{\smash{0.4}}}%
    \put(0.4801262,0.05642575){\makebox(0,0)[lb]{\smash{0.6}}}%
    \put(0.59875987,0.05642575){\makebox(0,0)[lb]{\smash{0.8}}}%
    \put(0.73078767,0.05642575){\makebox(0,0)[lb]{\smash{1}}}%
    \put(0.83602722,0.05642575){\makebox(0,0)[lb]{\smash{1.2}}}%
    \put(0.95466089,0.05642575){\makebox(0,0)[lb]{\smash{1.4}}}%
    \put(0.07974348,0.11677409){\makebox(0,0)[lb]{\smash{0}}}%
    \put(0.05104179,0.24880189){\makebox(0,0)[lb]{\smash{0.2}}}%
    \put(0.05104179,0.38082969){\makebox(0,0)[lb]{\smash{0.4}}}%
    \put(0.05104179,0.51094404){\makebox(0,0)[lb]{\smash{0.6}}}%
    \put(0.05104179,0.64297184){\makebox(0,0)[lb]{\smash{0.8}}}%
    \put(0.07974348,0.77308619){\makebox(0,0)[lb]{\smash{1}}}%
    \put(0.41132878,0.0072998){\makebox(0,0)[lb]{\smash{FPs per case}}}%
    \put(0.02481262,0.33258568){\rotatebox{90}{\makebox(0,0)[lb]{\smash{Sensitivity}}}}%
  \end{picture}%
\endgroup%

%% file: roc_smooth_chicos_1.pdf_tex
\begingroup%
  \makeatletter%
  \providecommand\color[2][]{%
    \errmessage{(Inkscape) Color is used for the text in Inkscape, but the package 'color.sty' is not loaded}%
    \renewcommand\color[2][]{}%
  }%
  \providecommand\transparent[1]{%
    \errmessage{(Inkscape) Transparency is used (non-zero) for the text in Inkscape, but the package 'transparent.sty' is not loaded}%
    \renewcommand\transparent[1]{}%
  }%
  \providecommand\rotatebox[2]{#2}%
  \ifx\svgwidth\undefined%
    \setlength{\unitlength}{415.686024bp}%
    \ifx\svgscale\undefined%
      \relax%
    \else%
      \setlength{\unitlength}{\unitlength * \real{\svgscale}}%
    \fi%
  \else%
    \setlength{\unitlength}{\svgwidth}%
  \fi%
  \global\let\svgwidth\undefined%
  \global\let\svgscale\undefined%
  \makeatother%
  \begin{picture}(1,0.80857848)%
    \put(0,0){\includegraphics[width=\unitlength]{roc_smooth_chicos_1.pdf}}%
    \put(0.14747675,0.04961678){\makebox(0,0)[lb]{\smash{0}}}%
    \put(0.31298628,0.04961678){\makebox(0,0)[lb]{\smash{1}}}%
    \put(0.48042035,0.04961678){\makebox(0,0)[lb]{\smash{2}}}%
    \put(0.64785441,0.04961678){\makebox(0,0)[lb]{\smash{3}}}%
    \put(0.81528847,0.04961678){\makebox(0,0)[lb]{\smash{4}}}%
    \put(0.98272254,0.04961678){\makebox(0,0)[lb]{\smash{5}}}%
    \put(0.04848311,0.19964293){\makebox(0,0)[lb]{\smash{0.1}}}%
    \put(0.04848311,0.34590716){\makebox(0,0)[lb]{\smash{0.3}}}%
    \put(0.04848311,0.4921714){\makebox(0,0)[lb]{\smash{0.5}}}%
    \put(0.04848311,0.63843564){\makebox(0,0)[lb]{\smash{0.7}}}%
    \put(0.04848311,0.78469988){\makebox(0,0)[lb]{\smash{0.9}}}%
    \put(0.41371129,0.00734208){\makebox(0,0)[lb]{\smash{FPs per case}}}%
    \put(0.02495634,0.33451209){\rotatebox{90}{\makebox(0,0)[lb]{\smash{Sensitivity}}}}%
  \end{picture}%
\endgroup%

%% file: roc_abs_grandes_1.pdf_tex
\begingroup%
  \makeatletter%
  \providecommand\color[2][]{%
    \errmessage{(Inkscape) Color is used for the text in Inkscape, but the package 'color.sty' is not loaded}%
    \renewcommand\color[2][]{}%
  }%
  \providecommand\transparent[1]{%
    \errmessage{(Inkscape) Transparency is used (non-zero) for the text in Inkscape, but the package 'transparent.sty' is not loaded}%
    \renewcommand\transparent[1]{}%
  }%
  \providecommand\rotatebox[2]{#2}%
  \ifx\svgwidth\undefined%
    \setlength{\unitlength}{413.3620242bp}%
    \ifx\svgscale\undefined%
      \relax%
    \else%
      \setlength{\unitlength}{\unitlength * \real{\svgscale}}%
    \fi%
  \else%
    \setlength{\unitlength}{\svgwidth}%
  \fi%
  \global\let\svgwidth\undefined%
  \global\let\svgscale\undefined%
  \makeatother%
  \begin{picture}(1,0.81455965)%
    \put(0,0){\includegraphics[width=\unitlength]{roc_abs_grandes_1.pdf}}%
    \put(0.14830589,0.06137721){\makebox(0,0)[lb]{\smash{0}}}%
    \put(0.29926316,0.06137721){\makebox(0,0)[lb]{\smash{0.2}}}%
    \put(0.46763857,0.06137721){\makebox(0,0)[lb]{\smash{0.4}}}%
    \put(0.63601398,0.06137721){\makebox(0,0)[lb]{\smash{0.6}}}%
    \put(0.80438938,0.06137721){\makebox(0,0)[lb]{\smash{0.8}}}%
    \put(0.98824759,0.06137721){\makebox(0,0)[lb]{\smash{1}}}%
    \put(0.08783222,0.12672191){\makebox(0,0)[lb]{\smash{0}}}%
    \put(0.05880198,0.26026102){\makebox(0,0)[lb]{\smash{0.2}}}%
    \put(0.05880198,0.39380014){\makebox(0,0)[lb]{\smash{0.4}}}%
    \put(0.05880198,0.52540391){\makebox(0,0)[lb]{\smash{0.6}}}%
    \put(0.05880198,0.65894303){\makebox(0,0)[lb]{\smash{0.8}}}%
    \put(0.08783222,0.7905468){\makebox(0,0)[lb]{\smash{1}}}%
    \put(0.41603725,0.00738336){\makebox(0,0)[lb]{\smash{FPs per case}}}%
    \put(0.02509665,0.33639278){\rotatebox{90}{\makebox(0,0)[lb]{\smash{Sensitivity}}}}%
  \end{picture}%
\endgroup%

%% file: roc_abs_chicos_1.pdf_tex
\begingroup%
  \makeatletter%
  \providecommand\color[2][]{%
    \errmessage{(Inkscape) Color is used for the text in Inkscape, but the package 'color.sty' is not loaded}%
    \renewcommand\color[2][]{}%
  }%
  \providecommand\transparent[1]{%
    \errmessage{(Inkscape) Transparency is used (non-zero) for the text in Inkscape, but the package 'transparent.sty' is not loaded}%
    \renewcommand\transparent[1]{}%
  }%
  \providecommand\rotatebox[2]{#2}%
  \ifx\svgwidth\undefined%
    \setlength{\unitlength}{415.588023bp}%
    \ifx\svgscale\undefined%
      \relax%
    \else%
      \setlength{\unitlength}{\unitlength * \real{\svgscale}}%
    \fi%
  \else%
    \setlength{\unitlength}{\svgwidth}%
  \fi%
  \global\let\svgwidth\undefined%
  \global\let\svgscale\undefined%
  \makeatother%
  \begin{picture}(1,0.80448666)%
    \put(0,0){\includegraphics[width=\unitlength]{roc_abs_chicos_1.pdf}}%
    \put(0.14751152,0.07104094){\makebox(0,0)[lb]{\smash{0}}}%
    \put(0.27071046,0.07104094){\makebox(0,0)[lb]{\smash{0.5}}}%
    \put(0.42470912,0.07104094){\makebox(0,0)[lb]{\smash{1}}}%
    \put(0.54983303,0.07104094){\makebox(0,0)[lb]{\smash{1.5}}}%
    \put(0.7038317,0.07104094){\makebox(0,0)[lb]{\smash{2}}}%
    \put(0.82703063,0.07104094){\makebox(0,0)[lb]{\smash{2.5}}}%
    \put(0.98295427,0.07104094){\makebox(0,0)[lb]{\smash{3}}}%
    \put(0.05277703,0.19540751){\makebox(0,0)[lb]{\smash{0.1}}}%
    \put(0.05277703,0.34170624){\makebox(0,0)[lb]{\smash{0.3}}}%
    \put(0.05277703,0.48800497){\makebox(0,0)[lb]{\smash{0.5}}}%
    \put(0.05277703,0.6343037){\makebox(0,0)[lb]{\smash{0.7}}}%
    \put(0.05277703,0.78060243){\makebox(0,0)[lb]{\smash{0.9}}}%
    \put(0.41380885,0.00734381){\makebox(0,0)[lb]{\smash{FPs per case}}}%
    \put(0.02496222,0.33459097){\rotatebox{90}{\makebox(0,0)[lb]{\smash{Sensitivity}}}}%
  \end{picture}%
\endgroup%

%% file: roc_clas_1.pdf_tex
\begingroup%
  \makeatletter%
  \providecommand\color[2][]{%
    \errmessage{(Inkscape) Color is used for the text in Inkscape, but the package 'color.sty' is not loaded}%
    \renewcommand\color[2][]{}%
  }%
  \providecommand\transparent[1]{%
    \errmessage{(Inkscape) Transparency is used (non-zero) for the text in Inkscape, but the package 'transparent.sty' is not loaded}%
    \renewcommand\transparent[1]{}%
  }%
  \providecommand\rotatebox[2]{#2}%
  \ifx\svgwidth\undefined%
    \setlength{\unitlength}{494.258bp}%
    \ifx\svgscale\undefined%
      \relax%
    \else%
      \setlength{\unitlength}{\unitlength * \real{\svgscale}}%
    \fi%
  \else%
    \setlength{\unitlength}{\svgwidth}%
  \fi%
  \global\let\svgwidth\undefined%
  \global\let\svgscale\undefined%
  \makeatother%
  \begin{picture}(1,0.56039154)%
    \put(0,0){\includegraphics[width=\unitlength]{roc_clas_1.pdf}}%
    \put(0.07602103,0.03692808){\makebox(0,0)[lb]{\smash{0}}}%
    \put(0.21360099,0.03692808){\makebox(0,0)[lb]{\smash{0.5}}}%
    \put(0.37869696,0.03692808){\makebox(0,0)[lb]{\smash{1}}}%
    \put(0.51789551,0.03692808){\makebox(0,0)[lb]{\smash{1.5}}}%
    \put(0.68137289,0.03692808){\makebox(0,0)[lb]{\smash{2}}}%
    \put(0.82057144,0.03692808){\makebox(0,0)[lb]{\smash{2.5}}}%
    \put(0.9856674,0.03692808){\makebox(0,0)[lb]{\smash{3}}}%
    \put(0.06145373,0.05796972){\makebox(0,0)[lb]{\smash{0}}}%
    \put(0.03717492,0.155085){\makebox(0,0)[lb]{\smash{0.2}}}%
    \put(0.03717492,0.25220027){\makebox(0,0)[lb]{\smash{0.4}}}%
    \put(0.03717492,0.34769695){\makebox(0,0)[lb]{\smash{0.6}}}%
    \put(0.03717492,0.44481222){\makebox(0,0)[lb]{\smash{0.8}}}%
    \put(0.06145373,0.54030891){\makebox(0,0)[lb]{\smash{1}}}%
    \put(0.45153341,0.00617491){\makebox(0,0)[lb]{\smash{FPs per case}}}%
    \put(0.02098904,0.24410733){\rotatebox{90}{\makebox(0,0)[lb]{\smash{Sensitivity}}}}%
  \end{picture}%
\endgroup%